\documentclass{article}

\PassOptionsToPackage{numbers, compress}{natbib}

\usepackage[preprint]{neurips_2023}

\usepackage[utf8]{inputenc} 
\usepackage[T1]{fontenc}    
\usepackage{hyperref}       
\usepackage{url}            
\usepackage{booktabs}       
\usepackage{amsfonts}       
\usepackage{nicefrac}       
\usepackage{microtype}      
\usepackage{xcolor}         

\usepackage{graphicx}
\usepackage{float}
\usepackage{multirow}
\usepackage{threeparttable}
\usepackage{amsmath,amsfonts}
\usepackage{wrapfig}

\title{MPPN: Multi-Resolution Periodic Pattern Network For Long-Term Time Series Forecasting}

\author{%
  Xing Wang,  Zhendong Wang,  Kexin Yang,  Junlan Feng\thanks{Corresponding author}, Zhiyan Song,  Chao Deng, Lin zhu\\
  JIUTIAN Team,
  China Mobile Research Institute,
  Beijing, China \\
  \texttt {\{wangxing, wangzhendongai, yangkexin\}@chiamobile.com} \\
\texttt{\{fengjunlan, songzhiyan, dengchao, zhulinyj\}@chinamobile.com}
  }

\begin{document}

\maketitle

\begin{abstract}
  Long-term time series forecasting plays an important role in various real-world scenarios. Recent deep learning methods for long-term series forecasting tend to capture the intricate patterns of time series by decomposition-based or sampling-based methods. However, most of the extracted patterns may include unpredictable noise and lack good interpretability. 
  Moreover, the multivariate series forecasting methods usually ignore the individual characteristics of each variate, which may affecting the prediction accuracy. 
  To capture the intrinsic patterns of time series, we propose a novel deep learning network architecture, named Multi-resolution Periodic Pattern Network (MPPN), for long-term series forecasting. 
  We first construct context-aware multi-resolution semantic units of time series and employ multi-periodic pattern mining to capture the key patterns of time series. Then, we propose a channel adaptive module to capture the perceptions of multivariate towards different patterns.
  In addition, we present an entropy-based method for evaluating the predictability of time series and providing an upper bound on the prediction accuracy before forecasting. 
  Our experimental evaluation on nine real-world benchmarks demonstrated that MPPN significantly outperforms the state-of-the-art Transformer-based, decomposition-based and sampling-based methods for long-term series forecasting.


\end{abstract}

\section{Introduction}
Time series forecasting is a long-standing problem and has been widely used in weather forecasting, energy management, traffic flow scheduling, and financial planning. Long-term time series forecasting (LTSF) means predicting further into the future, which can provide sufficient reference for long-term planning applications and is of great importance. This paper focuses on long-term time series forecasting problem. Most of the typical methods for LTSF task before treated time series as a sequence of values, similar to the sequence in speech and natural language processing. Specifically, the encoding of a lookback window of historical time series values, along with time feature embedding (e.g., Hour of Day, Day of Week and Day of Month) and positional encoding, are combined as the model input sequence. And then the convolution-based \cite{wang2023micn} or Transformer-based techniques \cite{zhou2021informer, liu2021pyraformer} are used to extract the intricate correlations or high-dimensional features of time series to achieve long-term sequence prediction. 

Different from other types of sequential data, time series data only record scalars at each moment. Data of solitary points cannot provide adequate semantic information and might contain noise. 
Therefore, some works implement sub-series \cite{wu2021autoformer} or segments \cite{wang2023micn, zhang2023crossformer} as the basic semantic tokens aiming to capture the inherent patterns of time series. However, the patterns of time series are intricate and usually entangled and overlapped with each other, which are extremely challenging to clarify. Without making full use of the properties of time series (e.g., period), relying solely on the self-attention or convolution techniques to capture the overlapped time series patterns can hardly avoid extracting noise patterns. 
In addition, most of the multivariate time series prediction methods \cite{NEURIPS2022_7b102c90, zhang2023crossformer}  mainly focus on modeling the correlations between variates and ignore the individual characteristics of each variate, which may affecting the prediction accuracy.


Existing typical methods based on capturing time series patterns for LTSF tasks can be categorized into decomposition-based and sampling-based. Decomposition-based methods attempt to decompose the time series into more predictable parts and predict them separately before aggregating the results \cite{wu2021autoformer, zhou2022fedformer, wang2023micn, oreshkin2019n, Zeng2022AreTE}. FEDformer \cite{zhou2022fedformer} and MICN \cite{wang2023micn} have further proposed multi-scale hybrid decomposition approach based on Moving Average to extract various seasonal and trend-cyclical parts of time series. However, the real-world time series are usually intricate which are influenced by multiple factors and can be hardly disentangled simply by seasonal-trend decomposition. To the best of our knowledge, most sampling-based methods implement downsampling techniques. A typical motivation is that downsampling a time series can still preserve the majority of its information \cite{liu2022scinet, zhang2022less}. Although current sampling-based methods can partially degrade the complexity of time series and improve the predictability of the original series \cite{liu2022scinet}, they can easily suffer from the influence of outliers or noise in time series, which reduces the quality of the extracted patterns and affects their performance in LTSF tasks.

\begin{wrapfigure}{r}{5cm}
\centering
\includegraphics[width=0.38\textwidth]{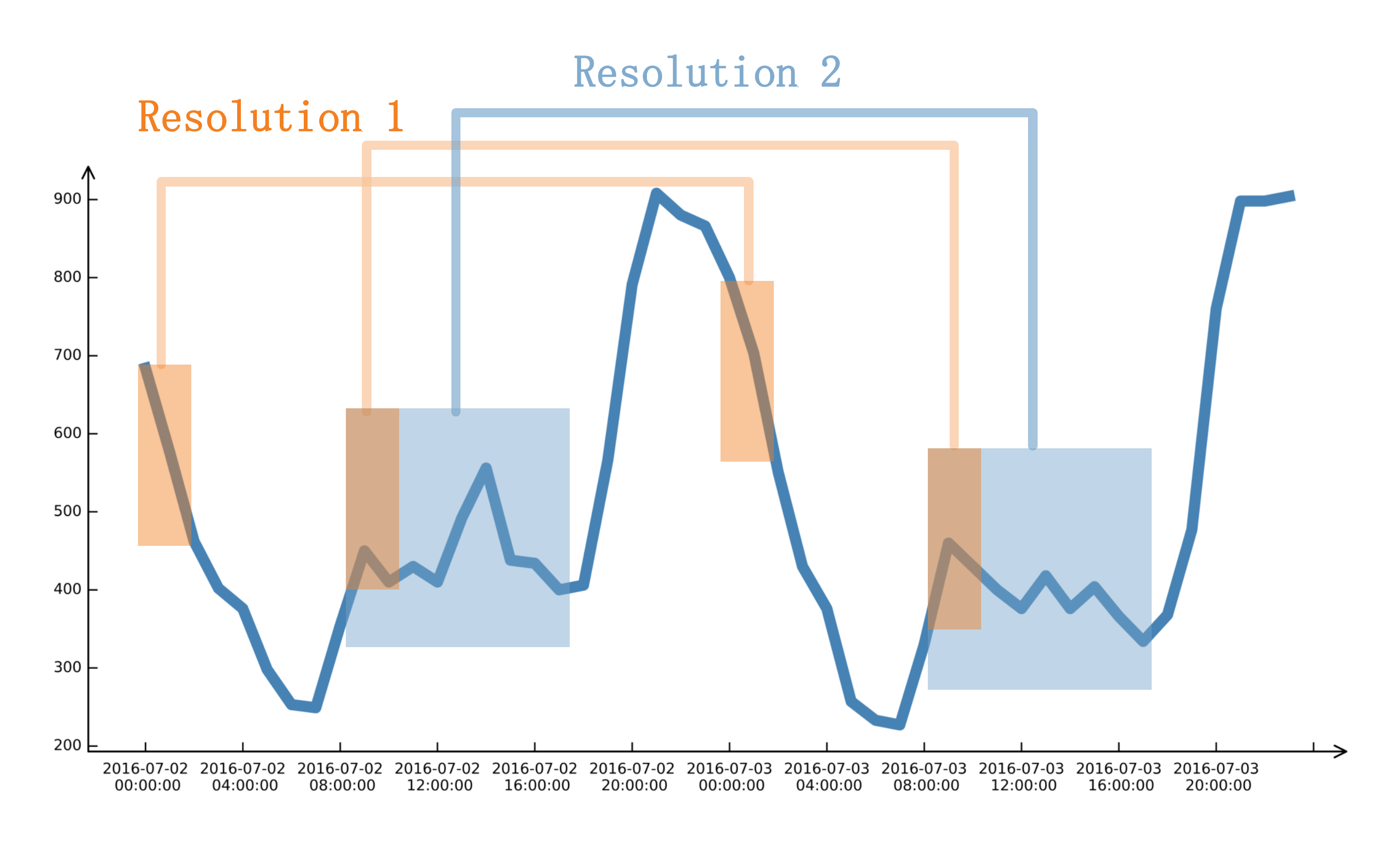}
\caption{An example of time series with the multi-resolution periodic patterns. 
It displays a client's electricity consumption for two days and shows that the series generally peaks between 8 PM and midnight, while during daytime, it maintains a low range, presumably implying that the resident is outside working.
}
\label{patch}
\end{wrapfigure}

We believe that, analogous to speech and natural language, time series have their own distinctive patterns that can represent them. The challenge lies in how to extract these patterns. For the same variate, we observe that time series with larger resolutions often exhibit stronger periodicity, whereas those with smaller resolutions tend to have more fluctuations, as shown in Figure \ref{patch}. Motivated by this, we thought that a time series can be seen as an overlay of multi-resolution patterns. 
Moreover, time series possess regular patterns, which is why we can predict them. One obvious observation is that real-world time series, such as electricity consumption and traffic, usually exhibit daily and weekly periods. Therefore, we attempt to capture the multi-periodicity of time series to decode their unique characteristics. Further, for multivariate series prediction task, each variate has its own characteristic and perception of different patterns. Existing methods frequently employ the same model parameters, which can only model the commonalities among the multiple variates, without taking into account the individualities of each variate. 


Based on the above motivations, we propose a novel deep learning network architecture, named Multi-resolution Periodic Pattern Network (MPPN) for long-term time series forecasting.
Firstly, we construct context-aware multi-resolution semantic units of the time series and propose a multi-periodic pattern mining mechanism for capturing the distinctive patterns in time series. Secondly, we propose a channel adaptive module to infer the variate embedding (attributes) from data during training and to perform adaptive weighting on the mined patterns. In addition, previous works suggested that series decomposition may help extract more predictable components \cite{wu2021autoformer}, but no theoretical proof was provided. We argue that before predicting a time series, it should be evaluated whether the series is predictable or not. Therefore, in this paper, we introduce an entropy-based method for evaluating the predictability of time series and providing an upper bound on how predictable the time series is before carrying out predictions. The contributions of this paper are summarized as follows:

\begin{itemize}
\item We propose a novel framework MPPN to explicitly capture the inherent multi-resolution and multi-periodic patterns of time series for efficient and accurate long-term series forecasting.
\item We propose a channel adaptive module to adaptively model different perceptions of multivariate series towards various temporal patterns, further improving the prediction performance.
\item Experimental evaluations on nine real-world benchmarks demonstrate that our MPPN significantly outperforms the state-of-the-art Transformer-based, decomposition-based and sampling-based methods in LTSF tasks, while maintaining linear computational complexity.

\end{itemize}

\section{Related work}
In the past several decades, numerous methods for time series forecasting have been developed, evolving from conventional statistics (such as ARIMA \cite{williams2003modeling}) and machine learning (such as Prophet \cite{taylor2018forecasting}) to the current deep learning. 
Especially, deep learning has gained popularity owing to its strong representation ability and nonlinear modeling capacity.
Typical deep learning-based methods include RNN \cite{lai2018modeling}, TCN \cite{bai2018empirical} and Transformer \cite{vaswani2017attention}. 
Transformer-based methods with self-attention mechanism are frequently used for LTSF task \cite{zhou2022fedformer, wu2021autoformer, zhang2023crossformer,zhou2021informer, liu2021pyraformer}. To alleviate the issue of quadratic complexity in sequence length with self-attention, LogTrans \cite{li2019enhancing} proposes the LogSparse attention reducing the complexity to $\mathcal{O}(L(\log L)^2)$. Similarly, Informer \cite{zhou2021informer} introduces the ProbSparse attention based on KL-divergence, achieving $\mathcal{O}(L\log L)$ complexity. Autoformer \cite{wu2021autoformer} proposes the series-wise attention based on auto-correlation to improve computational efficiency and the utilization of sequence information, along with an inner decomposition block specifically designed to capture intricate temporal patterns in long-term time series. FEDFormer \cite{zhou2022fedformer} enhances the decomposed Transformer by incorporating the frequency domain representations of time series for attention computation. Additionally, FEDFormer achieves linear computational complexity and memory cost through the random selection of a constant number of Fourier components.

Although Transformer-based methods have achieved impressive performance, recent research \cite{Zeng2022AreTE} have questioned whether they are suitable for LTSF tasks, especially since the permutation-invariant self-attention mechanism causes loss of temporal information. They have shown that an embarrassingly simple linear model outperforms all Transformer-based models. This highlights the importance of focusing on intrinsic properties of time series.
Recently, sampling-based methods in conjunction with convolution have achieved remarkable results for LTSF tasks. SCINet \cite{liu2022scinet} adopts a recursive downsample-convolve-interact architecture that downsamples the sequence into two sub-sequences (odd and even) recursively to extract time series patterns. MICN \cite{wang2023micn} implements a multi-scale branch structure with down-sampled convolution for local features extraction and isometric convolution for capturing global correlations. Although these methodologies exhibit better performance compared to Transformer-based models in LTSF task, they neglect intrinsic properties of time series and patterns extracted based on global indiscriminate downsampling may contain noise.


\section{Methodology}
\label{method}

\begin{figure}[htbp]
    \centering
 \includegraphics[width=5.2in,height=2in]{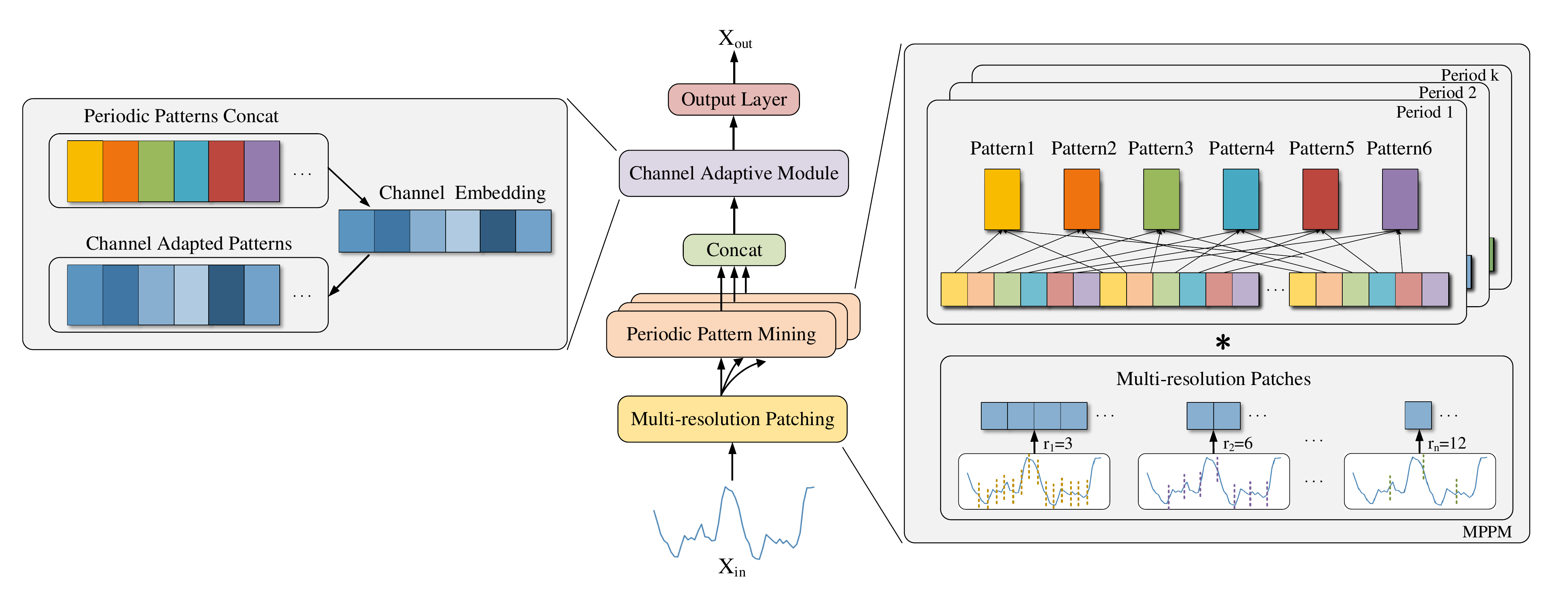}
    \caption{The overall architecture of Multi-resolution Periodic Pattern Network (MPPN).}
    \label{model}
\end{figure}

In this section, we first present the problem definition of the multivariate time series forecasting task and introduce a quantitative evaluation of predictability. Then we introduce our proposed MPPN method. 
The overall architecture of the MPPN model is illustrated in Figure \ref{model}. It consists of Multi-resolution Periodic Pattern Mining (MPPM), a channel adaptive module, and an output layer.

\subsection{Problem definition}
\label{problem}
Multivariate time series prediction aims to forecast future values of multiple variates based on their historical observations. Considering a multivariate time series $\boldsymbol{X}=\left[\boldsymbol{x}_{1}, \ldots, \boldsymbol{x}_{t}, \ldots, \boldsymbol{x}_{T}\right]^{T}\in\mathbb{R}^{T\times C}$ consisting of $T$ time steps and $C$ recorded variates, where $\boldsymbol{x}_t\in\mathbb{R}^C$ represents an observation of the multivariate time series at time step $t$. We set the look-back window length as $L$ and the length of the forecast horizon as $H$.
Then, given the historical time series $\boldsymbol{X}_{in}=\left[\boldsymbol{x}_{t-L}, \ldots, \boldsymbol{x}_{t-1}\right]^{T}\in \mathbb{R}^{L\times C}$, the forecasting objective is to learn a mapping function $\mathcal{F}$ that predicts the values for the next $H$ time steps  $\boldsymbol{X}_{out}=\left[\boldsymbol{x}_{t}, \ldots, \boldsymbol{x}_{t+H}\right]^{T}\in \mathbb{R}^{H\times C}$:
\begin{align}
\left[\boldsymbol{x}_{t-L}, \ldots, \boldsymbol{x}_{t-1}\right]^{T} \stackrel{\mathcal{F}}{\longrightarrow}\left[\boldsymbol{x}_{t}, \ldots, \boldsymbol{x}_{t+H}\right]^{T}. \label{pred_problem}
\end{align}

\subsection{Predictability}
\label{predictability}
Predictability is a measure that quantifies the confidence in the predictive capability for a time series, providing an upper bound on the accuracy possibly achieved by any forecasting approach. As the foundation of time series prediction, predictability explains to what extent the future can be foreseen, which is often overlooked by prior deep learning-based temporal forecasting methods. In the context of time series prediction, the foremost importance lies not in the construction of predictive models, but rather in the determination of whether the time series itself is predictable. Based on the determinations, it becomes possible to filter out time series with low predictability, such as random walk time series, thereby discerning the meaningfulness of the prediction problem at hand. There exists a multitude of seminal works in the domain of predictability \cite{song2010limits,xu2019predictability, guo2021can, smith2014refined}, among which the most commonly employed approach is based on entropy measures.

For long-term time series forecasting, we firstly evaluate the predictability following the method in \cite{song2010limits}, 
 which explored the predictability of human mobility trajectories using entropy rates. Firstly, we discretize the continuous time series into $Q$ discrete values. 
 Denote $\mathbb{X}=\{X_1,X_2,\cdots,X_n\}$ as a time series after discretization, its entropy rate is defined as follows:
\begin{align}
\mathcal{H}_u(\mathbb{X}) = \lim _{n \rightarrow \infty} \frac{1}{n} \sum_{i=1}^n H\left(X_i \mid X_{i-1}, \cdots, X_1\right),\label{entropy_rate}
\end{align}
which characterizes the average conditional entropy $H$ of the current variable given the values of all the past variables as $n\to\infty$. In order to calculate this theoretical value, we utilize an estimator based on Lempel-Ziv encoding \cite{kontoyiannis1998nonparametric}, which has been proven to be a consistent estimator of the real entropy rate $\mathcal{H}_u(\mathbb{X})$. For $\mathbb{X}$, the entropy rate is estimated by
\begin{align}
S=\left(\frac{1}{n} \sum_{i=1}^n \Lambda_i\right)^{-1} \ln (n),\label{Lempel}
\end{align}
where $\Lambda_i$ signifies the minimum length of the sub-string starting at position $i$ that has not been encountered before from position $1$ to $i-1$. We further derive the upper bound of predictability $\Pi^{\max}$ by solving the following Fano's inequality \cite{kontoyiannis1998nonparametric}:
\begin{align}
S\leq H(\Pi^{\max})+(1-\Pi^{\max})\log_2(N-1),\label{Fano}
\end{align}
where $H(\Pi^{\max})=-\Pi^{\max}\log_2(\Pi^{\max})-(1-\Pi^{\max})\log_2(1-\Pi^{\max})$ represents the binary entropy function and $N$ is the number of distinct values in $\mathbb{X}$. It is worth noting that the inequality \eqref{Fano} is tight, in the sense that the upper bound of predictability is attainable by some actual algorithm. As a result, the upper bound of predictability provides a theoretical guarantee for conducting long-term time series forecasting.


\subsection{Multi-resolution periodic pattern mining}
\label{MPPN}

The MPPM is composed of two key components, namely multi-resolution patching and periodic pattern mining, which are specially designed to capture intricate multi-resolution patterns inherent in time series data with multiple periods. For simplicity, we omit the channel dimension $C$ and denote the hidden state of the series as $D$.

\paragraph{Multi-resolution patching}
To capture the multi-resolution patterns in time series data, we first obtain context-aware semantic units of the time series. Specifically, as shown in Figure \ref{model}, we employ non-overlapping multi-scale convolutional kernels ( inception mechanism \cite{szegedy2016rethinking}) to partition the input historical time series $\boldsymbol{X}_{in}$ into multi-resolution patches. For instance, for a time series with a granularity of 1 hour and assuming a resolution of 3 hours, the above-mentioned convolution with a kernel size 3 is used to map $\boldsymbol{X}_{in}$ to the output $\boldsymbol{X}_{r}$. This process can be formulated as follows:
\begin{equation}
\boldsymbol{X}_{r}=\operatorname{Conv} 1 d\left(\operatorname{Padding}\left(\boldsymbol{X}_{in}\right)\right)_{\text {kernel }=r},
\end{equation}
where, $r$ denotes the convolutional kernel size that correspond to the pre-defined temporal resolution. For Conv1d, we set the $kernel$ and $stride$ both to be $r$. For the resolution selection, we choose a set of reasonable resolutions $r \in\{{r}_{1}, {r}_{2}, \cdots, {r}_{n}\}$ based on the granularity of the input time series (See Appendix A.4 
for more details). $\boldsymbol{X}_{r}$
denotes the obtained semantic units of the time series corresponding to resolution $r$.

\paragraph{Periodic pattern mining}
We implement periodic pattern mining to explicitly capture the multi-resolution and multi-periodic patterns in time series data. We firstly employ Fast Fourier Transform (FFT) to calculate the periodicity of the original time series, following the periodicity computation method proposed by Wu et al. \cite{wu2022timesnet}.
Briefly, we take the Fourier transform of the original time series $\boldsymbol{X}$ and calculate the amplitude of each frequency. We then select the top-$k$ amplitude values and obtain a set of the most significant frequencies $\left\{f_1, \cdots, f_k\right\}$ corresponding to the selected amplitudes, where $k$ is a hyperparameter. 
Consequently, we acquire $k$ periodic lengths $\left\{\operatorname{Period}_{1}, \cdots, \operatorname{Period}_{k}\right\}$ that correspond to these frequencies. Similar to \cite{wu2022timesnet}, we only consider frequencies within $\left\{1, \cdots,\left[\frac{T}{2}\right]\right\}$. The process is summarized as follows:
\begin{equation}
\mathbf{A}=\operatorname{Avg}\left(\operatorname{Amp}\left(\operatorname{FFT}\left(\boldsymbol{X}\right)\right)\right),\left\{f_1, \cdots, f_k\right\}=\underset{f_* \in\left\{1, \cdots,\left[\frac{T}{2}\right]\right\}}{\arg \operatorname{Topk}}(\mathbf{A}), \,
\operatorname{Period}_{i}=\left\lceil\frac{T}{f_i}\right\rceil, 
\end{equation}
where $i \in\{1, \cdots, k\}$, $\operatorname{FFT}(\cdot)$ represents the FFT, and $\operatorname{Amp}(\cdot)$ denotes the amplitude calculation. $\mathbf{A} \in \mathbb{R}^T$ denotes the calculated amplitudes, which are determined by taking the average of $C$ variates using $\operatorname{Avg}(\cdot)$ .

We then utilize the periodicity calculated above and employ dilated convolutions to achieve multi-periodic and multi-resolution pattern mining. Specifically, given a periodic length of $\operatorname{Period}_{i}$ and the resolution $r$, we set convolution dilation as $\left\lfloor\frac{\operatorname{Period}_{i}}{r}\right\rfloor$ and kernel size as $\left\lfloor\frac{L}{\operatorname{Period}_{i}}\right\rfloor$ for the convolution operation on $\boldsymbol{X}_{r}$. To obtain regularized patterns, we perform truncation on the outputs of dilated convolution. The process can be formulated as follows:
\begin{equation}
\boldsymbol{X}_{{\operatorname{Period}_{i}},r}=\operatorname{Truncate}\left(\operatorname{Conv} 1 d\left(\boldsymbol{X}_{r}\right)_{\text {kernel }=\left\lfloor\frac{L}{\operatorname{Period}_{i}}\right\rfloor, \text {dilation }=\left\lfloor\frac{\operatorname{Period}_{i}}{r}\right\rfloor}\right),
\end{equation}
where $\boldsymbol{X}_{{\operatorname{Period}_{i}},r} \in R^{\left\lfloor\frac{\operatorname{Period}_{i}}{r}\right\rfloor \times D}$ denotes the patterns extracted corresponding to $\operatorname{Period}_{i}$ and resolution $r$. For the same period, we concatenate all the corresponding $\boldsymbol{X}_{{\operatorname{Period}_{i}},r}$ of different resolutions $r$ to obtain its whole pattern $\boldsymbol{X}_{\operatorname{Period}_{i}}$. We then concatenate the patterns of multiple periods to obtain the final multi-periodic pattern of the time series, denoted as $\boldsymbol{X}_{\operatorname{Pattern}} \in R^{P \times D}$, formulated as follows:
\begin{align}
\boldsymbol{X}_{\operatorname{Period}_i}=
\underset{j=1}{\overset{n}{\|}} 
\boldsymbol{X}_{\operatorname{Period}_{i},r_j},\,
\boldsymbol{X}_{\operatorname{Pattern}}=\underset{i=1}{\overset{k}{\|}}
\boldsymbol{X}_{\operatorname{Period}_i}, \,
P=\sum_{i=1}^k \sum_{j=1}^n \left\lfloor\frac{\operatorname{Period}_i}{r_j} \right\rfloor. \label{multi-periodic}
\end{align}




\subsection{Channel adaptive module}
To achieve adaptivity for each variate, we propose a channel adaptive mechanism.
We firstly define a learnable variate embeddings matrix $\boldsymbol{E} \in R^{C \times P}$, which can be updated during model training, where $P$ represents the number of pattern modes extracted by the above MPPM.
Next, we apply the $\operatorname{sigmoid}$ function to activate the learned variate representation $\boldsymbol{E}$ and then perform broadcasting multiplication with the obtained multi-resolution periodic pattern $\boldsymbol{X}_{\operatorname{Pattern}}$, producing the final channel adaptive patterns $\boldsymbol{X}_{\operatorname{AdpPattern}} \in R^{C \times P \times D}$, formulated as follows:
\begin{equation}
\boldsymbol{X}_{\operatorname{AdpPattern}}=\boldsymbol{X}_{\operatorname{Pattern}}\cdot\operatorname{sigmoid}\left(\boldsymbol{E}\right),
\end{equation}

At last, we implement the output layer with one fully connected layer to generate the final long-term prediction $\boldsymbol{X}_{out} \in \mathbb{R}^{H\times C}$. The output layer can be formulated as follows:
\begin{equation}
    \boldsymbol{X}_{out}=\operatorname {Reshape}\left(\boldsymbol{X}_{\operatorname{AdpPattern}}\right)\cdot {W}+b,
\end{equation}
where $W\in \mathbb{R}^{(PD)\times H}$ and $b \in \mathbb{R}^H$ are learnable parameters. $\boldsymbol{X}_{out}$ is the final output of the MPPN.
Finally, we adopt the Mean Squared Error (MSE) as the training loss to optimize the model.


\section{Experiments}
In this section, we present the experimental evaluation of our MPPN model compared to state-of-the-art baseline models. Additionally, we conduct comprehensive ablation studies and perform model analysis to demonstrate the effectiveness of each module in MPPN. More detailed information can be found in the Appendix.

\subsection{Experimental settings}

\paragraph{Datasets} We conduct extensive experiments on nine widely-used time series datasets, including four \emph{ETT} \cite{zhou2021informer} (ETTh1, ETTh2, ETTm1, ETTm2), \emph{Electricity}, \emph{Exchange-Rate} \cite{lai2018modeling}, \emph{Traffic}, \emph{Weather} and \emph{ILI} dataset. A brief description of these datasets is presented in Table~\ref{dataset-table}. We provide a detailed dataset description in Appendix A.1.

\begin{table}[htbp]
	\caption{Dataset statistics.}
	\label{dataset-table}
	\centering
	\resizebox{\linewidth}{!}{
	\begin{tabular}{cccccccc}
		\toprule
		Datasets    & Electricity & Weather & Traffic & Exchange-Rate & ILI & ETTh1\&ETTh2 & ETTm1\&ETTm2 \\
		\midrule
		Timesteps & 26,304  & 52,696 & 17,544 & 7,588 & 966 & 17,420 & 69,680\\
		Features     & 321 & 21 & 862 & 8 & 7 & 7 & 7  \\
		Granularity     & 1hour & 10min & 1hour & 1day & 1week & 1hour & 15min \\
		\bottomrule
	\end{tabular}
	}
\end{table}

\paragraph{Baselines} We adopt two up-to-date Linear-based models: DLinear and NLinear \cite{Zeng2022AreTE}, as well as four latest state-of-the-art Transformer-based models: Crossformer \cite{zhang2023crossformer}, FEDformer \cite{zhou2022fedformer}, Autoformer \cite{wu2021autoformer}, Informer \cite{zhou2021informer}, and two CNN-based models: MICN \cite{wang2023micn} and SCINet \cite{liu2022scinet} as baselines. For FEDformer, we select the better one (FEDformer-f, which utilizes Fourier transform) for comparison. Additional information about the baselines can be found in Appendix A.3.

\paragraph{Implementation details} Our model is trained with L2 loss, using the ADAM \cite{kingma2014adam} optimizer with an initial learning rate of 1e-3 and weight decay of 1e-5. The training process is early stopped if there is no loss reduction on the validation set after three epochs. All experiments are conducted using PyTorch and run on a single NVIDIA Tesla V100 GPU. Following previous work \cite{zhou2022fedformer, wu2021autoformer, wang2023micn}, we use Mean Square Error (MSE) and Mean Absolute Error (MAE) as evaluation metrics. See Appendix A.4
for more detailed information.

\subsection{Main results}

\begin{table}[htbp]
	\caption{Predictability and periodicity results of the nine benchmark datasets.}
	\label{table_pre_per}
	\centering
	\resizebox{\linewidth}{!}{
	\begin{tabular}{cccccccccc}
		\toprule
		Datasets    & Electricity & Weather & Traffic & Exchange-Rate & ILI & ETTh1 & ETTh2 & ETTm1 & ETTm2 \\
		\midrule
		Timesteps & 26,304  & 52,696 & 17,544 & 7,588 & 966 & 17,420 & 17,420 & 69,680 & 69,680 \\
		Predictability     & 0.876 & 0.972 & 0.934 & 0.973 & 0.917 & 0.853 & 0.927 & 0.926 & 0.967  \\
		 Top-1 period     & 24 & 144 & 12 & -- & -- & 24 & --& 96 & --  \\
		\bottomrule
	\end{tabular}
	}
\end{table}

\paragraph{Predictability analysis}
As a prerequisite, we investigate the predictability of the nine public datasets before constructing prediction models. For each dataset, a quantitative metric is provided in accordance with the method outlined in Section \ref{predictability}. We designate the average predictability of distinct univariate datasets as the measure of predictability for each benchmark dataset in question. The corresponding results are summarized in Table \ref{table_pre_per}. It can be observed from Table \ref{table_pre_per} that all predictability results exceed 0.85, indicating the nine benchmark datasets exhibit a notable level of predictability. This provides sufficient confidence and theoretical assurance for constructing excellent prediction models upon the nine benchmark datasets.

\paragraph{Periodicity analysis} 
To capture the intricate temporal patterns and multiple-periodicity within the time series, we employ the method of FFT as discussed in Section \ref{method} to extract the Top-1 period for each of the nine benchmark datasets. See the last row of Table \ref{table_pre_per} for details. It can be concluded from Table \ref{table_pre_per} that  Electricity, Weather, 
Traffic, ILI, ETTh1, and ETTm1 exhibit noticeable periodic patterns, while the remaining datasets do not possess discernible periodicities, indicating the manifestation of more complex time series patterns. 

\begin{table}[h]
	\caption{\textbf{Multivariate} long-term time series forecasting results with different prediction length $O\in\{24, 36, 48, 60\}$ for ILI dataset and $O\in\{96, 192, 336, 720\}$ for others. The SOTA results are \textbf{bolded}, while the sub-optimal results are \underline{underlined}. Correspondingly, IMP. shows the percentage improvement of MPPN over the state-of-the-art baseline models.}
	\label{multivariate-table}
	\centering
	\begin{threeparttable}
		\renewcommand{\multirowsetup}{\centering}
		\resizebox{\linewidth}{!}{
			\begin{tabular}{lcllllllllllllllllllc}
				\toprule
				\multicolumn{2}{c}{Models} & \multicolumn{2}{c}{\textbf{MPPN}} & \multicolumn{2}{c}{NLinear} & \multicolumn{2}{c}{DLinear} & \multicolumn{2}{c}{SCINet} & \multicolumn{2}{c}{MICN} & \multicolumn{2}{c}{FEDformer} & \multicolumn{2}{c}{Autoformer} & \multicolumn{2}{c}{Crossformer} & \multicolumn{2}{c}{Informer} & {IMP.}\\
				\cmidrule(lr){3-4} \cmidrule(lr){5-6} \cmidrule(lr){7-8} \cmidrule(lr){9-10} \cmidrule(lr){11-12} \cmidrule(lr){13-14} \cmidrule(lr){15-16} \cmidrule(lr){17-18} \cmidrule(lr){19-20} \cmidrule(lr){21-21}
				\multicolumn{2}{c}{Metric} &MSE & MAE & MSE & MAE & MSE & MAE & MSE & MAE & MSE & MAE & MSE & MAE & MSE & MAE & MSE & MAE & MSE & MAE & MSE\\
				\toprule
				\multirow{4}{*}{\rotatebox{90}{Weather}} & 96  & \textbf{0.144} & \textbf{0.196} & 0.182 & \underline{0.232} & 0.174 & 0.233 & 0.184 & 0.242 & \underline{0.170} & 0.235 & 0.219 & 0.300 & 0.263 & 0.332 & 2.320 & 0.877 & 0.375 & 0.437  & 15.29\% \\
				& 192 & \textbf{0.189} & \textbf{0.240} & 0.225 & \underline{0.269} & \underline{0.218} & 0.278 & 0.244 & 0.298 & 0.223 & 0.285 & 0.271 & 0.331 & 0.295 & 0.354 & 2.370 & 0.919 & 0.483 & 0.496  & 13.30\% \\
				& 336 & \textbf{0.240} & \textbf{0.281} & 0.271 & \underline{0.301} & \underline{0.263} & 0.314 & 0.287 & 0.322 & 0.278 & 0.339 & 0.318 & 0.354 & 0.346 & 0.385 & 3.839 & 1.243 & 0.588 & 0.542 & 8.75\% \\
				& 720 & \textbf{0.310} & \textbf{0.333} & 0.339 & \underline{0.349} & \underline{0.332} & 0.374 & 0.346 & 0.360 & 0.342 & 0.386 & 0.410 & 0.419 & 0.428 & 0.433 & 5.161 & 1.494 & 1.061 & 0.755 & 6.63\%  \\
				\midrule
				\multirow{4}{*}{\rotatebox{90}{Traffic}} & 96 & \textbf{0.387} & \textbf{0.271} & \underline{0.412} & \underline{0.282} & 0.413 & 0.287 & 0.444 & 0.281 & 0.524 & 0.307 & 0.588 & 0.368 & 0.644 & 0.415 & 0.530 & 0.293 & 0.746 & 0.416 &  6.07\%   \\
				& 192 & \textbf{0.396} & \textbf{0.273} & 0.425 & \underline{0.287} & \underline{0.424} & 0.290 & 0.528 & 0.321 & 0.541 & 0.315 & 0.606 & 0.373 & 0.623 & 0.386 & 0.560 & 0.307 & 0.483 & 0.496 & 6.60\% \\
				& 336 & \textbf{0.410} & \textbf{0.279} & \underline{0.437} & \underline{0.293} & 0.438 & 0.299 & 0.531 & 0.321 & 0.540 & 0.312 & 0.629 & 0.390 & 0.620 & 0.385 & 0.585 & 0.324 & 0.588 & 0.542 & 6.18\%  \\
				& 720 & \textbf{0.449} & \textbf{0.301} & \underline{0.465} & \underline{0.311} & 0.466 & 0.316 & 0.620$^\ast$ & 0.394$^\ast$ & 0.599 & 0.324 & 0.627 & 0.381 & 0.677 & 0.418 & 0.591 & 0.314 & 1.046 & 0.588  & 3.44\%\\
				\midrule
				\multirow{4}{*}{\rotatebox{90}{Electricity}} & 96  & \textbf{0.131} & \textbf{0.226} & 0.141 & \underline{0.237} & \underline{0.140} & \underline{0.237} & 0.167 & 0.269 & 0.163 & 0.269 & 0.189 & 0.305 & 0.202 & 0.317 & 0.220 & 0.303 & 0.323 & 0.409  &  6.43\% \\
				& 192 & \textbf{0.145} & \textbf{0.239} & \underline{0.154} & \underline{0.249} & \underline{0.154} & 0.250 & 0.174 & 0.280 & 0.178 & 0.286 & 0.197 & 0.312 & 0.233 & 0.338 & 0.279 & 0.347 & 0.347 & 0.431  & 5.84\%  \\
				& 336 & \textbf{0.162} & \textbf{0.256} & 0.171 & \underline{0.265} & \underline{0.169} & 0.268 & 0.186 & 0.292 & 0.184 & 0.293 & 0.214 & 0.329 & 0.260 & 0.359 & 0.336 & 0.370 & 0.349 & 0.432  &  4.14\% \\
				& 720 & \textbf{0.200} & \textbf{0.289} & 0.210 & \underline{0.298} & \underline{0.204} & 0.300 & 0.231$^\ast$ & 0.316$^\ast$ & 0.214 & 0.324 & 0.244 & 0.351 & 0.256 & 0.361 & 0.429 & 0.441 & 0.394 & 0.457 & 1.96\% \\
				\midrule
				\multirow{4}{*}{\rotatebox{90}{Exchange}} & 96  & 0.089 & \textbf{0.204} & 0.089 & 0.208 & \underline{0.085} & 0.209 & 0.116 & 0.254 & \textbf{0.082} & \underline{0.205} & 0.134 & 0.262 & 0.143 & 0.273 & 0.257 & 0.383 & 0.947 & 0.771   &  -- \\
				& 192 & 0.177 & \textbf{0.295} & 0.180 & 0.300 & \underline{0.162} & \underline{0.296} & 0.218 & 0.345 & \textbf{0.157} & 0.298 & 0.261 & 0.372 & 0.271 & 0.380 & 0.878 & 0.732 & 1.244 & 0.882  &  -- \\
				& 336 & 0.344 & 0.418 & 0.331 & 0.415 & 0.333 & 0.441 & \underline{0.294} & \underline{0.413} & \textbf{0.269} & \textbf{0.402} & 0.442 & 0.494 & 0.456 & 0.506 & 1.149 & 0.858 & 1.792 & 1.070 & -- \\
				& 720 & 0.929 & 0.731  & 0.943 & 0.728 & \underline{0.898} & \underline{0.725} & 1.110 & 0.767 & \textbf{0.701} & \textbf{0.653} & 1.125 & 0.820 & 1.090 & 0.812 & 1.538 & 1.002 & 2.936 & 1.415 & -- \\
				\midrule
				\multirow{4}{*}{\rotatebox{90}{ILI}} & 24  & \textbf{1.796} & \textbf{0.860} & 2.285 & \underline{0.983} & \underline{2.280} & 1.061 & 3.409 & 1.245 & 3.031 & 1.180 & 3.221 & 1.242 & 3.410 & 1.296 & 3.197 & 1.199 & 5.248 & 1.580   &  21.23\% \\
				& 36 & \textbf{1.748} & \textbf{0.840} & \underline{2.119} & \underline{0.938} & 2.235 & 1.059 & 3.200 & 1.204 & 2.507 & 1.011 & 2.660 & 1.071 & 3.365 & 1.252 & 3.191 & 1.210 & 5.057 & 1.561   &  17.51\%  \\
				& 48 & \textbf{1.692} & \textbf{0.840} & \underline{2.062} & \underline{0.933} & 2.298 & 1.079 & 2.943 & 1.187 & 2.427 & 1.013 & 2.717 & 1.101 & 3.125 & 1.200 & 3.450 & 1.205 & 5.110 & 1.564 & 17.94\% \\
				& 60 & \textbf{1.840} & \textbf{0.881} & \underline{2.258} & \underline{0.994} & 2.573 & 1.157 & 2.719 & 1.189 & 2.654 & 1.085 & 2.840 & 1.148 & 2.847 & 1.146 & 3.505 & 1.264 & 5.397 & 1.606 & 18.51\% \\
				\midrule
				\multirow{4}{*}{\rotatebox{90}{ETTh1}} & 96  & \textbf{0.371} & \textbf{0.393} & \underline{0.374} & \underline{0.394} & 0.384 & 0.405 & 0.405 & 0.428 & 0.405 & 0.431 & 0.377 & 0.416 & 0.436 & 0.448 & 0.426 & 0.436 & 0.934 & 0.764 & 0.80\% \\
				& 192 & \textbf{0.405} & \textbf{0.413} & \underline{0.408} & \underline{0.415} & 0.443 & 0.450 & 0.470 & 0.470 & 0.501 & 0.489 & 0.424 & 0.446 & 0.444 & 0.451 & 0.585 & 0.547 & 1.006 & 0.785 & 0.74\% \\
				& 336 & \textbf{0.426} & \textbf{0.425} & \underline{0.429} & \underline{0.427} & 0.447 & 0.448 & 0.530 & 0.514 & 0.541 & 0.528 & 0.450 & 0.463 & 0.516 & 0.493 & 0.552 & 0.521 & 1.036 & 0.783   & 0.70\%\\
				& 720 & \textbf{0.436} & \textbf{0.452} & \underline{0.440} & \underline{0.453} & 0.504 & 0.515 & 0.584 & 0.561 & 0.822 & 0.700 & 0.474 & 0.491 & 0.500 & 0.501 & 0.655 & 0.604 & 1.174 & 0.856 & 0.91\%\\
				\midrule
				\multirow{4}{*}{\rotatebox{90}{ETTh2}} & 96  & \underline{0.278} & \textbf{0.335} & \textbf{0.277} & \underline{0.338} & 0.290 & 0.353 & 0.397 & 0.434 & 0.292 & 0.355 & 0.339 & 0.381 & 0.397 & 0.430 & 0.843 & 0.669 & 2.978 & 1.360 & --  \\
				& 192 & \textbf{0.344} & \textbf{0.380} & \textbf{0.344} & \underline{0.381} & 0.388 & 0.422 & 0.594 & 0.548 & 0.441 & 0.454 & 0.429 & 0.437 & 0.440 & 0.441 & 0.472 & 0.492 & 6.203 & 2.078 &  --   \\
				& 336 & \underline{0.362} & \textbf{0.400} & \textbf{0.357} & \textbf{0.400} & 0.463 & 0.473 & 0.615 & 0.559 & 0.545 & 0.515 & \underline{0.445} & 0.461 & 0.477 & 0.481 & 0.898 & 0.687 & 5.437 & 1.961 & --  \\
				& 720 & \textbf{0.393} & \textbf{0.434} & \underline{0.394} & \underline{0.436} & 0.733 & 0.606 & 1.079 & 0.764 & 0.834 & 0.688 & 0.455 & \underline{0.475} & 0.482 & 0.489 & 1.250 & 0.830 & 4.115 & 1.692 & 0.25\% \\
				\midrule
				\multirow{4}{*}{\rotatebox{90}{ETTm1}} & 96  & \textbf{0.287} & \textbf{0.335} & 0.306 & 0.348 & \underline{0.301} & \underline{0.345} & 0.339 & 0.386 & 0.315 & 0.365 & 0.349 & 0.401 & 0.520 & 0.487 & 0.392 & 0.425 & 0.624 & 0.558  & 4.65\%  \\
				& 192 & \textbf{0.330} & \textbf{0.360} & 0.349 & 0.375 & \underline{0.336} & \underline{0.366} & 0.381 & 0.413 & 0.361 & 0.388 & 0.390 & 0.423 & 0.543 & 0.498 & 0.472 & 0.492 & 0.725 & 0.618 & 1.79\% \\
				& 336 & \textbf{0.369} & \textbf{0.382} & 0.375 & \underline{0.388} & \underline{0.372} & 0.389 & 0.414 & 0.436 & 0.387 & 0.416 & 0.433 & 0.450 & 0.652 & 0.543 & 0.527 & 0.525 & 1.006 & 0.750  & 0.81\% \\
				& 720 & \textbf{0.426} & \textbf{0.414} & 0.433 & \underline{0.422} & \underline{0.427} & 0.423 & 0.475 & 0.470 & 0.445 & 0.454 & 0.480 & 0.474 & 0.707 & 0.570 & 0.608 & 0.564 & 0.967 & 0.741 & 0.23\% \\
				\midrule
				\multirow{4}{*}{\rotatebox{90}{ETTm2}} & 96  & \textbf{0.162} & \textbf{0.250} & \underline{0.167} & \underline{0.255} & 0.172 & 0.267 & 0.196 & 0.294 & 0.178 & 0.272 & 0.189 & 0.280 & 0.254 & 0.321 & 0.360 & 0.426 & 0.382 & 0.463 & 2.99\% \\		
				& 192 & \textbf{0.217} & \textbf{0.288} & \underline{0.221} & \underline{0.293} & 0.237 & 0.314 & 0.369 & 0.424 & 0.236 & 0.310 & 0.255 & 0.322 & 0.273 & 0.331 & 0.580 & 0.568 & 0.849 & 0.724 & 1.81\%  \\
				& 336 & \textbf{0.273} & \textbf{0.325} & \underline{0.275} & \underline{0.327} & 0.295 & 0.359 & 0.410 & 0.447 & 0.299 & 0.351 & 0.323 & 0.363 & 0.340 & 0.371 & 1.623 & 0.799 & 1.423 & 0.915 & 0.73\% \\
				& 720 & \textbf{0.368} & \textbf{0.383} & \underline{0.370} & \underline{0.385} & 0.427 & 0.439 & 0.583 & 0.535 & 0.435 & 0.452 & 0.421 & 0.419 & 0.453 & 0.439 &1.954 & 1.015 & 3.929 & 1.469 & 0.54\% \\
				\bottomrule
			\end{tabular}
		}
	\begin{tablenotes}
		\tiny
		\item Results$^\ast$ are from SCINet \cite{liu2022scinet} due to out-of-memory. Other results are implemented by us. 
	\end{tablenotes}
	\end{threeparttable}
\end{table}

\paragraph{Multivariate results} For multivariate long-term forecasting, our proposed MPPN outperforms all baseline models and achieves the state-of-the-art performance on most of the datasets (Table \ref{multivariate-table}). Specifically, under the prediction length setting of 96, compared to the previous state-of-the-art results, MPPN lowers MSE by \textbf{15.29\%} (0.232\textrightarrow0.144) in Weather, \textbf{6.07\%} (0.412\textrightarrow0.387) in Traffic, \textbf{6.43\%} (0.140\textrightarrow0.131) in Electricity, \textbf{4.65\%} (0.301\textrightarrow0.287) in ETTm1 and \textbf{2.99\%} (0.167\textrightarrow0.162) in ETTm2. For ILI dataset of predicted length 24, MPPN gives \textbf{21.23\%} (2.280\textrightarrow1.796) MSE reduction. 
For datasets with evident periodicity, such as Electricity, Weather, and Traffic (Table \ref{table_pre_per}), our MPPN shows stronger capabilities in capturing and modeling their inherent time series patterns compared to other models. As for data without clear periodic patterns, like the Exchange-Rate and ETTh2, MPPN still provides commendable predictions. 
It is worth noting that MPPN exhibits significant improvements under the predict-96 setting and maintains stable performance for longer prediction horizons. This impressive pattern mining aptitude of MPPN for LTSF tasks makes it a practical choice for real-world applications. 
Besides, the overall sampling-based and decomposition-based methods perform better than other baseline models, highlighting the importance of capturing specific patterns in time series data.
We also list the univariate long-term forecasting results in Appendix B.

\subsection{Ablation studies}

\begin{table}[htbp]
\centering
\caption{Ablation studies: multivariate long-term series prediction results on Weather and Electricity with input length $720$ and prediction length in $\{96,192,336,720\}$. Three variants of MPPN are evaluated, with the best results highlighted in bold.}
\label{ablation}

\resizebox{\linewidth}{!}{
\begin{tabular}{cccccccccccc}
\toprule
\multicolumn{2}{c}{Methods}&\multicolumn{2}{c}{MPPN}&\multicolumn{2}{c}{w/o multi-resolution}&\multicolumn{2}{c}{w/o periodic sampling}&\multicolumn{2}{c}{w/o channel adaption}\\
\midrule
\multicolumn{2}{c}{Metric} & MSE  & MAE & MSE & MAE& MSE  & MAE & MSE  & MAE\\
\midrule
\multirow{4}{*}{\rotatebox{90}{Weather}} 

& 96 & \textbf{0.144} & \textbf{0.196} & 0.165 & 0.226 & 0.147 & 0.200 & 0.167 & 0.222  \\

& 192 & \textbf{0.189} & \textbf{0.240} & 0.209 & 0.261 & 0.196 & 0.249 & 0.212 &  
0.259  \\

& 336 & \textbf{0.240} & \textbf{0.281} & 0.258 & 0.302 & 0.246 & 0.289 & 0.258 & 0.295 \\

& 720 & \textbf{0.310} & \textbf{0.333} & 0.313 & 0.336  & 0.312 & 0.335 & 0.322 & 0.341 \\
\midrule
\multirow{4}{*}{\rotatebox{90}{Electricity}} 

& 96 & \textbf{0.131} & \textbf{0.226} & 0.156 & 0.264 & 0.133 & 0.228 & 0.133  & 0.228    \\

& 192 & \textbf{0.145} & \textbf{0.239} & 0.171 & 0.276 & 0.148 & 0.242 & 0.147  &   0.241  \\

& 336 & \textbf{0.162} & \textbf{0.256} & 0.186 & 0.290 & 0.164 & 0.258 & 0.164  & 0.258\\

& 720 & \textbf{0.200} & \textbf{0.289} &0.223  & 0.319 & 0.203 & 0.292 &0.203  &0.291  \\
\bottomrule
\end{tabular}
}
\end{table}

In this section, we conduct ablation studies on Weather and Electricity to assess the effect of each module in MPPN. Three variants of MPPN are evaluated: 1) \textbf{w/o multi-resolution}: we remove the multi-resolution patching and instead employ a single resolution and a single period for sampling; 2) \textbf{w/o periodic sampling}: we eliminate periodic pattern mining and directly adopt multi-resolution patching followed by a channel adaptive module and an output layer; 3) \textbf{w/o channel adaption}: we drop channel adaptive module and treat each channel equally; The experimental results are summarized in Table \ref{ablation} with best results bolded. As can be seen from Table \ref{ablation},  omitting multi-resolution or periodic pattern mining leads to significant performance degradation. As discussed in Section \ref{method}, multivariate time series data often exhibit complex multi-period patterns and dependencies. Taking into account the contextual information surrounding each data point in the time series helps compensate for the information scarcity of individual points. Therefore, employing multi-resolution patching and multiple periodic pattern mining facilitates better exploration of the intrinsic patterns in times series. Channel adaption also brings noticeable performance improvement for both datasets. Compared to Electricity containing only electricity consumption data, the impact of channel adaption is more pronounced on Weather. Since Weather dataset comprises distinct meteorological indicators, such as wind velocity and air temperature, it is conducive to regard different channels distinctively rather than treating them equally. Overall, MPPN enjoys the best performance across different datasets and prediction horizons, which demonstrates the effectiveness of each modeling mechanism. The collaborative integration of different modules facilitates our model in achieving SOTA performance across different datasets. Further ablation experiment results can be found in the Appendix.

\subsection{Model analysis}

\begin{figure*}[h]
	\begin{center}
		\centerline{\includegraphics[width=\columnwidth]{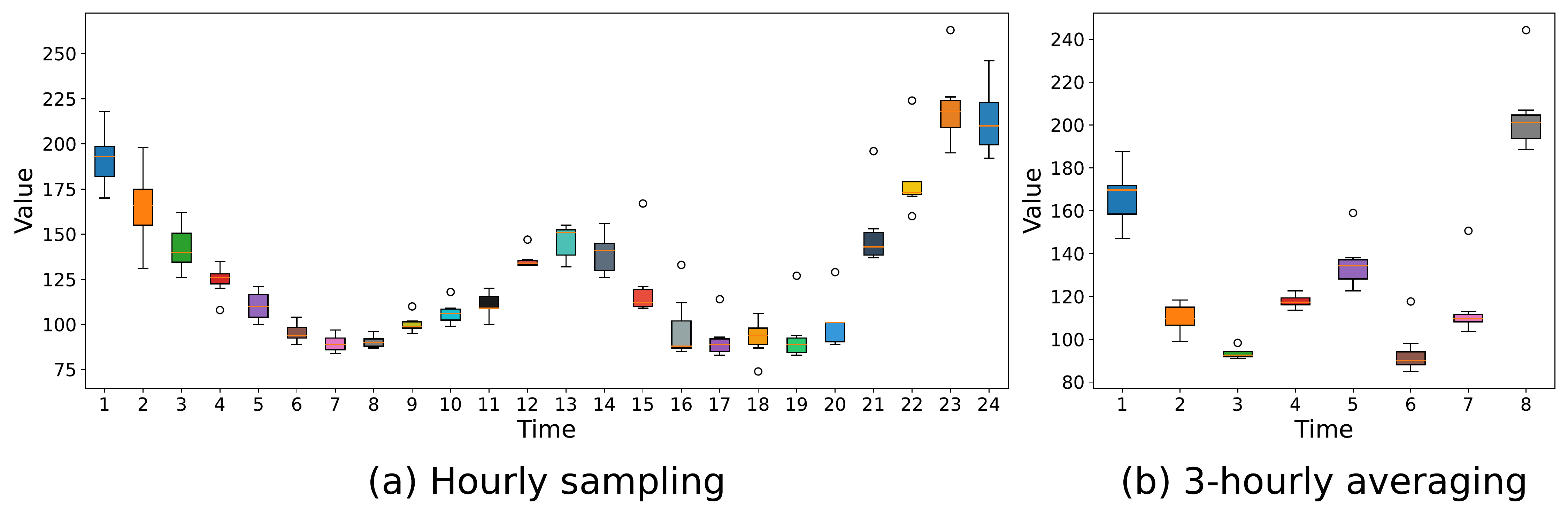}}
		\caption{Period pattern analysis on the Electricity dataset. }
		\label{fig-period}
	\end{center}
\end{figure*}

\paragraph{Periodic pattern}
As shown in Figure \ref{fig-period}(a), we randomly select a variate from the Electricity dataset with hourly interval and sample its historical data over 7 days. We find that the data at the same time point for each day exhibits fluctuations within a relatively small range, while the magnitude of the fluctuations varies at different time points. 
Our findings confirm the existence of periodic patterns in the analysed time series, demonstrating that our proposed MPPM in Section \ref{method} which can extract these patterns could improve the performance. Meanwhile, we also investigate the patterns of three-hour resolution by taking the mean value of the adjacent three time points, as shown in Figure \ref{fig-period}(b). 
Time series data exhibits periodic patterns across different resolutions, thus integrating multi-resolution patterns of the series can enhance modeling accuracy. 
We also investigate the effectiveness of multi-resolution patching with multiple periodicity, which actually can further improve the performance.

\paragraph{Channel adaptive modeling}



\begin{figure}[htbp]
  \centering

  \begin{minipage}[b]{0.45\textwidth}
    \centering
    \includegraphics[width=\textwidth]{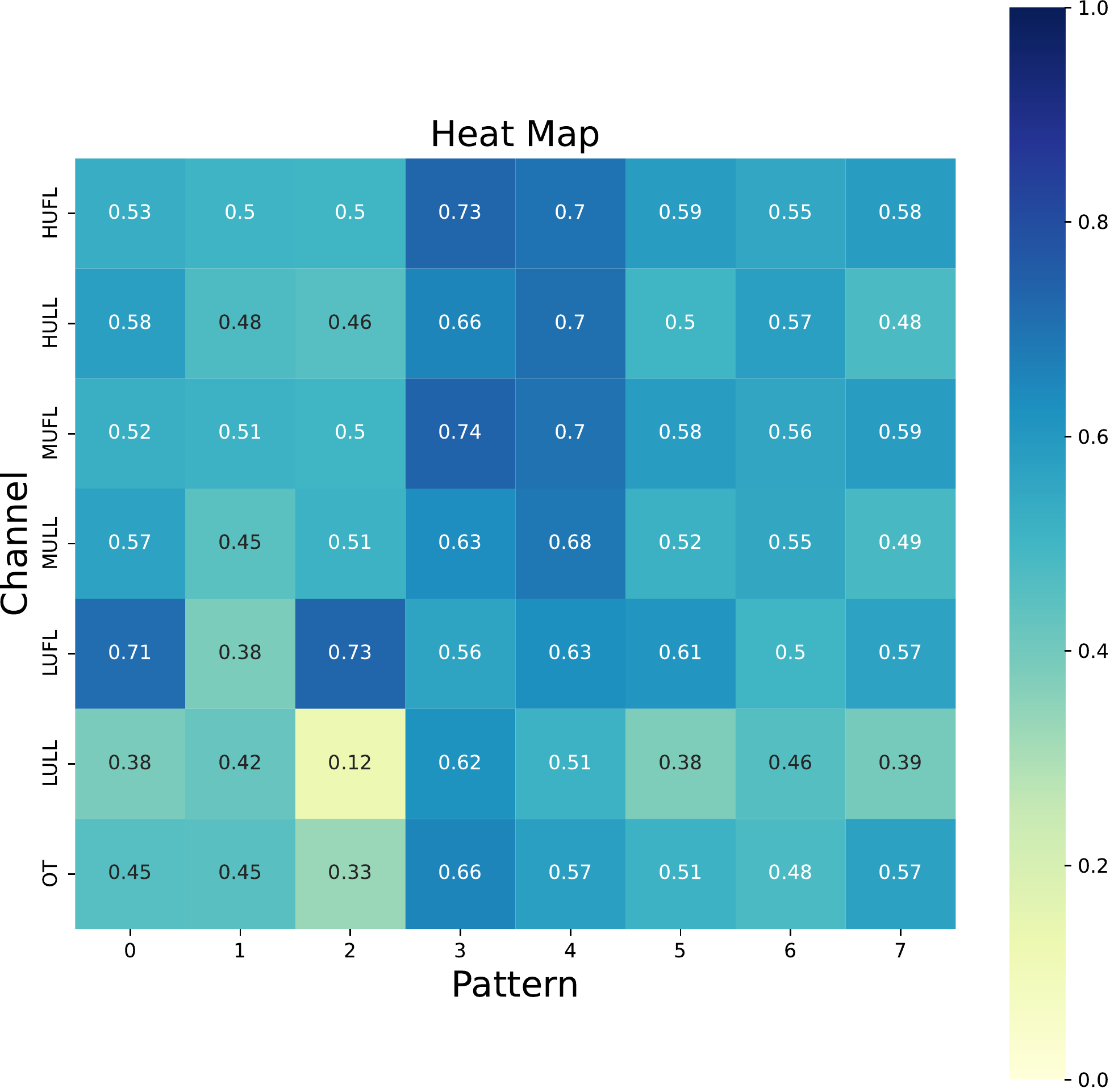}
    \caption{Heat map of channel adaption on ETTh1 with eight extracted patterns.}
    \label{fig-node}
  \end{minipage}
  \hfill 
  \begin{minipage}[b]{0.46\textwidth}
    \centering
    \includegraphics[width=\textwidth]{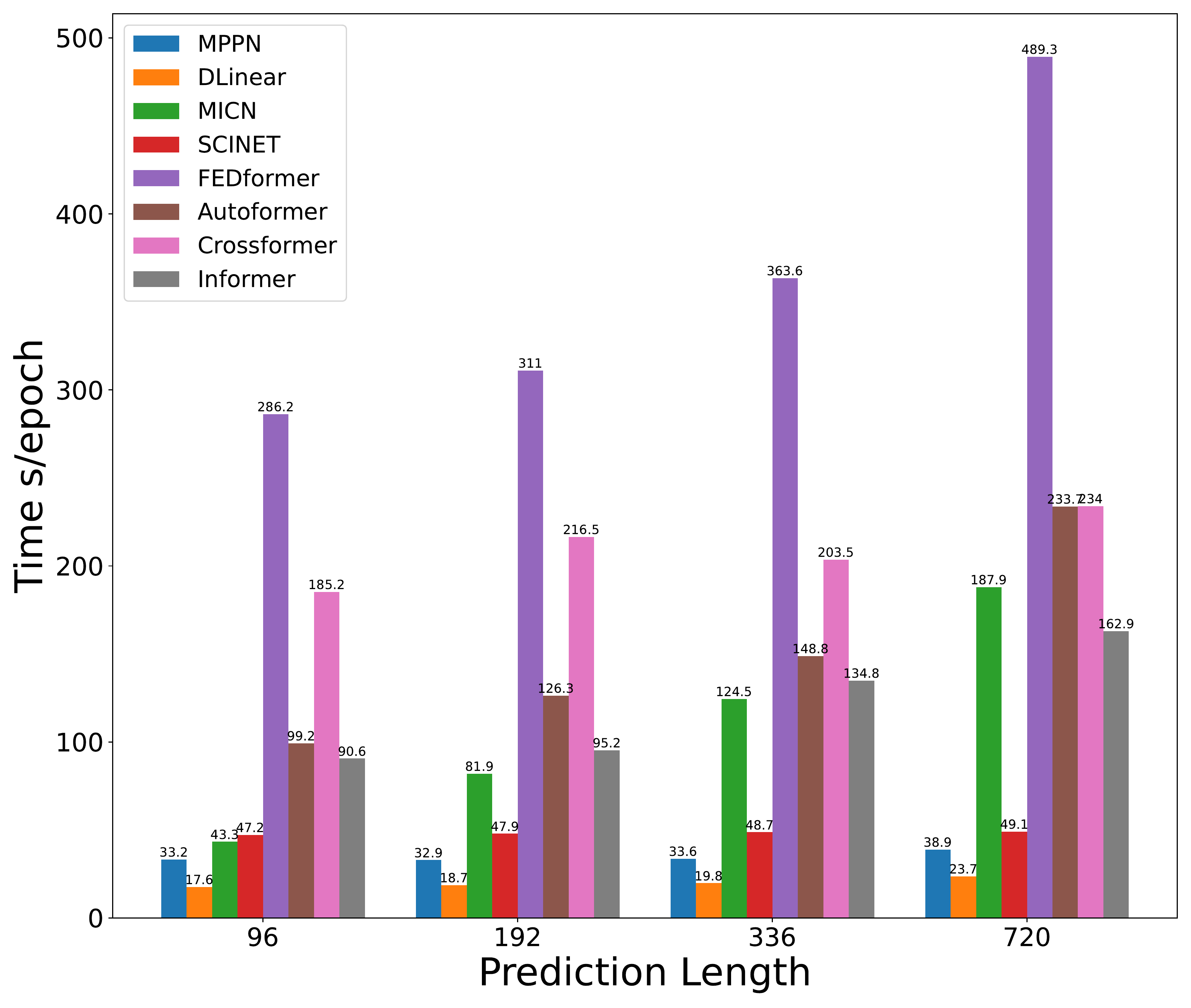}
    \caption{Comparison of the training time for different baseline models and our MPPN.}
    \label{fig-efficiency}
  \end{minipage}

\end{figure}

To illustrate the effect of the channel adaptive module, we visualize the channel embedding matrix on \text{ETTh1} dataset with eight patterns. We set the look-back window $L=336$ and the prediction horizon to be $96$. In Figure \ref{fig-node}, the varying hues and numbers in each block represent the sensitivity of various channels to distinct temporal patterns. It can be seen that most variates (channels) are significantly influenced by the third and fourth patterns, with the exception of `LULF', which denotes \textbf{L}ow \textbf{U}se\textbf{F}ul \textbf{L}oad. The visualization results reveal that different channels in multivariate time series are often affected by distinct temporal patterns. Therefore, it is crucial to differentiate and treat them individually according to their characteristics. The channel adaptive module in MPPN helps capture the perceptions of multivariate towards different patterns, while also providing interpretability to our approach.

\paragraph{Efficiency analysis} We compare the training time for one epoch of our MPPN with serval baseline models on the Weather dataset, and the results are shown in Figure \ref{fig-efficiency}. In general, Transformer-based models exhibit the highest time complexity, followed by CNN-based models. While MPPN demonstrates sightly higher time complexity compared to the single-layer DLinear, the difference is not significant under the premise of better prediction accuracy. As the prediction length increases, the training time of certain models, such as MICN, FEDformer, and Autoformer, shows a noticeable growth. Meanwhile, models like SCINet, Crossformer, and Informer do not show a significant increase as the prediction length grows, but they still have considerably higher training time compared to MPPN. 
Our MPPN model exhibits superior efficiency in handling long-term time series forecasting.



\section{Conclusion}
In this paper, we propose a novel deep learning network architecture MPPN for long-term time series forecasting. We construct multi-resolution contextual-aware semantic units of time series and propose the multi-period pattern mining mechanism to explicitly capture key time series patterns.
Furthermore, we propose a channel-adaptive module to model each variate's perception of different extracted patterns for multivariate series prediction. Additionally, we present an entropy-based method for evaluating the predictability and providing an upper bound on the  prediction accuracy before carrying out predictions. 
Extensive experiments on nine real-world datasets demonstrate the superiority of our method in long-term forecasting tasks compared to state-of-the-art methods.
For the future works, we will further explore the general patterns in time series data and build task-general models to support multiple downstream time series analysis tasks.

\medskip  

\bibliographystyle{plain}
\bibliography{neurips_2023.bib}

\newpage

\appendix
\section*{Appendix}

\section{Experimental details}

\subsection{Data descriptions}
\label{Data-descriptions}
We conduct extensive experiments on nine widely-used time series datasets. (1) \emph{ETT} \cite{zhou2021informer} contains data collected from electricity transformers located in two regions of China, between July 2016 and July 2018. The dataset includes two different granularities: ETTh for 1 hour and ETTm for 15 minutes. Each data point consists an `oil temperature' value and six external power load features. (2) \emph{Electricity}\footnote{\url{https://archive.ics.uci.edu/ml/datasets/ElectricityLoadDiagrams20112014}} contains electricity consumption data for 321 clients in kilowatts (kW) of 1 hour from 2012 to 2014, which is released from the UCL Machine Learning Repository. (3) \emph{Exchange-Rate} \cite{lai2018modeling} records the daily exchange rates of eight different countries, spanning from 1990 to 2016. (4) \emph{Traffic}\footnote{\url{http://pems.dot.ca.gov/}} contains hourly road occupancy rates recorded by the real-time individual detectors located on San Francisco Bay area freeways between 2015 to 2016. (5) \emph{Weather}\footnote{\url{https://www.bgc-jena.mpg.de/wetter/}} contains 21 meteorological indicators recorded every 10 minutes in Germany during the year 2020, including air temperature, wind velocity, etc. (6) \emph{ILI}\footnote{\url{https://gis.cdc.gov/grasp/fluview/fluportaldashboard.html}} consists of the ratio of patients seen with influenza-like illness to the total number of patients, which is collected by the United States Centers for Disease Control and Prevention on a weekly basis from 2002 to 2021. Following the standard protocol outlined in \cite{wu2021autoformer}, we partition all datasets into training, validation, and testing sets chronologically, with a ratio of 6:2:2 for the ETT dataset and 7:1:2 for other datasets.

\subsection{Evaluation metrics}
In this paper, we perform zero-mean normalization to the data and select the Mean Absolute Error (MAE) and the Mean Squared Error (MSE) as evaluation metrics: 
\begin{equation}
\text{MAE}=\frac{1}{H}\sum_{k=0}^H ||\boldsymbol{x}_{t+k}-\hat{\boldsymbol{x}}_{t+k}||,
\end{equation}
\begin{equation}
\text{MSE}=\frac{1}{H}\sum_{k=0}^H ||\boldsymbol{x}_{t+k}-\hat{\boldsymbol{x}}_{t+k}||^2,
\end{equation}
where $H$ is the output length, $\hat{\boldsymbol{x}}_{t+k}$ is the prediction of the model, and $\boldsymbol{x}_{t+k}$ is the ground truth.

\subsection{Baseline models}
\label{Baseline-models}
We compare MPPN with a variety of state-of-the-art models in the domain of multivariate long-term time series forecasting, including Linear-based models, CNN-based models and Transformer-based models. We summarize them as follows.
\begin{itemize}
\item NLinear \cite{Zeng2022AreTE}: NLinear performs a simple normalization, where the input is first subtracted by the last value of the sequence, then goes through a single-layer linear model. NLinear demonstrates an effective capability to address distribution shifts within datasets.
\item DLinear \cite{Zeng2022AreTE}: DLinear first decomposes the raw time series into a trend component by moving average and a seasonal component. Then it adopts two one-layer linear layers to each component and sum up the two features to produce the final prediction.
\item MICN \cite{wang2023micn}: MICN initially decomposes the input series into trend-cyclical and seasonal parts. It then uses down-sampled convolution and isometric convolution to capture both the local and global correlations presented in time series.
\item SCINet \cite{liu2022scinet}: SCINet is a hierarchical convolution network with downsample-convolve-interact operations. The input sequence is downsampled into multiple resolution sub-sequences, which are then iteratively extracted and exchanged to facilitate the learning of an effective representation for the time series.
\item FEDformer \cite{zhou2022fedformer}: FEDformer incorporates a set of decomposition blocks with different sizes to effectively capture the global profile of the time series. Additionally, it performs the attention mechanism with low-rank approximation in the frequency domain with Fourier transform to enhance the performance for long-term prediction.
\item Autoformer \cite{wu2021autoformer}: Autoformer is a deep decomposition architecture that incorporates the built-in decomposition block and the series-wise attention based on auto-correlation mechanism. The decomposition block is designed to progressively isolate the long-term trend information from the predicted hidden variables.
\item Crossformer \cite{zhang2023crossformer}: Crossformer embeds the input series into a 2D vector with time and dimension features, and two attention layers are proposed to capture the cross-time and cross-dimension dependencies, respectively.
\item Informer \cite{zhou2021informer}: Informer introduces the ProbSparse self-attention mechanism and self-attention distilling operation to address the problems of quadratic time complexity and quadratic memory usage in vanilla Transformer. 
\item LogTrans \cite{li2019enhancing}: LogTrans introduces convolutional self-attention to involve local context and LogSparse Transformer to reduce space complexity in vanilla Transformer models.
\end{itemize}

\subsection{Hyperparameter}
\label{Implementation-details}
By default, MPPN employs multi-resolution patching (non-overlapping patches) and the multiple temporal resolutions (convolutional kernel sizes) are set as [1, 3, 4, 6]. 
Referring to the results in Table \ref{table_pre_per}, for datasets with obvious periodicity, we select the top-2 period ($k=2$) computed by the method of FFT as discussed in Section \ref{method} for periodic pattern mining, including Electricity, Weather, Traffic, ILI, ETTh1 and ETTm1.
For datasets with less obvious periodicity, we set the period to the default value of 24 in the case of Exchange-Rate, ILI, and ETTh2. 
For ETTm2 with a granularity of 15 minutes, we set the period to 96, which corresponds exactly to the default daily period.
At last, we set the hyperparameter $D=48$ which is the hidden state of the series for all datasets.

For MPPN, we employ different input length on different datasets. Specifically, experiments conducted on the Weather, Traffic, and Electricity dataset employ an input length of 720. For the Exchange-Rate and ETT dataset,  we opt an input length of 336. Lastly, for the ILI dataset, we employ the input length as 96. We adopt the default hyperparameters of each baseline model mentioned in their papers to train them. Increasing the length of the input length offers substantial performance benefits for Linear-based models, while Transformer-based models tend to overfit temporal noises rather than extracting temporal information \cite{Zeng2022AreTE}. We set the input length $L=96$ for Transformer-based, and $L=336$ for CNN-based models and Linear-based models. Since the ILI dataset sampled at weekly granularity is smaller in size, we use a different set of parameters ($L=36$ for Transformer-based models and CNN-based models, and $L=104$ for Linear-based models).

\section{Univariate forecasting}
\label{Univariate-forecasting}
 We present the univariate time-series forecasting results in Table \ref{univariate-table}.
 Following the results from MICN \cite{wang2023micn}, we include LogTrans \cite{li2019enhancing} in comparison.
 Our MPPN model continues to demonstrate superior performance in comparison to the baseline models, particularly in the context of the weather dataset with obvious periodicity. MPPN achieves significant reductions in MSE, with decreases of \textbf{62.07\%}(0.0029\textrightarrow0.0011), \textbf{33.33\%}(0.0021\textrightarrow0.0014), \textbf{30.43\%}(0.0023\textrightarrow0.0016), and \textbf{32.26\%}(0.0031\textrightarrow0.0021) observed for forecast horizons of 96, 192, 336, and 720, respectively. The experimental results provide evidence for the effectiveness of MPPN in extracting essential characteristics of time series and performing univariate forecasting.

\begin{table}[tbp]
	\caption{\textbf{Univariate} long-term time series forecasting results with different prediction length $O\in\{24, 36, 48, 60\}$ for ILI dataset and $O\in\{96, 192, 336, 720\}$ for others. The SOTA results are \textbf{bolded}, while the sub-optimal results are \underline{underlined}. Correspondingly, IMP. shows the percentage improvement of MPPN over the state-of-the-art baseline models.}
	\label{univariate-table}
	\centering
	\begin{threeparttable}
		\renewcommand{\multirowsetup}{\centering}
		\resizebox{\linewidth}{!}{
			\begin{tabular}{lcllllllllllllllc}
				\toprule
				\multicolumn{2}{c}{Models} & \multicolumn{2}{c}{\textbf{MPPN}} &  \multicolumn{2}{c}{DLinear} &  \multicolumn{2}{c}{MICN$^\ast$} & \multicolumn{2}{c}{FEDformer$^\ast$} & \multicolumn{2}{c}{Autoformer$^\ast$} & \multicolumn{2}{c}{Informer$^\ast$} & \multicolumn{2}{c}{LogTrans$^\ast$} & {IMP.}\\
				\cmidrule(lr){3-4} \cmidrule(lr){5-6} \cmidrule(lr){7-8} \cmidrule(lr){9-10} \cmidrule(lr){11-12} \cmidrule(lr){13-14} \cmidrule(lr){15-16} \cmidrule(lr){17-17} 
				\multicolumn{2}{c}{Metric} &MSE & MAE & MSE & MAE & MSE & MAE & MSE & MAE & MSE & MAE & MSE & MAE & MSE & MAE & MSE \\
				\toprule
				\multirow{4}{*}{\rotatebox{90}{Weather}} & 96  & \textbf{0.0011} & \textbf{0.025} &  0.0057 & 0.063 & \underline{0.0029} & \underline{0.039} & 0.0062 & 0.062 & 0.011 & 0.081 & 0.0038 & 0.044 & 0.0046 & 0.052 &  62.07\%  \\
				& 192 & \textbf{0.0014} & \textbf{0.027} & 0.0062 & 0.066 & \underline{0.0021} & \underline{0.034} & 0.0060 & 0.062 & 0.0075 & 0.067 & 0.0023 & 0.040 & 0.0056 & 0.060 & 33.33\% \\
				& 336 & \textbf{0.0016} & \textbf{0.030} & 0.0064 & 0.068 & \underline{0.0023} & \underline{0.034} & 0.0041 & 0.050  & 0.0063 & 0.062 & 0.0041 & 0.049 & 0.0060 & 0.054 & 30.43\% \\
				& 720 & \textbf{0.0021} & \textbf{0.034}  & 0.0068 & 0.070 & 0.0048 & 0.054 & 0.0055 & 0.059 & 0.0085 & 0.070 & \underline{0.0031} & \underline{0.042} & 0.0071 & 0.063 & 32.26\% \\
				\midrule
				\multirow{4}{*}{\rotatebox{90}{Traffic}} & 96 & \textbf{0.115} & \textbf{0.191} & \underline{0.126} & \underline{0.202} & 0.158 & 0.241 & 0.207 & 0.312 & 0.246 & 0.346 & 0.257 & 0.353 & 0.226 & 0.317 & 8.73\% \\
				& 192 & \textbf{0.118} & \textbf{0.204} & \underline{0.129} & \underline{0.208} & 0.154 & 0.236 & 0.205 & 0.312 & 0.266 & 0.370 & 0.299 & 0.376 & 0.314 & 0.408 &  8.53\% \\
				& 336 & \textbf{0.118} & \textbf{0.200} & \underline{0.130} & \underline{0.213} & 0.165 & 0.243 & 0.219 & 0.323 & 0.263 & 0.371 & 0.312 & 0.387 & 0.387 & 0.453 & 9.23\% \\
				& 720 & \underline{0.146} & \underline{0.237} & \textbf{0.142} & \textbf{0.226} & 0.182 & 0.264 & 0.244 & 0.344 & 0.269 & 0.372 & 0.366 & 0.436 & 0.491 & 0.437 & \textbackslash \\
				\midrule
				\multirow{4}{*}{\rotatebox{90}{Electricity}} & 96  & \textbf{0.195} & \textbf{0.304} & \underline{0.203} & \underline{0.315} & 0.310 & 0.398 & 0.253 & 0.370 & 0.341 & 0.438 & 0.484 & 0.538 & 0.288 & 0.393 & 3.94\% \\
				& 192 & \textbf{0.230} & \textbf{0.329} & \underline{0.233} & \underline{0.337} & 0.300 & 0.394 & 0.282 & 0.386  & 0.345 & 0.428 & 0.557 & 0.558 & 0.432 & 0.483 &  1.29\% \\
				& 336 & \textbf{0.266} & \textbf{0.360} & \underline{0.268} & \underline{0.364} & 0.323 & 0.413 & 0.346 & 0.431  & 0.406 & 0.470 & 0.636 & 0.613 & 0.430 & 0.483 & 0.75\% \\
				& 720 & \underline{0.316} & \underline{0.413} &  \textbf{0.307} & \textbf{0.408} & 0.364 & 0.449 & 0.422 & 0.484 & 0.565 & 0.581 & 0.819 & 0.682 & 0.491 & 0.531 & \textbackslash \\
				\midrule
				\multirow{4}{*}{\rotatebox{90}{Exchange}} & 96 & \textbf{0.095}  & \textbf{0.232} & 0.155 & 0.288 & \underline{0.099} & \underline{0.240} & 0.154 & 0.304 & 0.241 & 0.387 & 0.591 & 0.615 & 0.237 & 0.377 & 4.04\% \\
				& 192 & 0.204 & \textbf{0.339} & \textbf{0.194} & \underline{0.351} & \underline{0.198} & 0.354 & 0.286 & 0.420 & 0.300 & 0.369 & 1.183 & 0.912 & 0.738 & 0.619 & \textbackslash \\
				& 336 & 0.431 & \underline{0.492} & \underline{0.420} & 0.508 & \textbf{0.302} & \textbf{0.447} & 0.511 & 0.555  & 0.509 & 0.524 & 1.367 & 0.984 & 2.018 & 1.070 & \textbackslash \\
				& 720 & 1.274 & 0.858 & \underline{0.814} & \underline{0.727} & \textbf{0.738} & \textbf{0.662} & 1.301 & 0.879 & 1.260 & 0.867 & 1.872 & 1.072 & 2.405 & 1.175 &  \textbackslash \\
				\midrule
				\multirow{4}{*}{\rotatebox{90}{ILI}} & 24  & \textbf{0.630} & \textbf{0.627} &  0.726 & 0.681 & \underline{0.674} & \underline{0.671} & 0.708 & \textbf{0.627} & 0.948 & 0.732 & 5.282 & 2.050 & 3.607 & 1.662 & 6.53\% \\
				& 36 & \textbf{0.517} & \textbf{0.573} &  0.793 & 0.745 & 0.712 & 0.733 & \underline{0.584} & \underline{0.617} & 0.634 & 0.650 & 4.554 & 1.916 & 2.407 & 1.363 & 11.47\% \\
				& 48 & \textbf{0.621} & \textbf{0.640} & 0.886 & 0.815 & 0.823 & 0.803 & \underline{0.717} & \underline{0.697} & 0.791 & 0.752 & 4.273 & 1.846 & 3.106 & 1.575 & 13.39\%\\
				& 60 & \textbf{0.695} & \textbf{0.688} & 0.960 & 0.860 & 0.992 & 0.892 & \underline{0.855} & \underline{0.774} & 0.874 & 0.797 & 5.214 & 2.057 & 3.698 & 1.733 & 18.71\% \\
				\bottomrule
			\end{tabular}
		}
	\begin{tablenotes}
		\tiny
		\item Results$^\ast$ are from MICN\cite{wang2023micn}; Other results are implemented by us.
	\end{tablenotes}
	\end{threeparttable}
\end{table}

\section{Error bars evaluation}
To evaluate the robustness of MPPN across different settings, we conduct experiments based on three independent runs on ETTh1, Weather and Electricity. As depicted in Table \ref{error-bar}, the standard deviation (Std.) is basically no more than 3\% of the mean values (Mean). These results highlight the robustness of MPPN in the face of diverse initialization settings.

\begin{table}[htbp]
	\caption{The error bars of MPPN with 3 runs, input length $I=720$ and output length $O\in\{96, 192, 336, 720\}$ on ETTh1, Weather and Electricity.}
	\label{error-bar}
	\centering
 	\resizebox{\linewidth}{!}{
	\begin{threeparttable}
		\begin{tabular}{llllllllllllllllll}
			\toprule
			\multicolumn{2}{c}{Dataset} & \multicolumn{5}{c}{Weather} &  \multicolumn{5}{c}{Electricity} & \multicolumn{5}{c}{ETTh1}  \\			
			\cmidrule(lr){3-7} \cmidrule(lr){8-12} \cmidrule(lr){13-17}
     \multicolumn{2}{c}{Metric} & Seed1 & Seed2 & Seed3 &  Mean & Std. & Seed1 & Seed2 & Seed3 & Mean & Std. & Seed1 & Seed2 & Seed3 & Mean & Std. \\
	\toprule
      \multirow{2}{*}{$O$=96}      
      & MSE & 0.1441 & 0.1459 & 0.1479 & 0.1460 & 0.0016 & 0.1311 & 0.1310 & 0.1312& 0.1311 & 0.0001 & 0.3715&0.3773 &0.3751 & 0.3746 & 0.0024\\
      & MAE & 0.1969 & 0.1986 & 0.2030 & 0.1995 & 0.0026 & 0.2263 & 0.2262 & 0.2263 & 0.2263 & 0.0001 &0.3931 & 0.3993 & 0.3974 & 0.3966 & 0.0026 \\
      \midrule
      \multirow{2}{*}{$O$=192}      
      & MSE & 0.1893 & 0.1899 & 0.1898 & 0.1897 & 0.0002 & 0.1460 &0.1459 &0.1460 & 0.1460& 0.0001 & 0.4055 &0.4117 &0.4118 & 0.4097 & 0.0029 \\
      & MAE & 0.2404 & 0.2409 & 0.2418 & 0.2410 & 0.0006 &0.2398 & 0.2399& 0.2399& 0.2399& 0.0000 & 0.4135 & 0.4201 & 0.4208 & 0.4181 & 0.0033  \\
    \midrule
    \multirow{2}{*}{$O$=336}      
      & MSE & 0.2401 & 0.2406 & 0.2396 & 0.2401 & 0.0004 &0.1621 &0.1621 &0.1618 &0.1620 & 0.0001&0.4269 &0.4505 & 0.4321 & 0.4365 & 0.0101 \\
      & MAE & 0.2814 & 0.2817 & 0.2808 & 0.2813 & 0.0004  &0.2565 & 0.2567 & 0.2566 & 0.2566 & 0.0001 & 0.4257 & 0.4492 & 0.4327 & 0.4359 & 0.0099 \\
    \midrule
    \multirow{2}{*}{$O$=720}      
      & MSE & 0.3104 & 0.3105 & 0.3106 & 0.3105 & 0.0001 & 0.2005 & 0.2006&0.2007 &0.2006 & 0.0001& 0.4366&0.4518 & 0.4455 & 0.4446 & 0.0062\\
      & MAE & 0.3332 & 0.3330 & 0.3346 & 0.3336 & 0.0007 &0.2898 &0.2898 & 0.2902& 0.2900& 0.0002& 0.4528 & 0.4619 & 0.4571 & 0.4573 & 0.0037  \\

		\bottomrule
		\end{tabular}
	\end{threeparttable}
	}
\end{table}

\section{Hyperparameter sensitivity} 

\subsection{Input length selection}  
In this section, we evaluate the impact of input length on model performance, as presented in Table \ref{input-length-table}. The experimental results indicate that the impact of input length on model performance varies across different datasets. Due to the differences in granularity and periodicity among datasets, fine-tuning the input length for each specific dataset is preferable over adopting a one-size-fits-all approach in real-world applications. For example, in datasets with simple patterns like ETTm1, an input length of 336 is sufficient to capture  substantial information. But for datasets characterized by intricate time series patterns, such as Electricity and Weather, a longer input length is required to extract more comprehensive information that can effectively contribute to the prediction task. Besides, the performances of MPPN exhibit a notable growth as the size of lookback window increases, which highlights the strong temporal relation extraction capability of our method \cite{Zeng2022AreTE}.

\begin{table}[tbp]
	\caption{Multivariate long-term series results on Weather, Electricity and ETTm1 dataset with prediction length 96, and input lengths $I$ in $\{96, 192, 336, 720\}$. Four differnet variants of MPPN are evaluated, with the best results highlighted in \textbf{bold}.}
	\label{input-length-table}
	\centering
	\begin{threeparttable}
		\begin{tabular}{lllllll}
			\toprule
			Dataset & \multicolumn{2}{c}{Weather} &  \multicolumn{2}{c}{Electricity} & \multicolumn{2}{c}{ETTm1}  \\			
			\cmidrule(lr){2-3} \cmidrule(lr){4-5} \cmidrule(lr){6-7}
			Metric & MSE & MAE & MSE & MAE & MSE & MAE \\
			\toprule
			{$I$=96} & 0.174 & 0.220 & 0.189 & 0.268 & 0.339 & 0.368  \\
			\midrule
			{$I$=192} & 0.158 & 0.206 & 0.144 & 0.237 & 0.301 & 0.343  \\
			\midrule
			{$I$=336} & 0.150 & 0.201 & 0.136 & 0.231 & \textbf{0.287} & \textbf{0.335}   \\
			\midrule
			{$I$=720} & \textbf{0.144} & \textbf{0.196} & \textbf{0.131} & \textbf{0.226} &  0.304 & 0.351   \\
			\bottomrule
		\end{tabular}
	\end{threeparttable}
\end{table}

\subsection{Multiple resolution selection} 
To evaluate the impact of multiple resolutions on prediction performance, we select three sets of distinct kernel sizes. The selection of different resolutions better aligns with the typical diurnal rhythm of daily activities. For Electricity at an hourly granularity, we have chosen a convolutional filter with a size of 6 to partition a day into ``morning, afternoon, evening, and late night''. For Weather at a 10-minute granularity, convolutional filters with sizes of 3 and 6 can be employed to uncover patterns and variations in the temporal data at intervals of every half-hour or every hour. The corresponding results are presented in Table \ref{ablation-kernel}. It can be seen from Table \ref{ablation-kernel} that the first set of kernel size enjoys the best performance across different scenarios for both Weather and Electricity. As a consequence, we adopt $[1,3,4,6]$ as the default kernel sizes in our main experimental results for the nine public datasets. In practice, the selection of the optimal resolution depends on the specific application context, data granularity, and periodicity, which requires a case-by-case analysis.

\begin{table}[htbp]
\centering
\caption{Multivariate long-term series prediction results on Weather and Electricity with input length $720$ and prediction length in $\{96,192,336,720\}$. Three different combinations of convolutional kernel sizes are evaluated, with the best results highlighted in bold.}
\label{ablation-kernel}

\begin{tabular}{cccccccccccc}
\toprule
\multicolumn{2}{c}{Multiple resolutions}&\multicolumn{2}{c}{[1,3,4,6]}&\multicolumn{2}{c}{[1,3,5,7]}&\multicolumn{2}{c}{[2,4,6,8]}\\
\midrule
\multicolumn{2}{c}{Metric} & MSE  & MAE & MSE & MAE& MSE  & MAE\\
\midrule
\multirow{4}{*}{\rotatebox{90}{Weather}}

& 96 & \textbf{0.144} & \textbf{0.196} & 0.147 & 0.203 & 0.153 & 0.211 \\

& 192 & \textbf{0.189} & \textbf{0.240} & 0.190 & 0.242 & 0.198 &  0.250  \\

& 336 & \textbf{0.240} & \textbf{0.281} & 0.241 & 0.283 & 0.244 &  0.287 \\

& 720 & \textbf{0.310} & \textbf{0.333} & 0.312 & 0.335  & 0.311 & 0.335 \\
\midrule
\multirow{4}{*}{\rotatebox{90}{Electricity}} 

& 96 & \textbf{0.131} & \textbf{0.226} & 0.132 & 0.227 & 0.140 &  0.241  \\

& 192 & \textbf{0.145} & \textbf{0.239} & 0.146 & 0.241 & 0.155 & 0.255  \\

& 336 & \textbf{0.162} & \textbf{0.256} & 0.162 & 0.257 & 0.169 & 0.267 \\

& 720 & \textbf{0.200} & \textbf{0.289} & \textbf{0.200} & 0.290 & 0.210 & 0.304  \\
\bottomrule
\end{tabular}
\end{table}

\section{Ablation of patching}  
This ablation study investigates the effect of not using patches, using overlapped and non-overlapped patches to the forecasting performance. In the case of overlapped patches, we set $stride=1$ in Conv1d of multi-resolution patching. From Table \ref{patch-table}, the prediction performance of the Traffic dataset experiences a significant decline when patches are not employed. Conversely, for the Electricity dataset, the impact of using or not using patches is minimal. This observation underscores the distinctive sensitivity of datasets towards patches. As a result, the design of patching emerges as an important factor in improving forecasting performance, as well as optimizing running time and memory utilization. 
Notably, prediction results on the Electricity dataset exhibits better than baseline models even without the utilization of patches, highlighting the effectiveness of the multi-periodic pattern mining and channel adaptive module in MPPN. 
Besides, the MSE and MAE scores show minimal variations between the models utilizing overlapped and non-overlapped patches. This observation highlights the insensitivity of our model to changes in patch configuration, thus attesting to its robustness. It is advisable to flexibly select different patching methods based on the characteristics of the dataset, which enables a more tailored and effective modeling of the specific dataset.

\begin{table}[tbp]
\caption{Ablation studies: multivariate long-term series results on Electricity and Traffic with input length 720 and prediction lengths $O$ in  $\{96, 192, 336, 720\}$. Three variants of patching are evaluated, with the best results highlighted in \textbf{bold}.}
\label{patch-table}
\centering
\begin{threeparttable}
\begin{tabular}{llllllll}
	\toprule
	\multicolumn{2}{c}{Patching} & \multicolumn{2}{c}{Overlap} &  \multicolumn{2}{c}{Non-overlap} & \multicolumn{2}{c}{w/o Patch}  \\			
	\cmidrule(lr){3-4} \cmidrule(lr){5-6} \cmidrule(lr){7-8}
	\multicolumn{2}{c}{Metric} & MSE & MAE & MSE & MAE & MSE & MAE  \\
	\toprule
	\multirow{4}{*}{\rotatebox{90}{Electricity}} & 96 & \textbf{0.130} & \textbf{0.225} & 0.131 & 0.226 & 0.131 & 0.225  \\
	& 192 & 0.145 & 0.239 & 0.145 & 0.239 & 0.146 & 0.239 \\
	& 336 & 0.162 & 0.256 & 0.162 & 0.256 & 0.162 & 0.256\\
	& 720 & 0.200 & 0.289 & 0.200 & 0.289 & 0.200 & 0.289 \\
	\midrule
	\multirow{4}{*}{\rotatebox{90}{Traffic}} & 96 & \textbf{0.386} & \textbf{0.270} & 0.387 & 0.271 & 0.414 & 0.305 \\
	& 192 & 0.397 & \textbf{0.272} & \textbf{0.396} & 0.273 & 0.457 & 0.341\\
	& 336 & 0.410 & 0.280 & 0.410 & \textbf{0.279} & 0.457 & 0.336 \\
	& 720 & 0.459 & 0.315 & \textbf{0.449} & \textbf{0.301} & 0.509 & 0.363 \\
	\bottomrule
\end{tabular}
\end{threeparttable}
\end{table}                                                                                                                     

\section{Spatial and temporal embedding} 

\begin{table}[htbp]
\centering
\caption{Ablation studies: multivariate long-term series prediction results on Traffic, Weather and Electricity with input length $720$ and prediction length in $\{96,192,336,720\}$. 
The original MPPN together with four embedding approaches is compared, with the best results highlighted in bold.}
\label{ablation-spembed}

\resizebox{\linewidth}{!}{
\begin{tabular}{cccccccccccc}
\toprule
\multicolumn{2}{c}{Methods}&\multicolumn{2}{c}{MPPN}&\multicolumn{2}{c}{+ L embedding}&\multicolumn{2}{c}{+H embedding}&\multicolumn{2}{c}{+ L \& H embedding}  &\multicolumn{2}{c}{+ S \& H embedding} \\
\midrule
\multicolumn{2}{c}{Metric} & MSE  & MAE & MSE & MAE& MSE  & MAE & MSE  & MAE & MSE & MAE\\
\midrule
\multirow{4}{*}{\rotatebox{90}{Traffic}} 

& 96 & \textbf{0.387} & \textbf{0.271} & 0.426 & 0.314 & 0.423 & 0.312 & 0.397 & 0.288 & 0.410 & 0.300 \\

& 192 & \textbf{0.396} & \textbf{0.273} & 0.433 & 0.312 &0.414 & 0.295 & 0.406 & 0.287  &  0.424 & 0.308\\

& 336 & \textbf{0.410} & \textbf{0.279} & 0.438 & 0.315 & 0.416 & 0.287 &  0.419 & 0.294  & 0.421 & 0.298 \\

& 720 & \textbf{0.449} & \textbf{0.301} & 0.467 &  0.326 & 0.478 & 0.336 & 0.457 & 0.316 & 0.454 & 0.312\\
\midrule

\multirow{4}{*}{\rotatebox{90}{Weather}} 

& 96 & \textbf{0.144} & \textbf{0.196} & 0.147 & 0.203 & 0.177 & 0.258 & 0.146  & 0.201  & 0.187  & 0.270 \\

& 192 & \textbf{0.189} & \textbf{0.240} & 0.193 & 0.253 &  0.213 & 0.284 & 0.192  & 0.245  & 0.240 &  0.315 \\

& 336 & \textbf{0.240} & \textbf{0.281} & 0.253 & 0.303 & 0.255 & 0.311 & 0.255  & 0.305  & 0.326 & 0.357\\

& 720 & \textbf{0.310} & \textbf{0.333} & 0.330  & 0.366 & 0.329 & 0.363 & 0.326 & 0.357 & 0.328 & 0.360\\

\midrule
\multirow{4}{*}{\rotatebox{90}{Etth1}} 

& 96 & \textbf{0.371} & \textbf{0.393} & 0.395 & 0.417 & 0.391 & 0.415 & 0.387  & 0.410 & 0.397& 0.421  \\

& 192 & \textbf{0.405} & \textbf{0.413} & 0.429 & 0.435 & 0.438 & 0.440 & 0.421  & 0.431 & 0.436 & 0.439 \\

& 336 & \textbf{0.426} & \textbf{0.425} & 0.453 & 0.453 & 0.460 & 0.456 & 0.469  & 0.463 & 0.470 & 0.462\\

& 720 & \textbf{0.436} & \textbf{0.452} & 0.484 & 0.487 & 0.495 & 0.498 & 0.511 & 0.508 & 0.491 & 0.496\\

\bottomrule
\end{tabular}
}
\end{table}

In this section, we perform ablation studies on Traffic, Weather and Electricity dataset to investigate the effect of different spatial and temporal embedding mechanisms: 1) \textbf{L embedding}: we add time stamp encoding of the historical $L$-step input data (Month, Day, Weekday, Hour, etc.); 2) \textbf{H embedding}: we add time stamp encoding of the $H$-step prediction output into MPPN; 3) \textbf{L \& H embedding}: we map the data at each time-step into a low dimensional vector both in the input and the output. 4) \textbf{S \& H embedding}: we consider spatial encoding for different channels of time series after pattern extraction together with the \textbf{H embedding}.  It can be observed from Table \ref{ablation-spembed} that none of the introduced spatio-temporal encoding mechanisms yields performance improvement for the original MPPN. This indicates that time encoding may be not necessary for time series modeling, as the multi-resolution and multi-period patterns extracted by MPPN could well capture the temporal features of time series, thus enabling efficient and accurate long-term time series prediction.

\section{Supplementary of main results}  
\begin{figure}[htbp]
  \centering
  \begin{minipage}{0.32\textwidth}
    \centerline{\includegraphics[width=\textwidth]{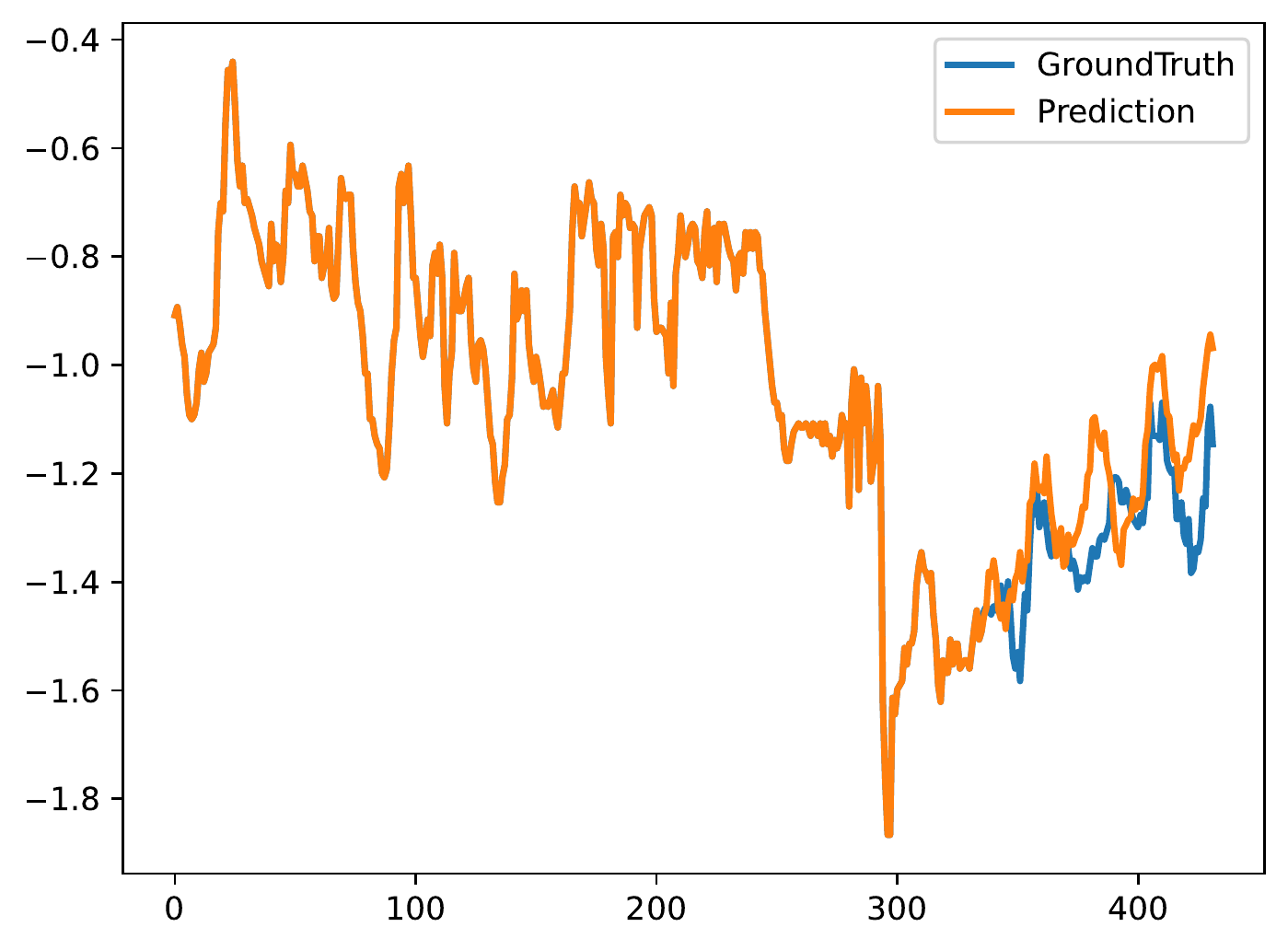}}
    \centerline{(a) MPPN}
    \centerline{\includegraphics[width=\textwidth]{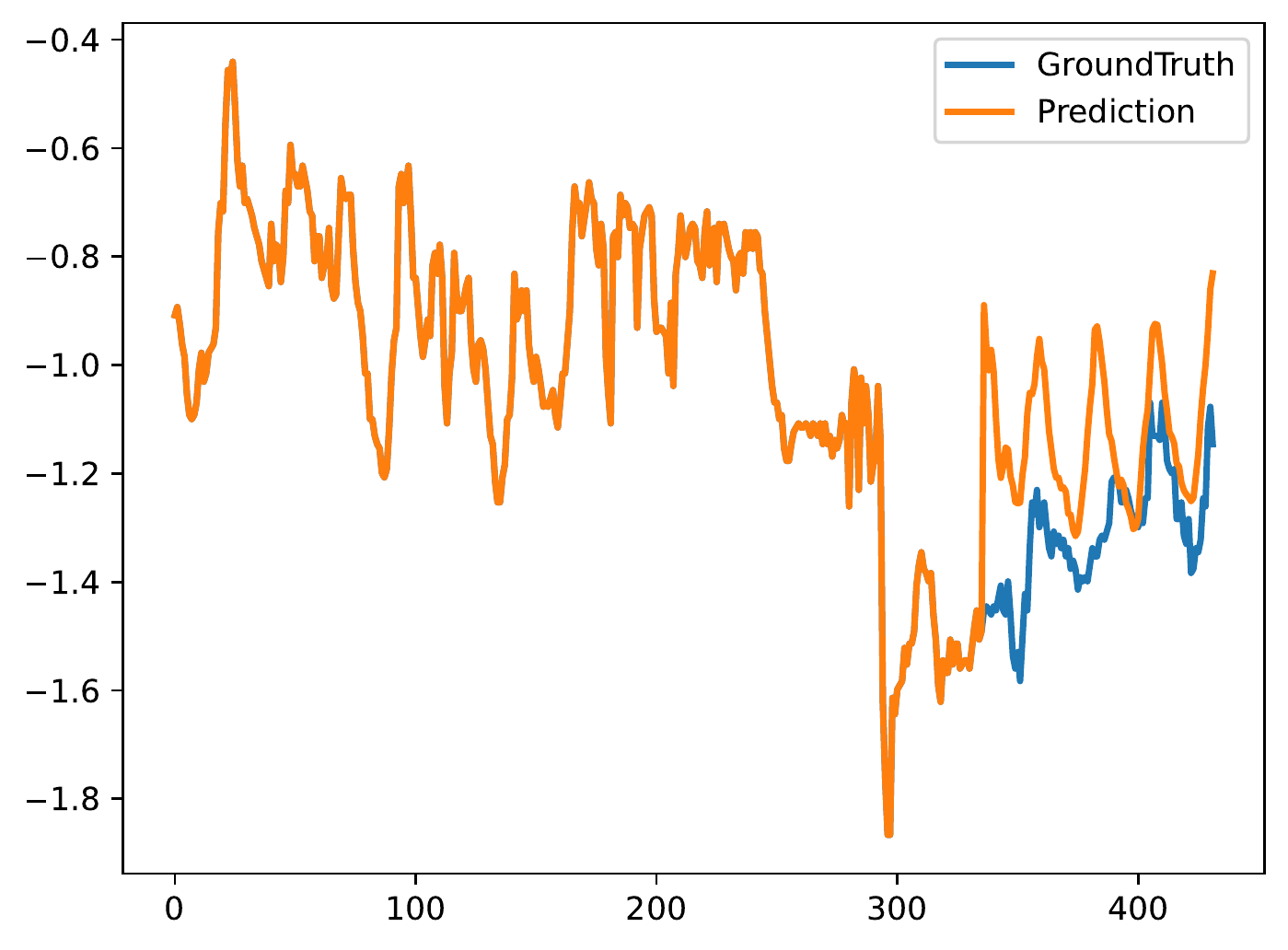}}
    \centerline{(d) FEDformer}
  \end{minipage}
  \hfill 
  \begin{minipage}{0.32\textwidth}
    \centerline{\includegraphics[width=\textwidth]{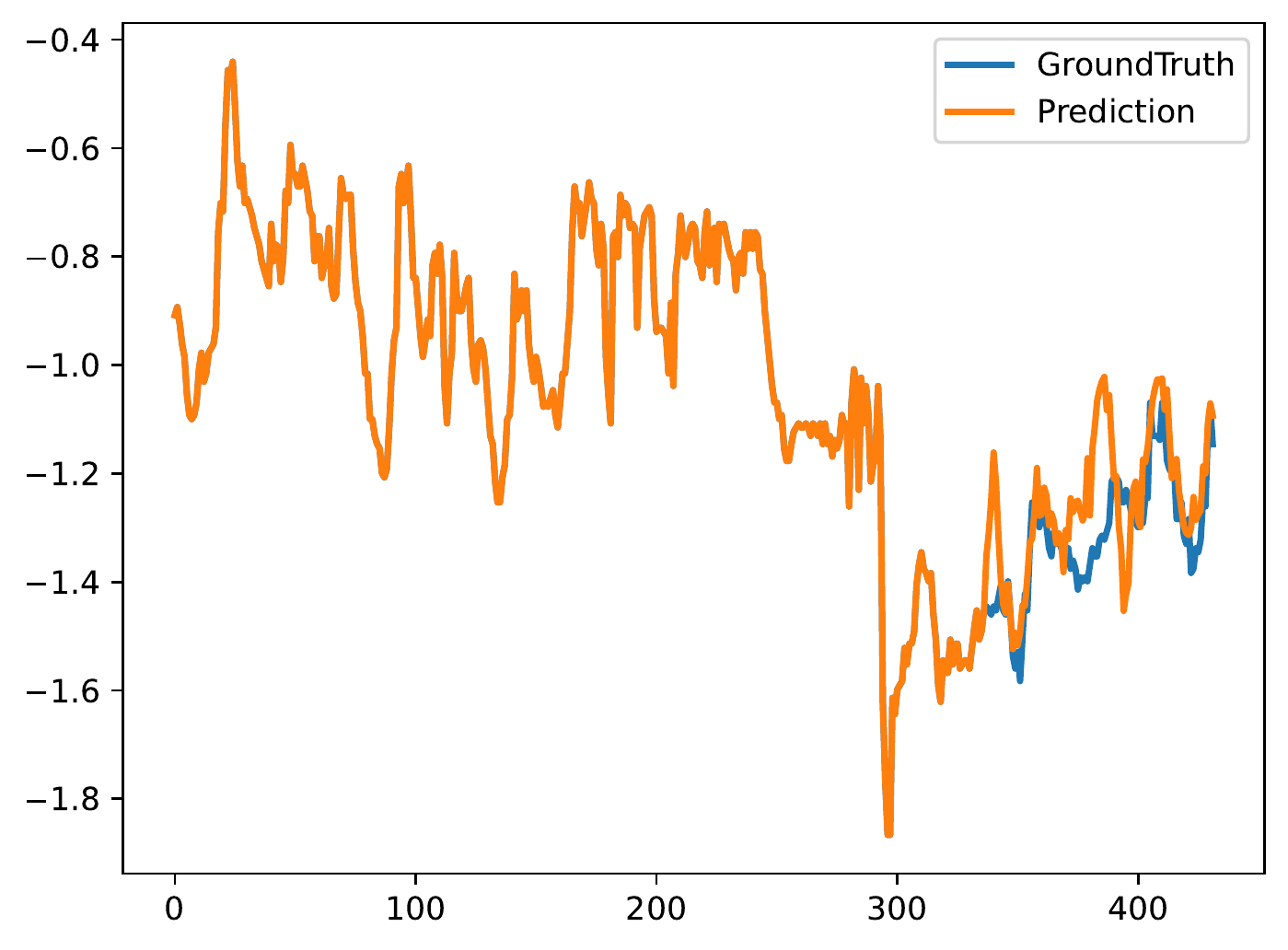}}
    \centerline{(b) DLinear}  
    \centerline{\includegraphics[width=\textwidth]{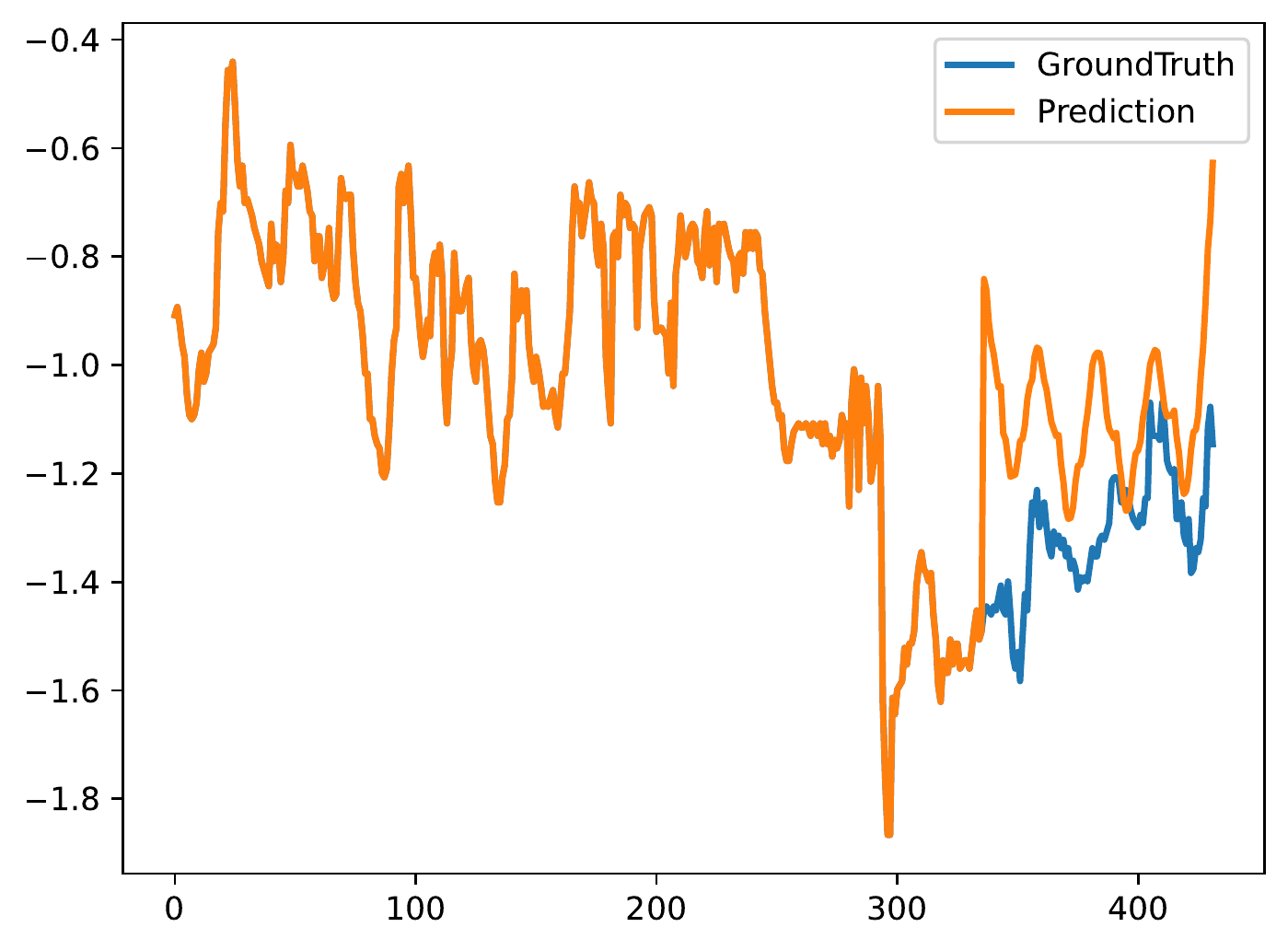}}
    \centerline{(e) Autoformer}
  \end{minipage}
  \hfill
  \begin{minipage}{0.32\textwidth}
    \centerline{\includegraphics[width=\textwidth]{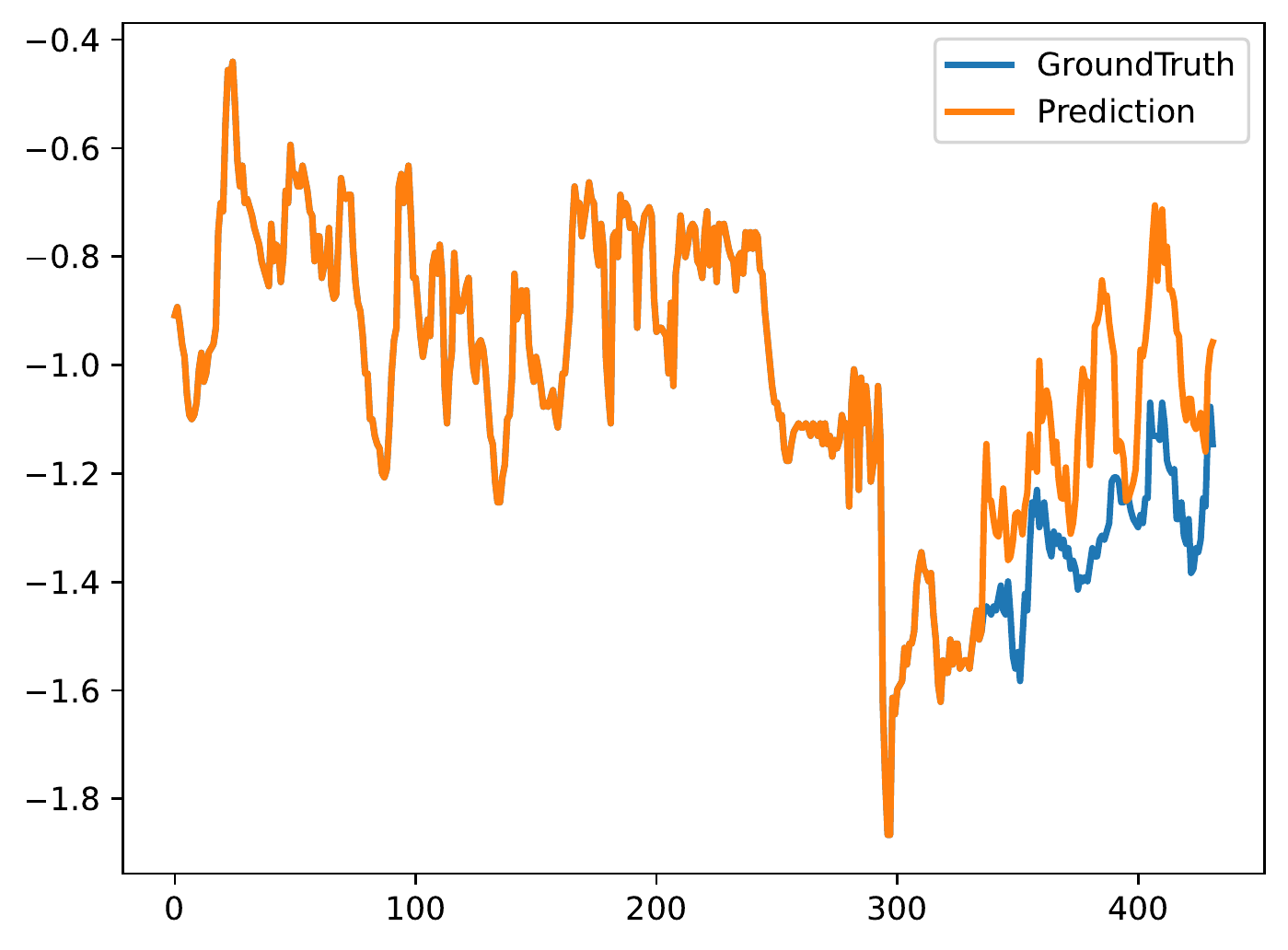}}
    \centerline{(c) MICN}
    \centerline{\includegraphics[width=\textwidth]{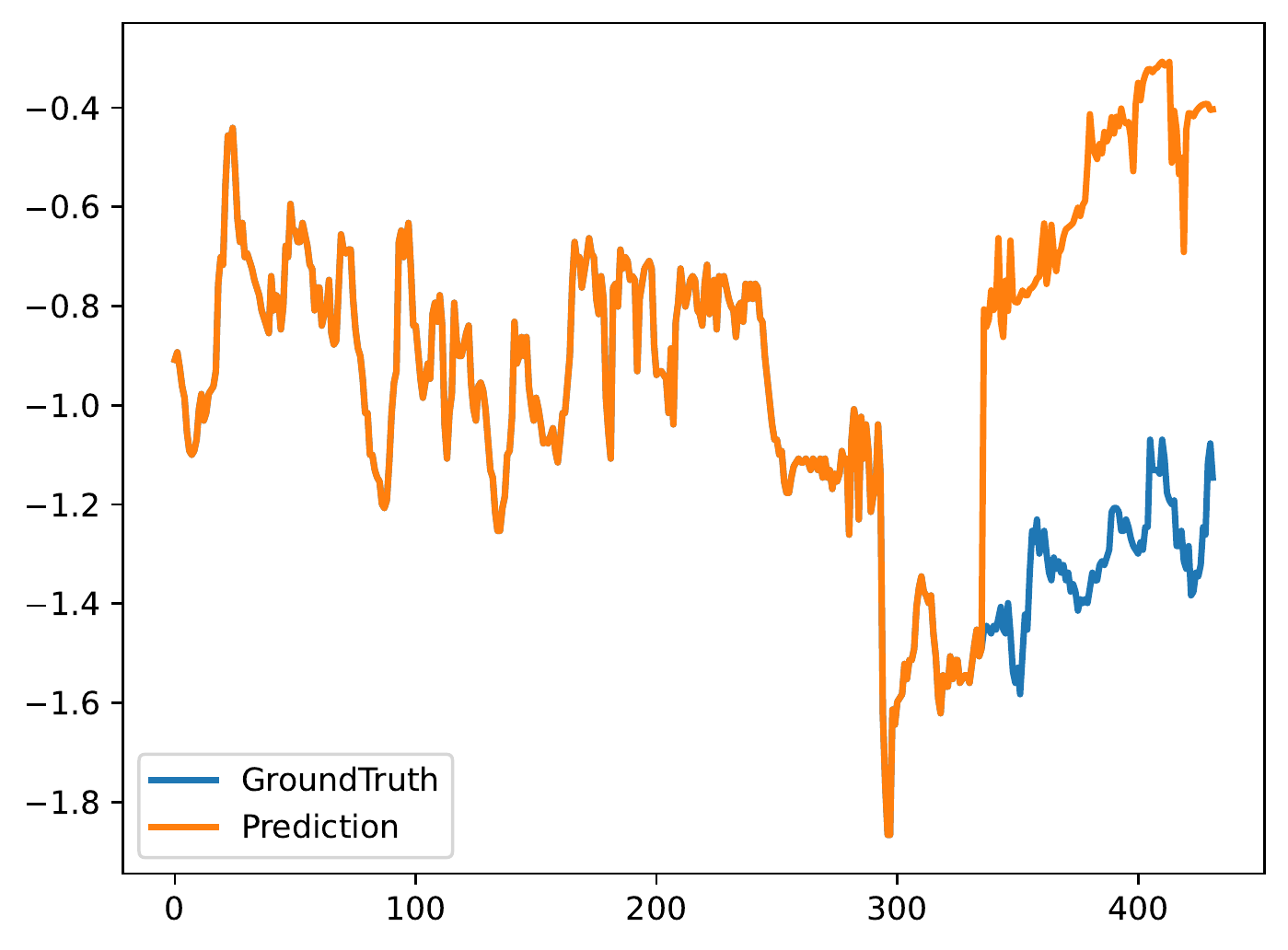}}
    \centerline{(f) Informer}
  \end{minipage}
  \caption{The prediction results on the ETTh1 dataset under the input-336-predict-96 settings.}
  \label{etth1_96_plot}
\end{figure}

\begin{figure}[htbp]
  \centering
  \begin{minipage}{0.32\textwidth}
    \centerline{\includegraphics[width=\textwidth]{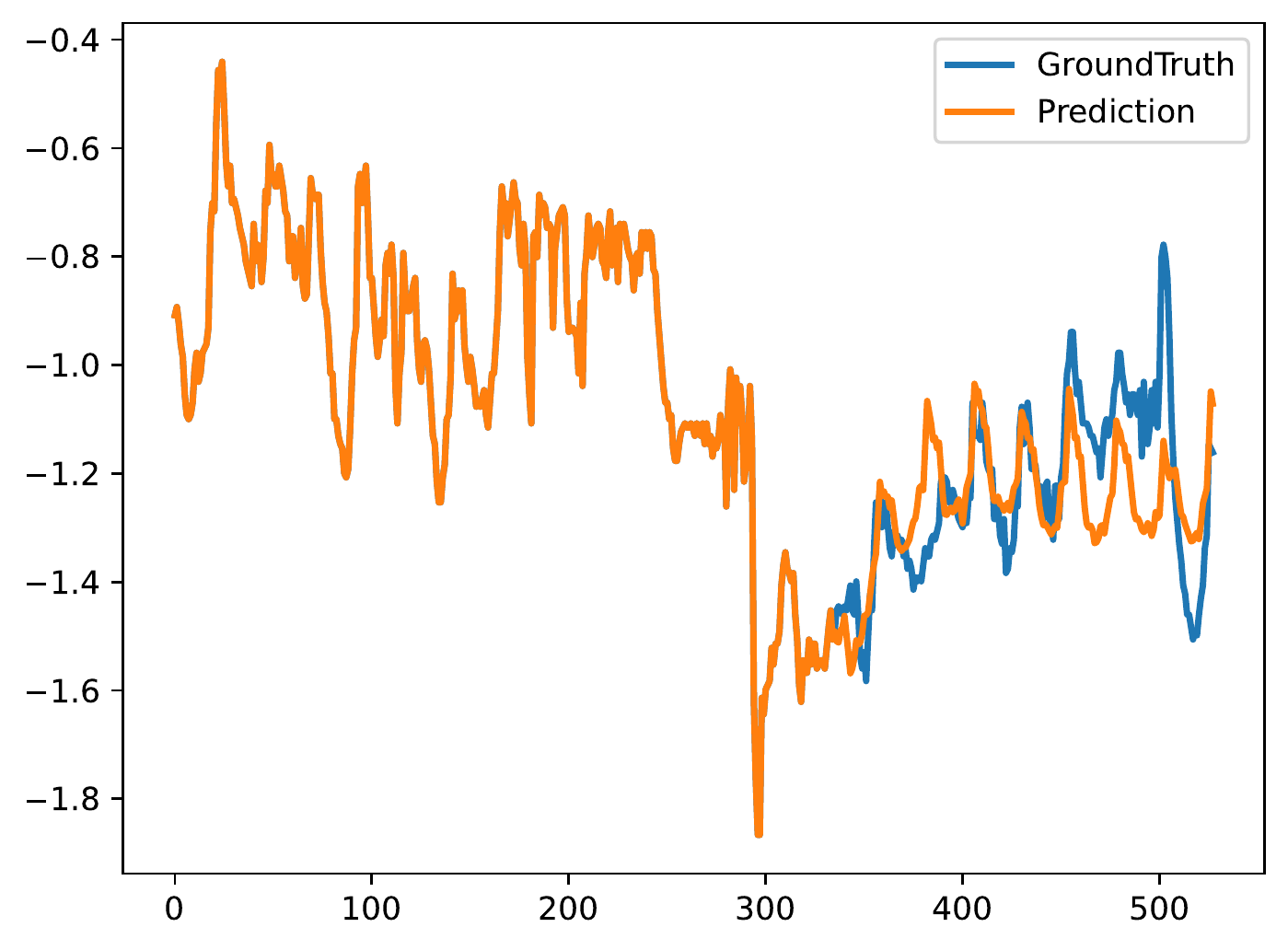}}
    \centerline{(a) MPPN}
    \centerline{\includegraphics[width=\textwidth]{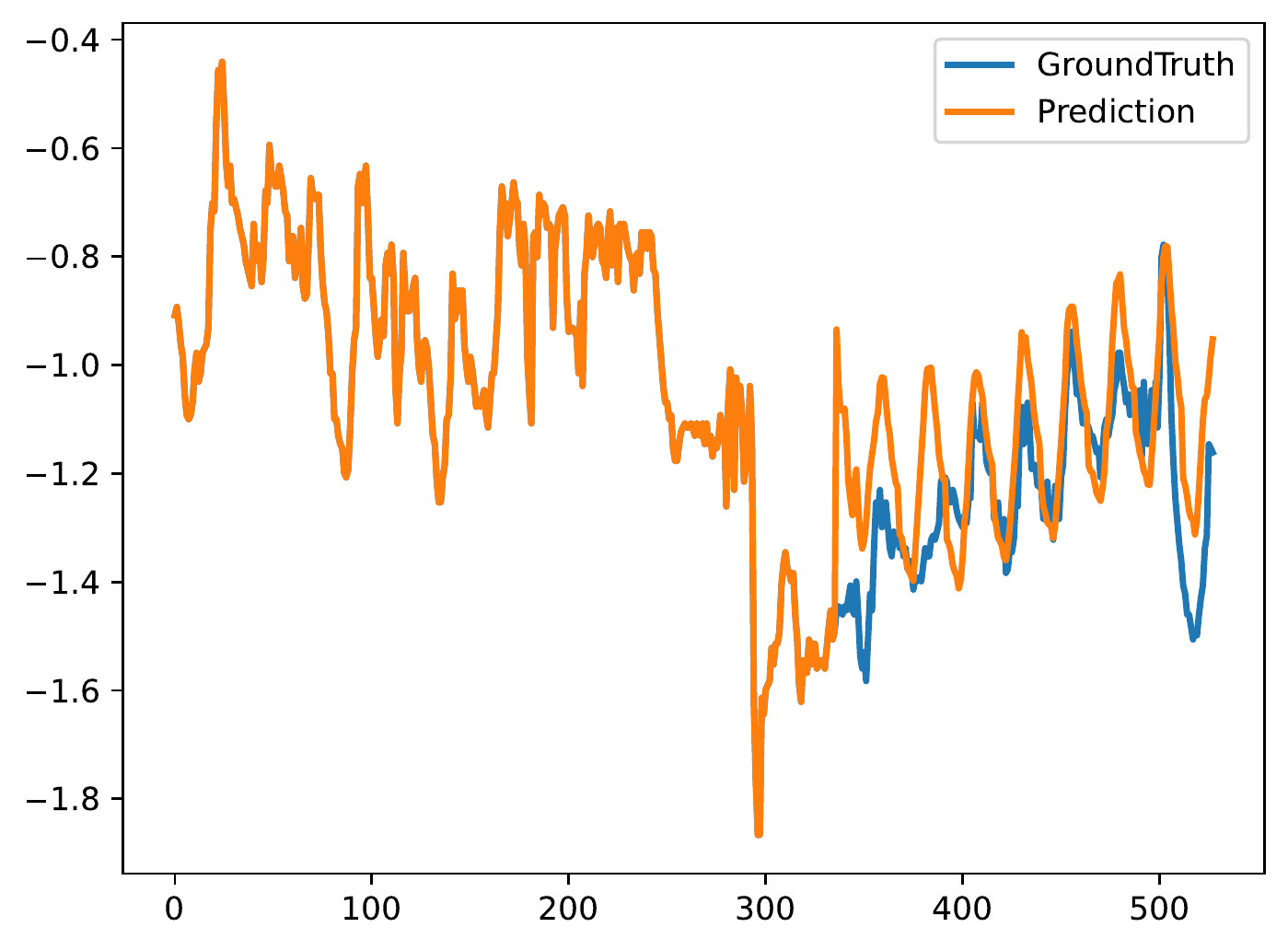}}
    \centerline{(d) FEDformer}
  \end{minipage}
  \hfill 
  \begin{minipage}{0.32\textwidth}
    \centerline{\includegraphics[width=\textwidth]{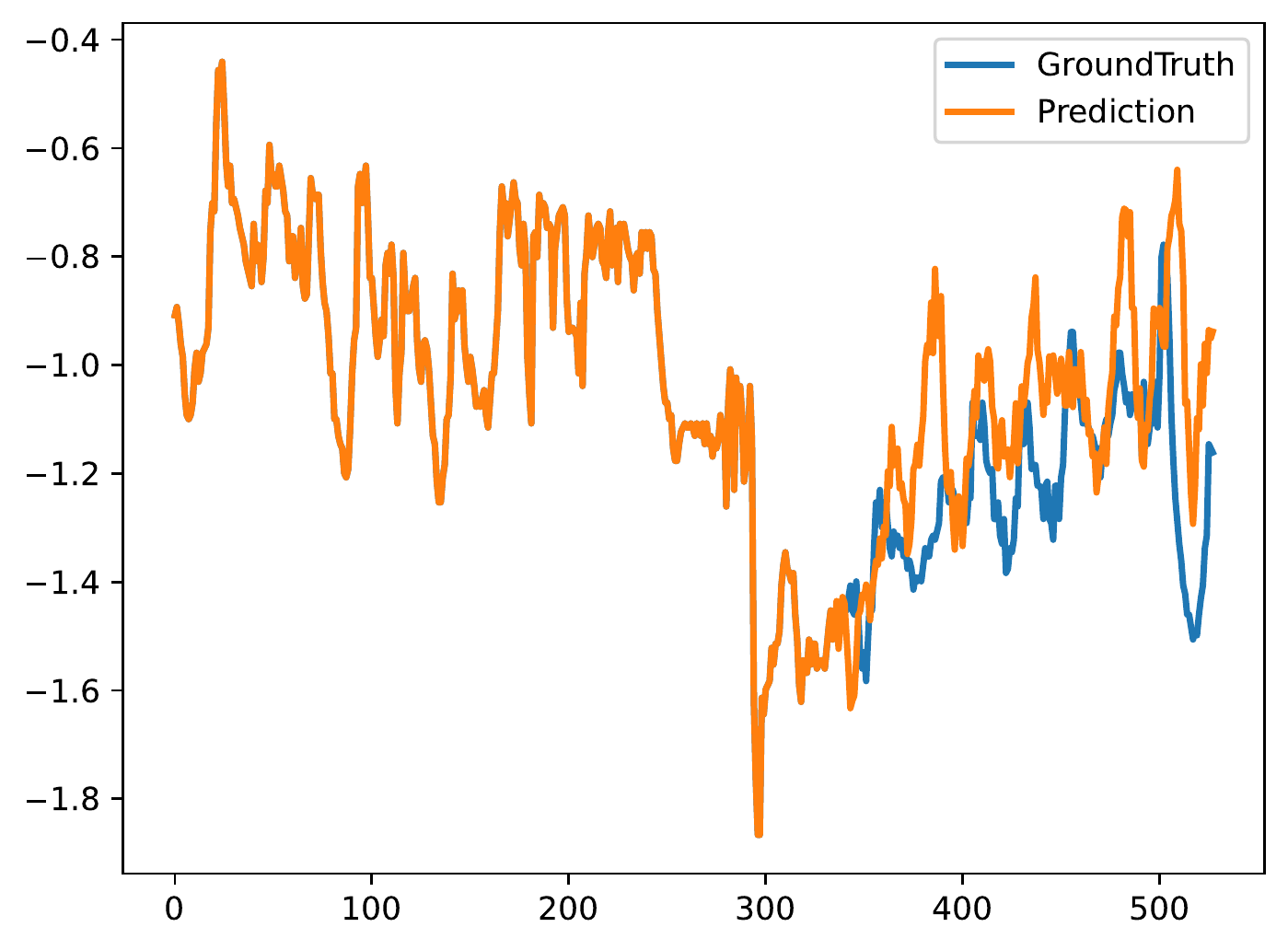}}
    \centerline{(b) DLinear}  
    \centerline{\includegraphics[width=\textwidth]{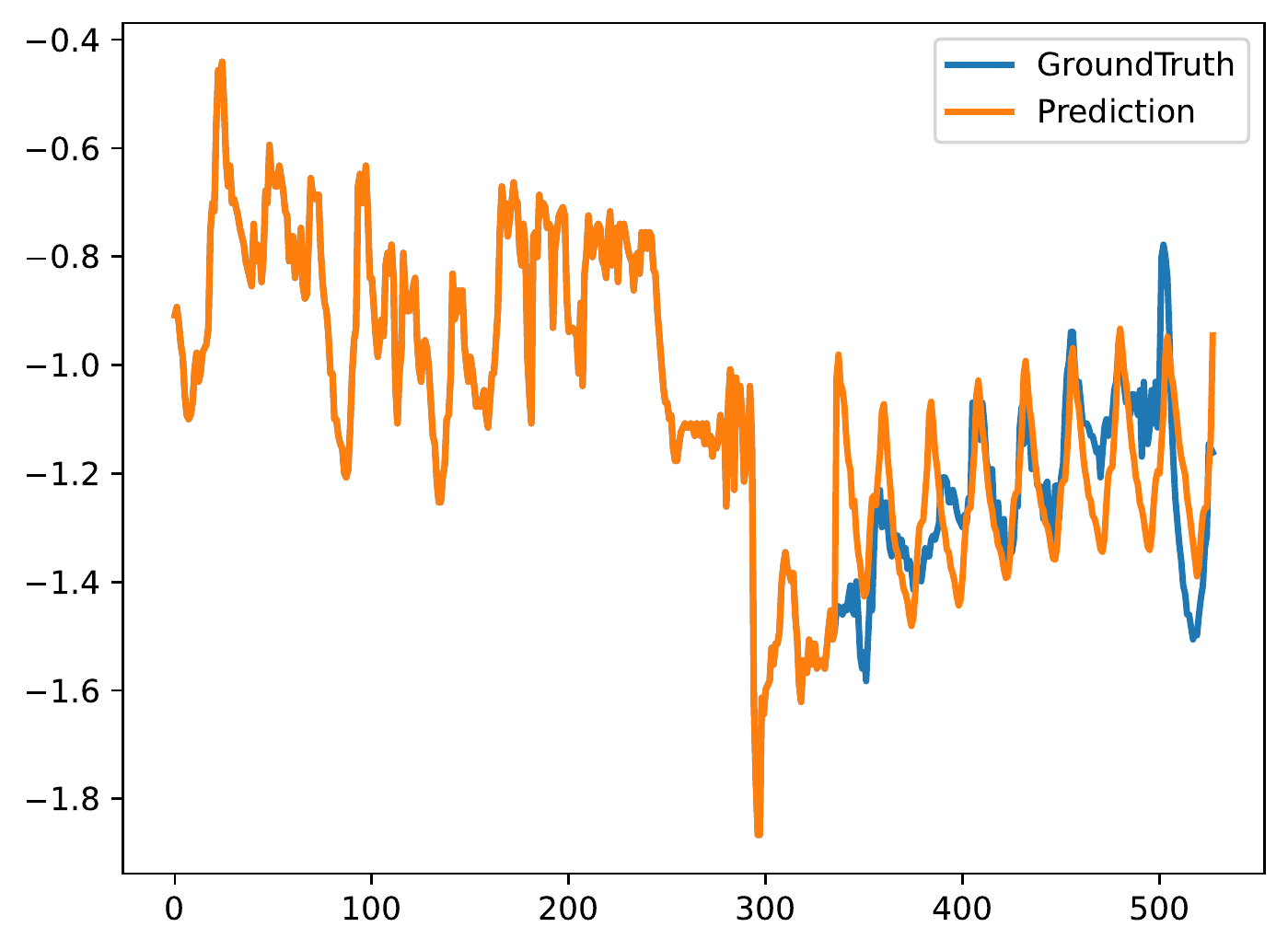}}
    \centerline{(e) Autoformer}
  \end{minipage}
  \hfill
  \begin{minipage}{0.32\textwidth}
    \centerline{\includegraphics[width=\textwidth]{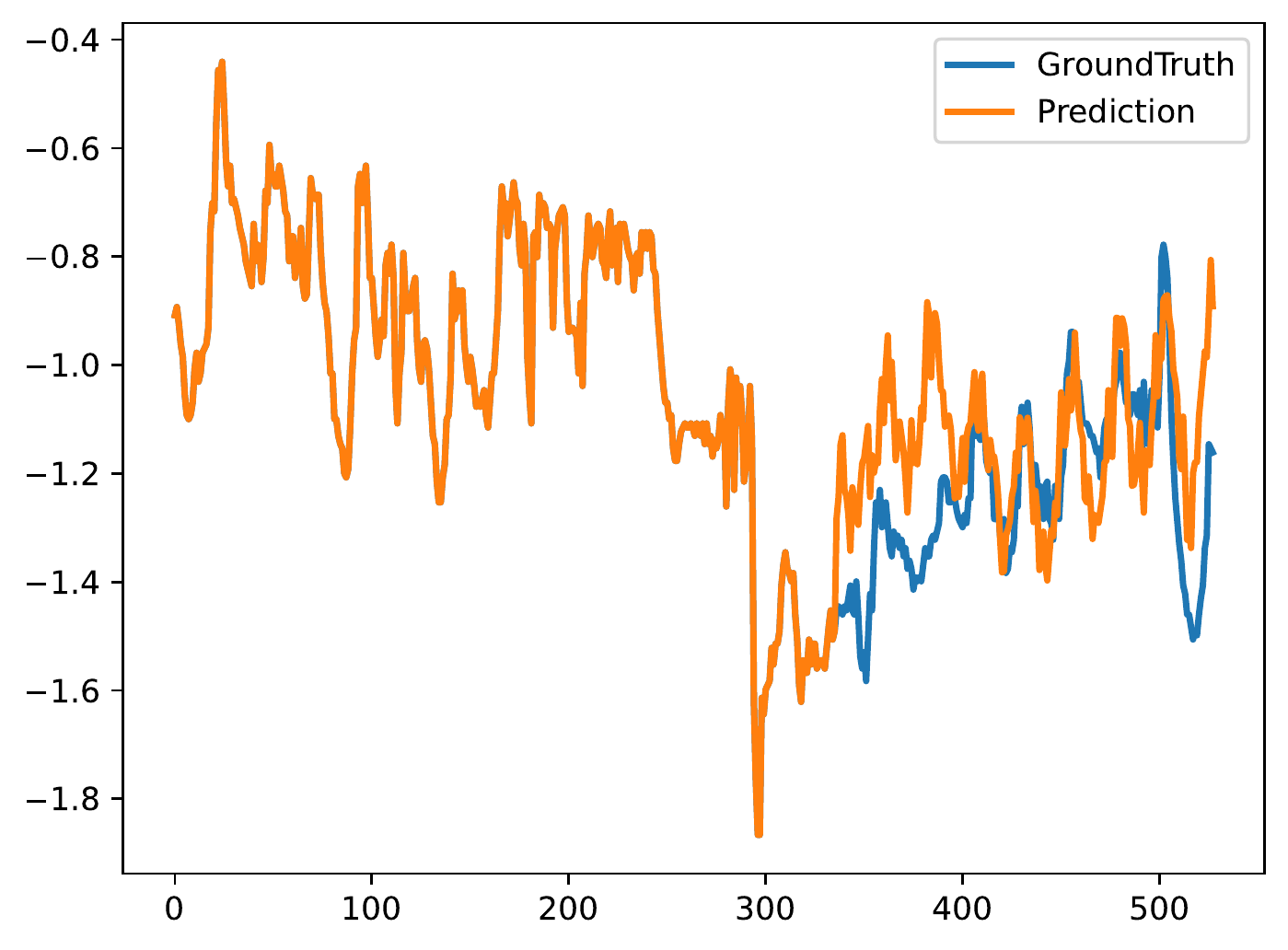}}
    \centerline{(c) MICN}
    \centerline{\includegraphics[width=\textwidth]{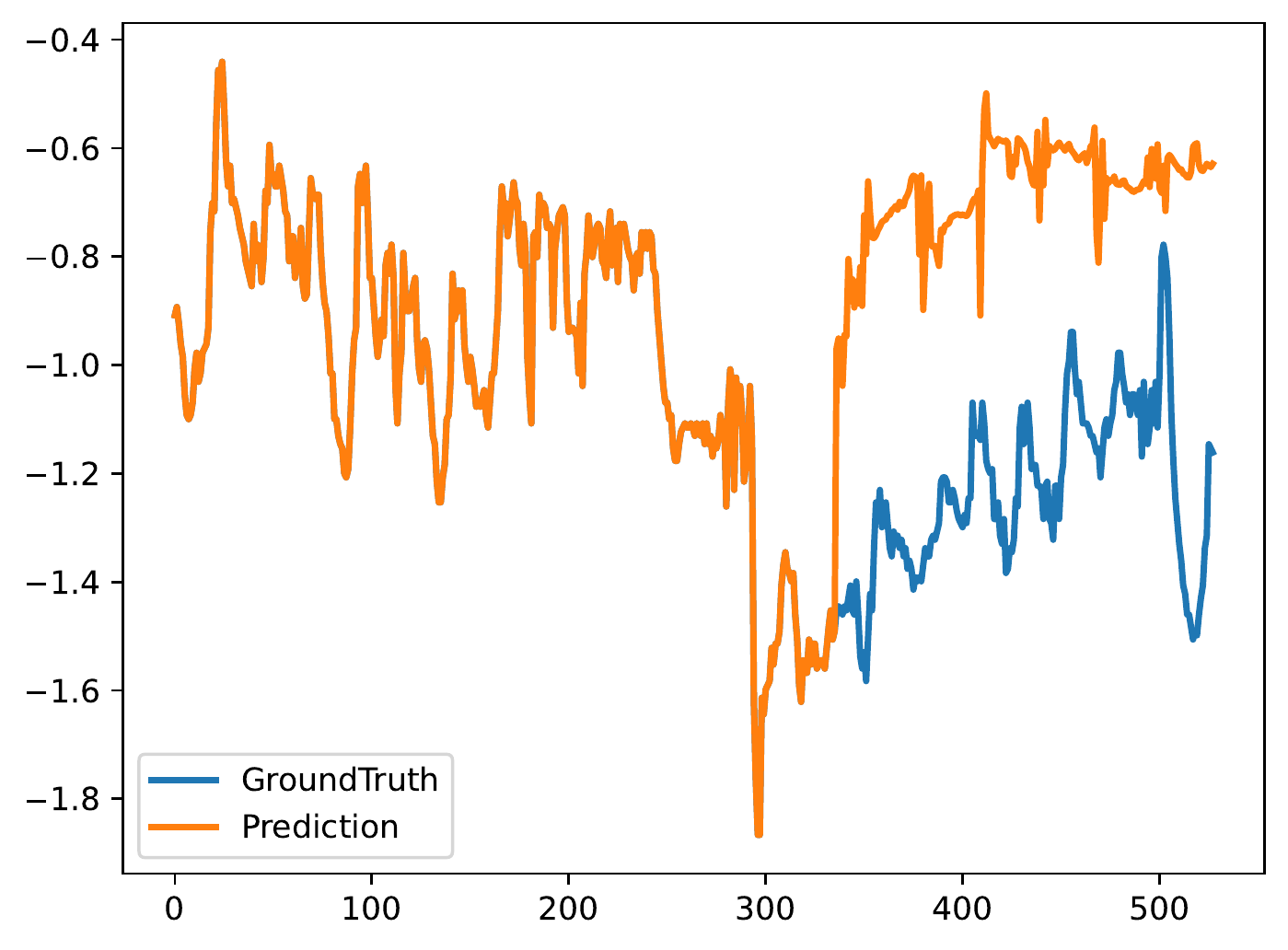}}
    \centerline{(f) Informer}
  \end{minipage}
  \caption{The prediction results on the ETTh1 dataset under the input-336-predict-192 settings.}
  \label{etth1_192_plot}
\end{figure}

\begin{figure}[htbp]
  \centering
  \begin{minipage}{0.32\textwidth}
    \centerline{\includegraphics[width=\textwidth]{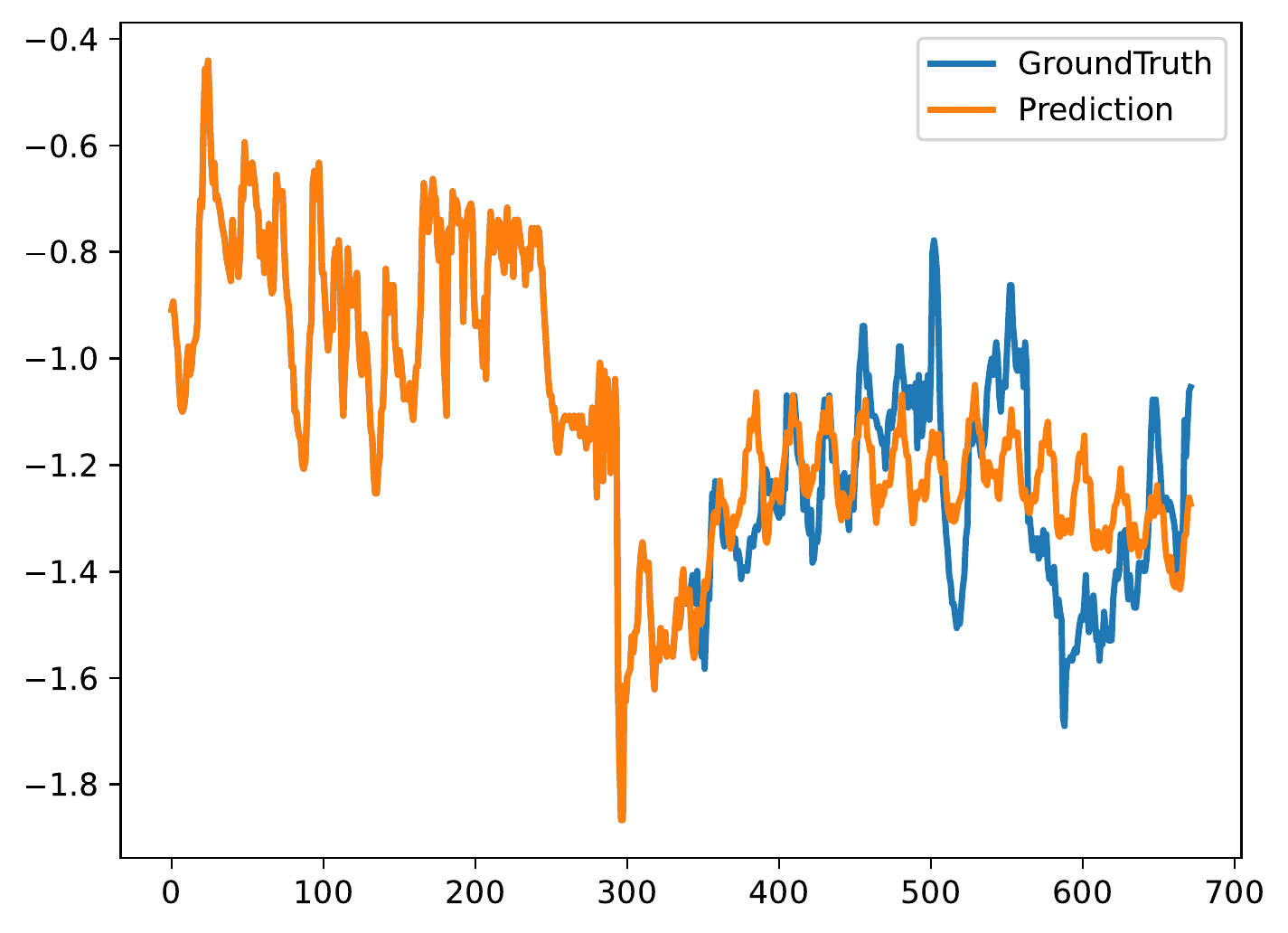}}
    \centerline{(a) MPPN}
    \centerline{\includegraphics[width=\textwidth]{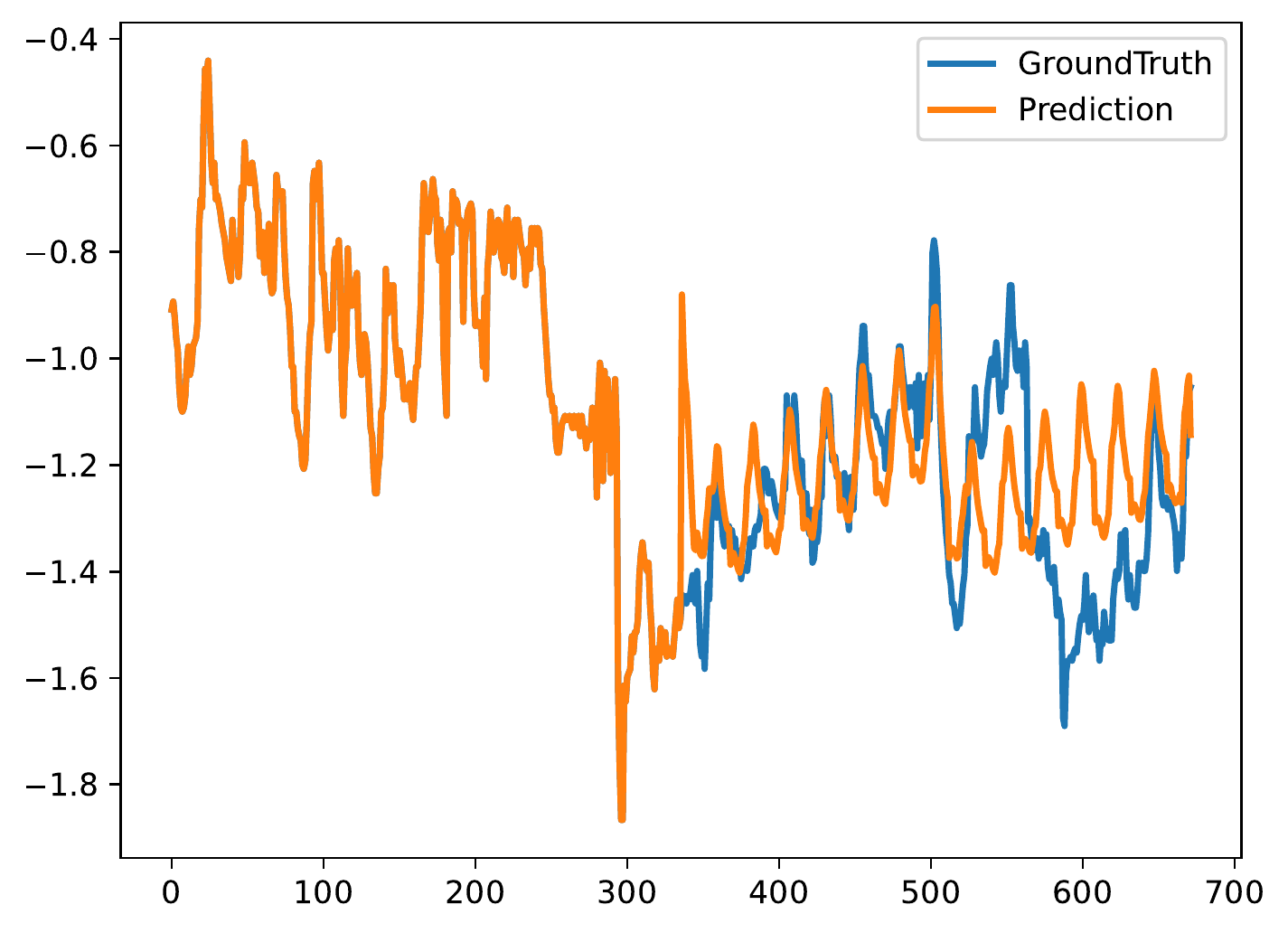}}
    \centerline{(d) FEDformer}
  \end{minipage}
  \hfill 
  \begin{minipage}{0.32\textwidth}
    \centerline{\includegraphics[width=\textwidth]{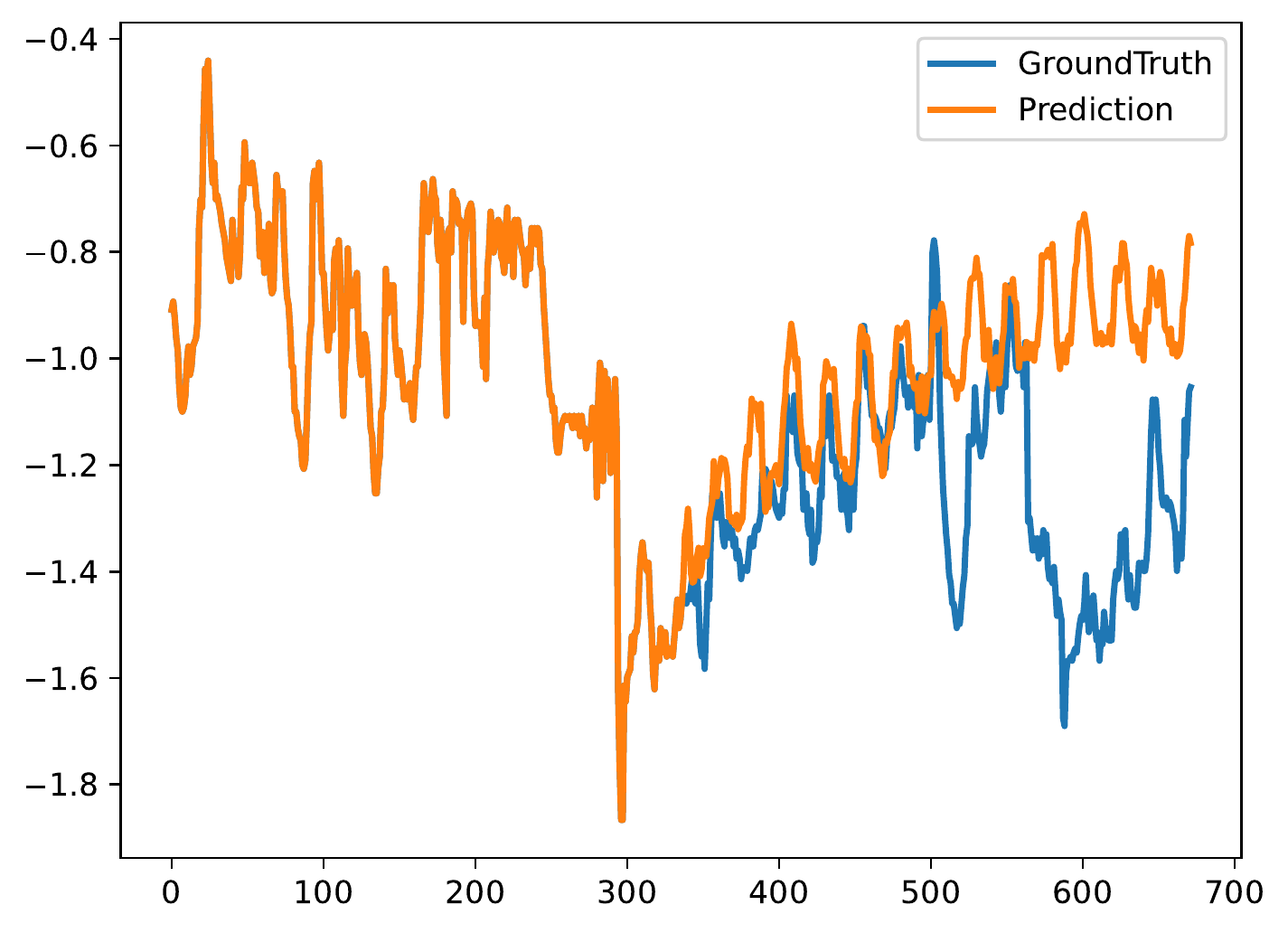}}
    \centerline{(b) DLinear}  
    \centerline{\includegraphics[width=\textwidth]{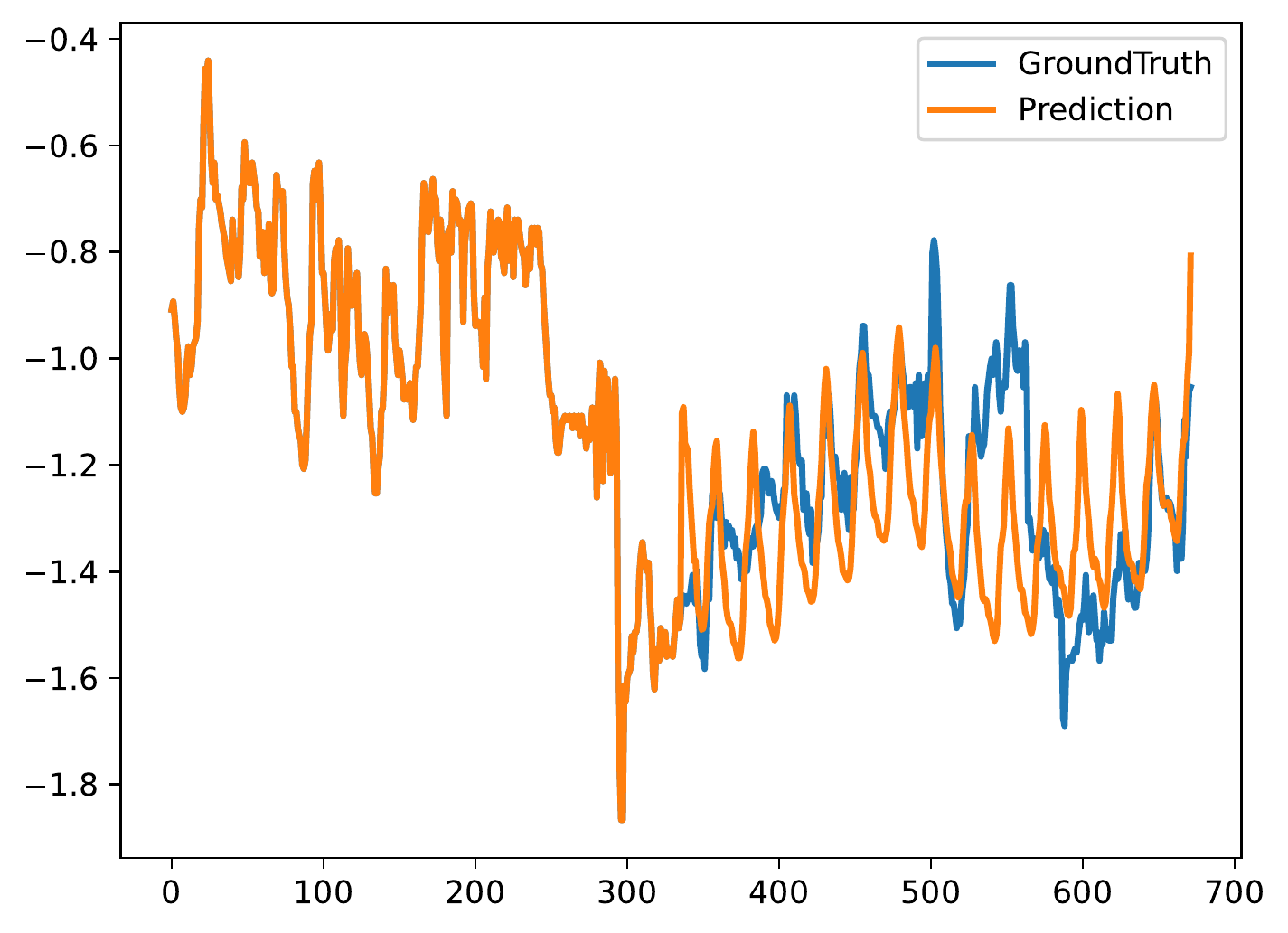}}
    \centerline{(e) Autoformer}
  \end{minipage}
  \hfill
  \begin{minipage}{0.32\textwidth}
    \centerline{\includegraphics[width=\textwidth]{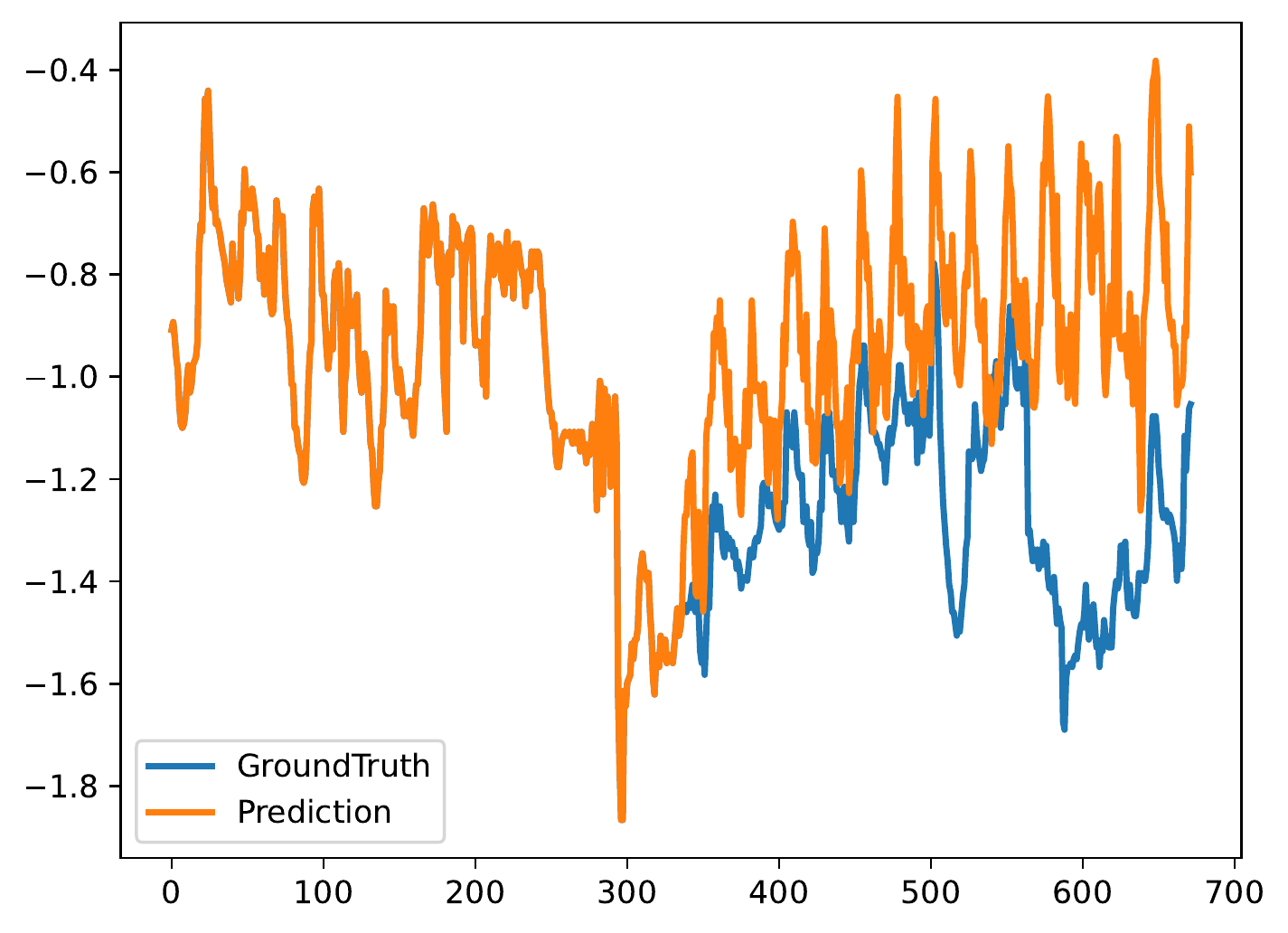}}
    \centerline{(c) MICN}
    \centerline{\includegraphics[width=\textwidth]{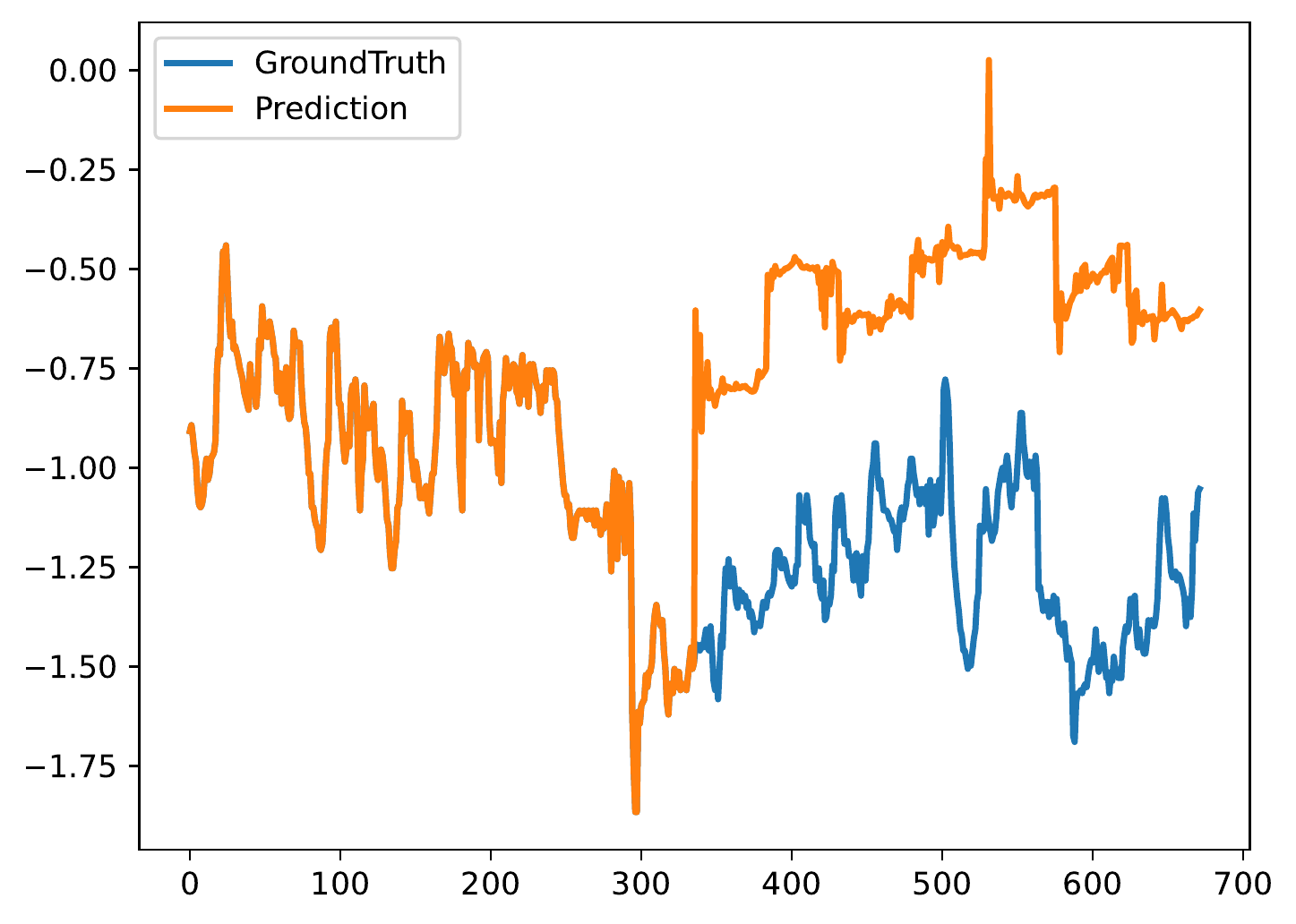}}
    \centerline{(f) Informer}
  \end{minipage}
  \caption{The prediction results on the ETTh1 dataset under the input-336-predict-336 settings.}
  \label{etth1_336_plot}
\end{figure}

\begin{figure}[htbp]
  \centering
  \begin{minipage}{0.32\textwidth}
    \centerline{\includegraphics[width=\textwidth]{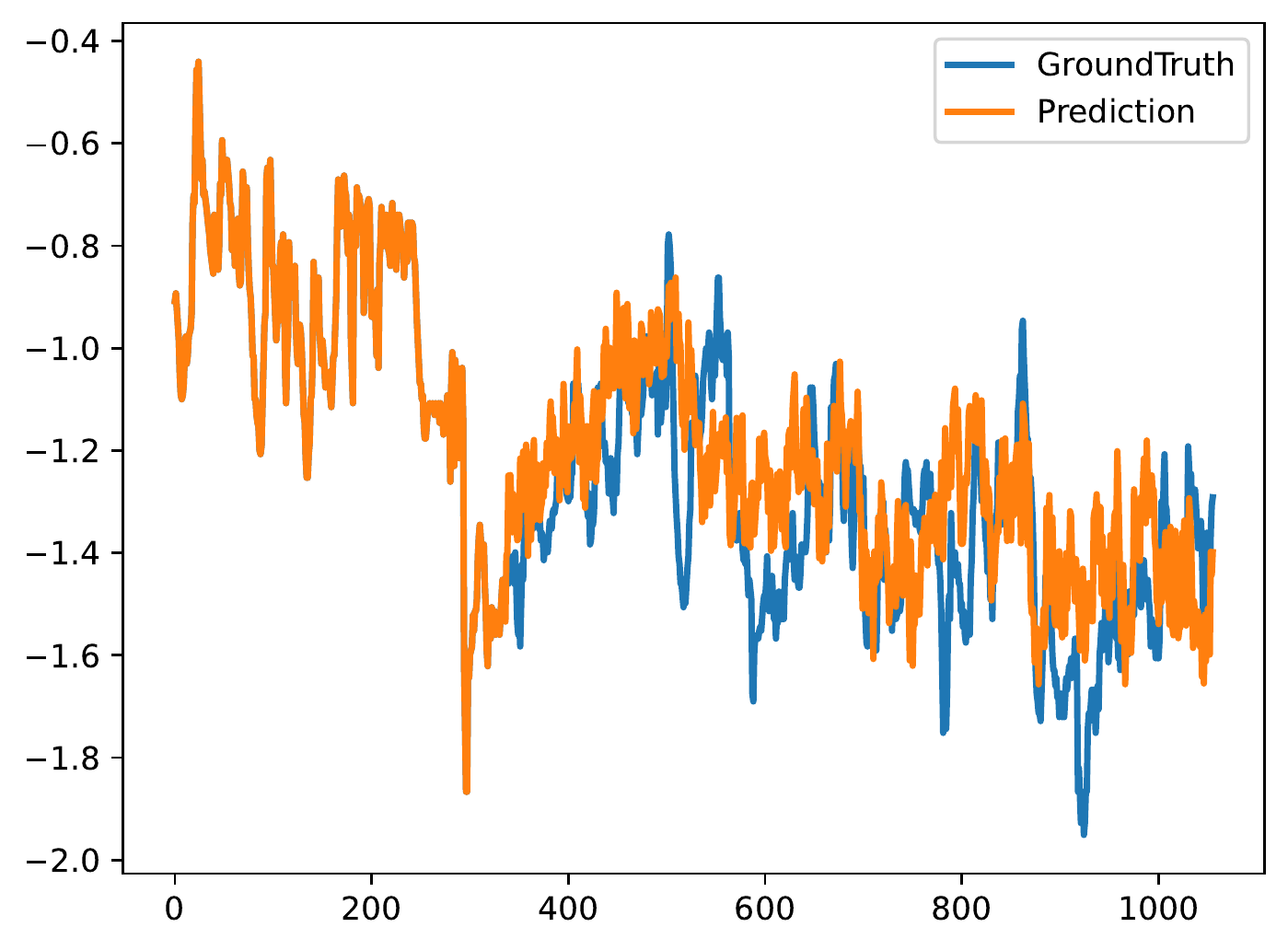}}
    \centerline{(a) MPPN}
    \centerline{\includegraphics[width=\textwidth]{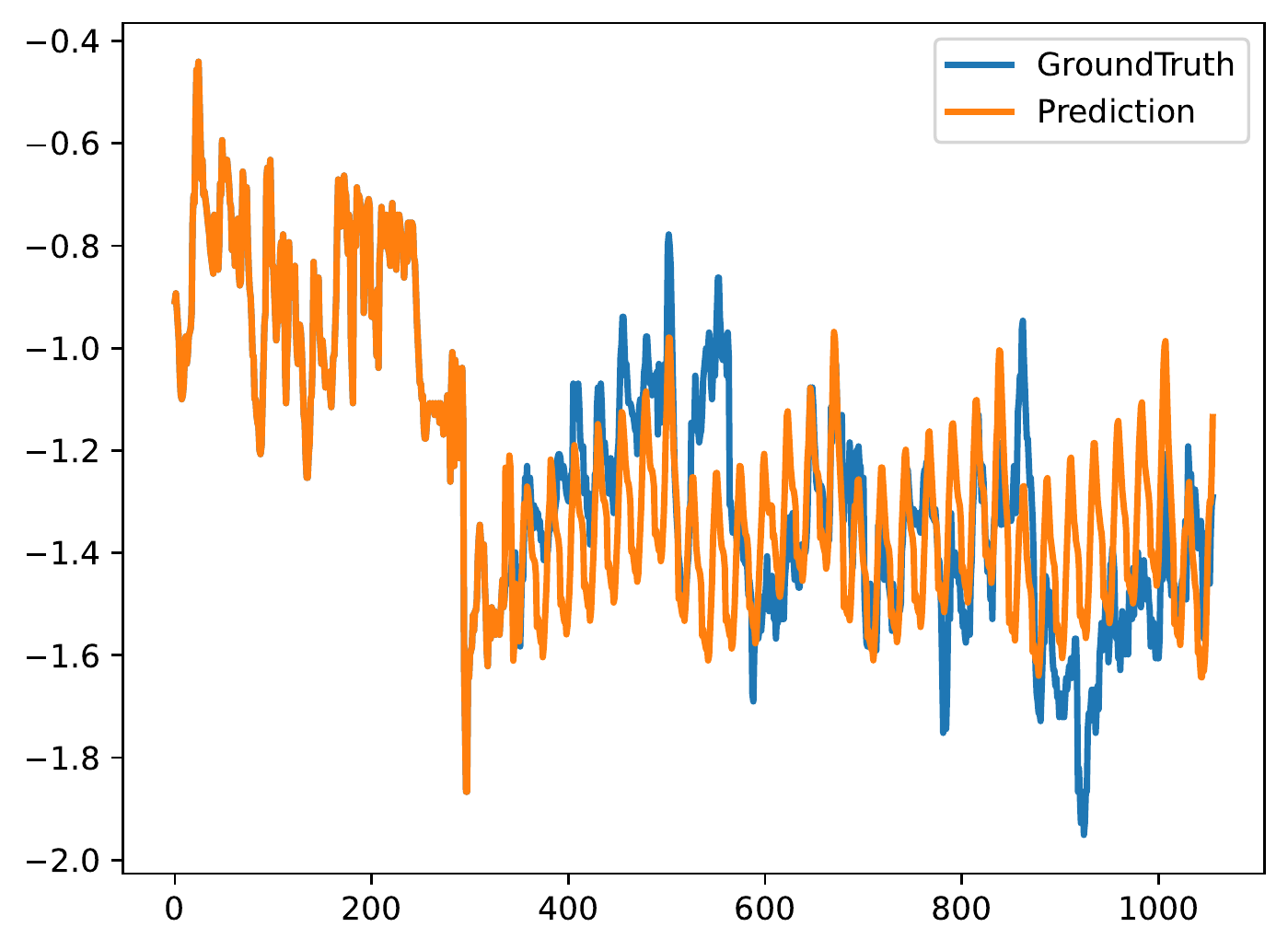}}
    \centerline{(d) FEDformer}
  \end{minipage}
  \hfill 
  \begin{minipage}{0.32\textwidth}
    \centerline{\includegraphics[width=\textwidth]{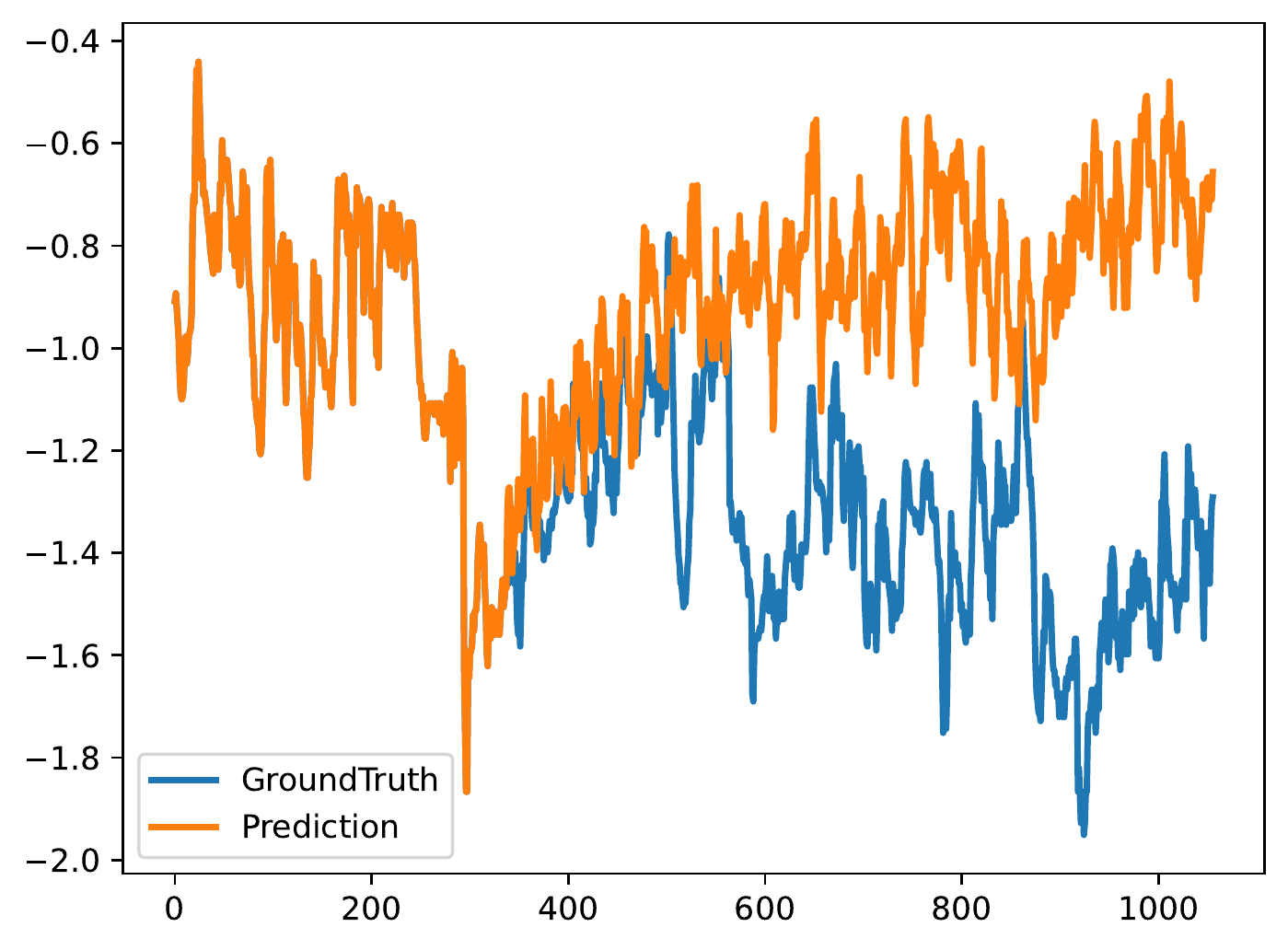}}
    \centerline{(b) DLinear}  
    \centerline{\includegraphics[width=\textwidth]{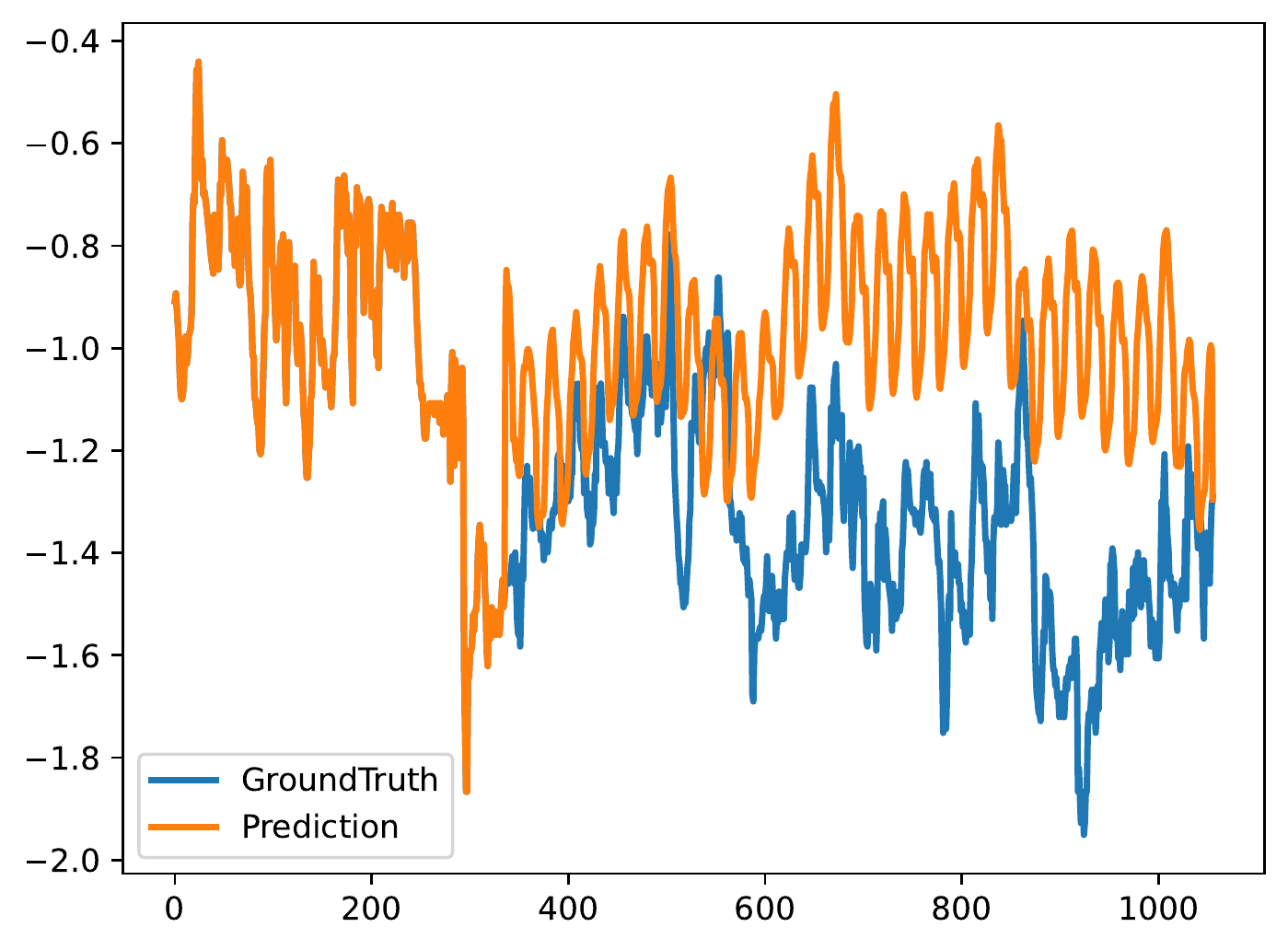}}
    \centerline{(d) Autoformer}
  \end{minipage}
  \hfill
  \begin{minipage}{0.32\textwidth}
    \centerline{\includegraphics[width=\textwidth]{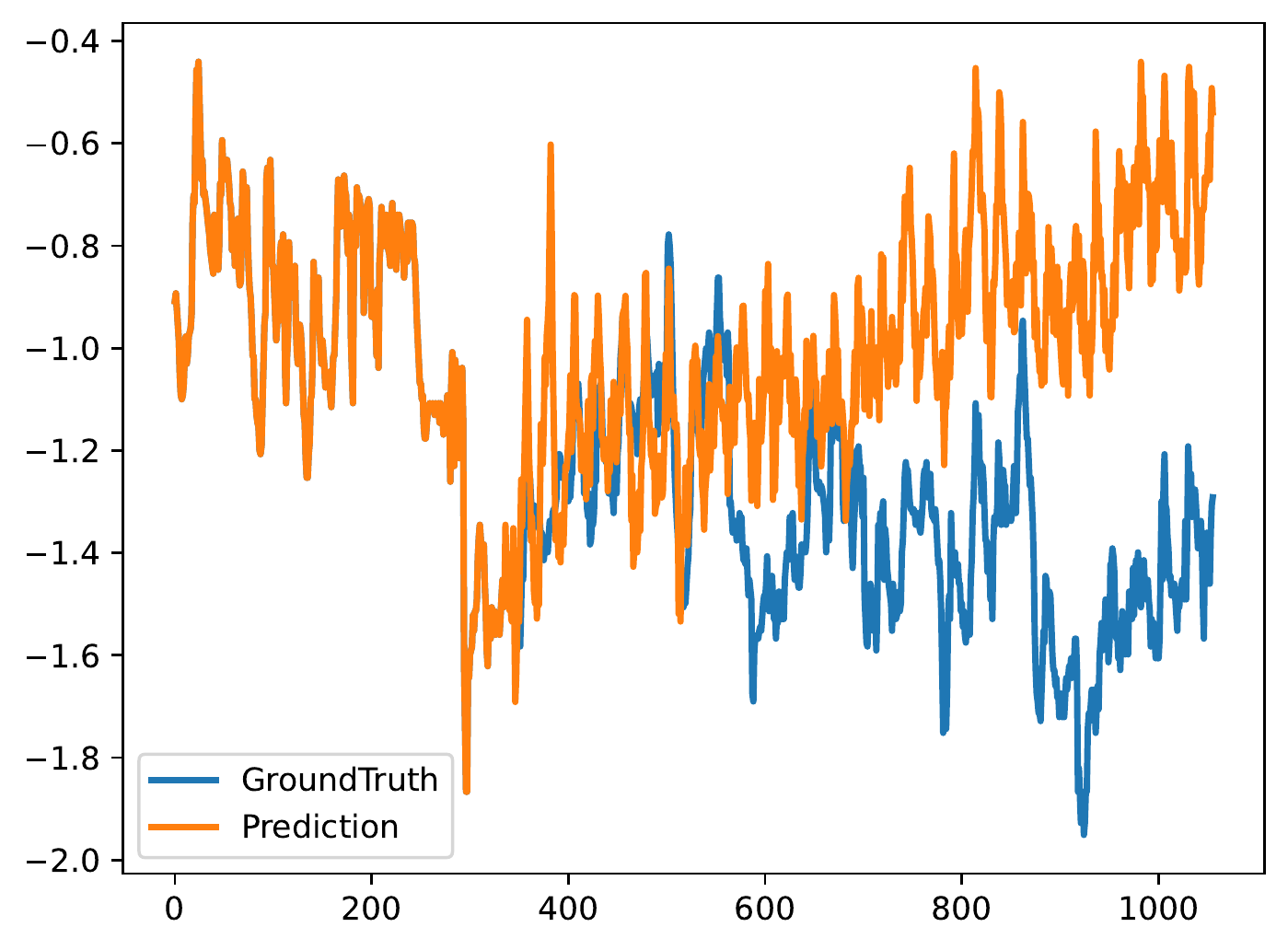}}
    \centerline{(c) MICN}
    \centerline{\includegraphics[width=\textwidth]{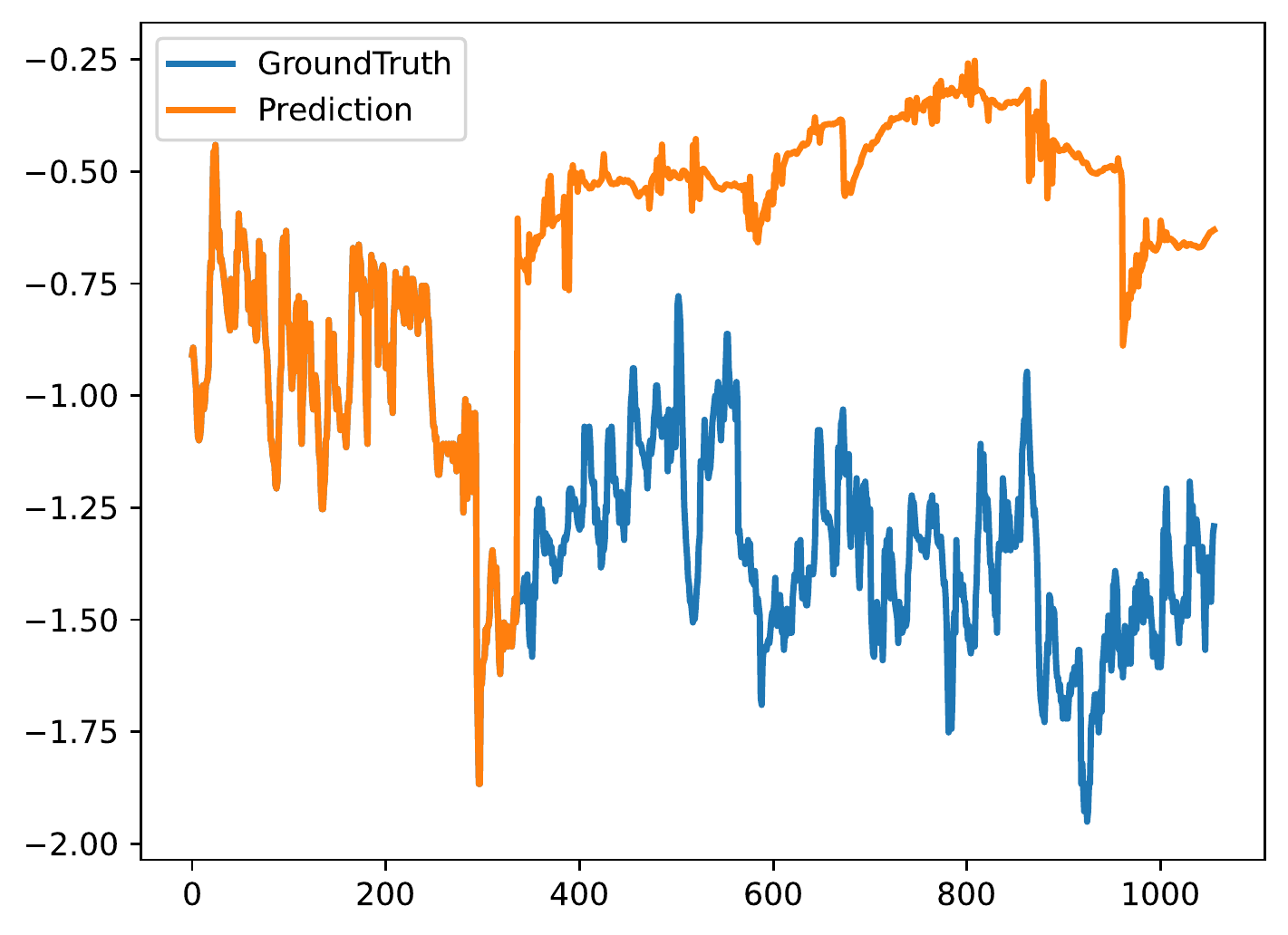}}
    \centerline{(d) Informer}
  \end{minipage}
  \caption{The prediction results on the ETTh1 dataset under the input-336-predict-720 settings.}
  \label{etth1_720_plot}
\end{figure}

\begin{figure}[htbp]
  \centering
  \begin{minipage}{0.32\textwidth}
    \centerline{\includegraphics[width=\textwidth]{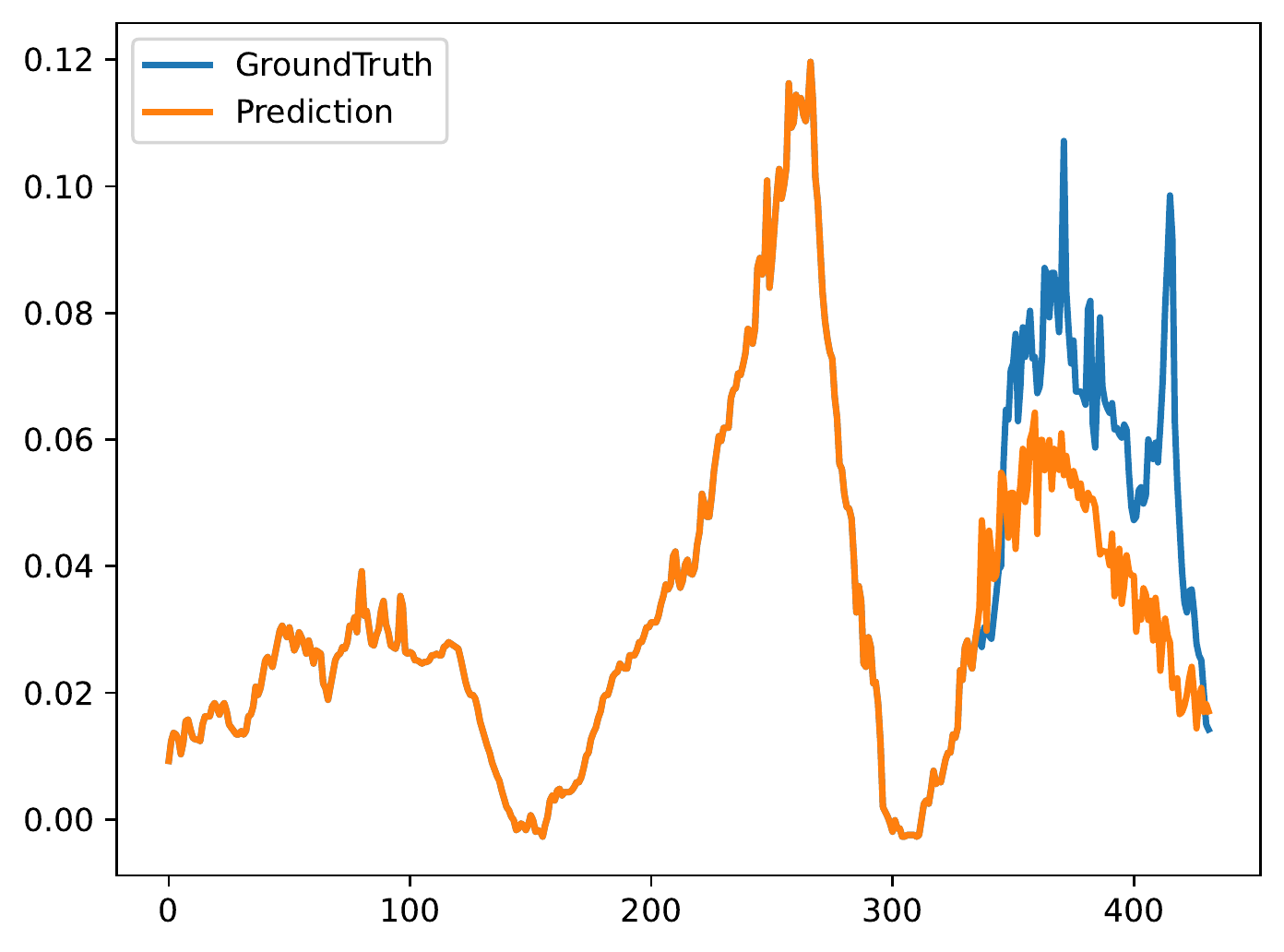}}
    \centerline{(a) MPPN}
    \centerline{\includegraphics[width=\textwidth]{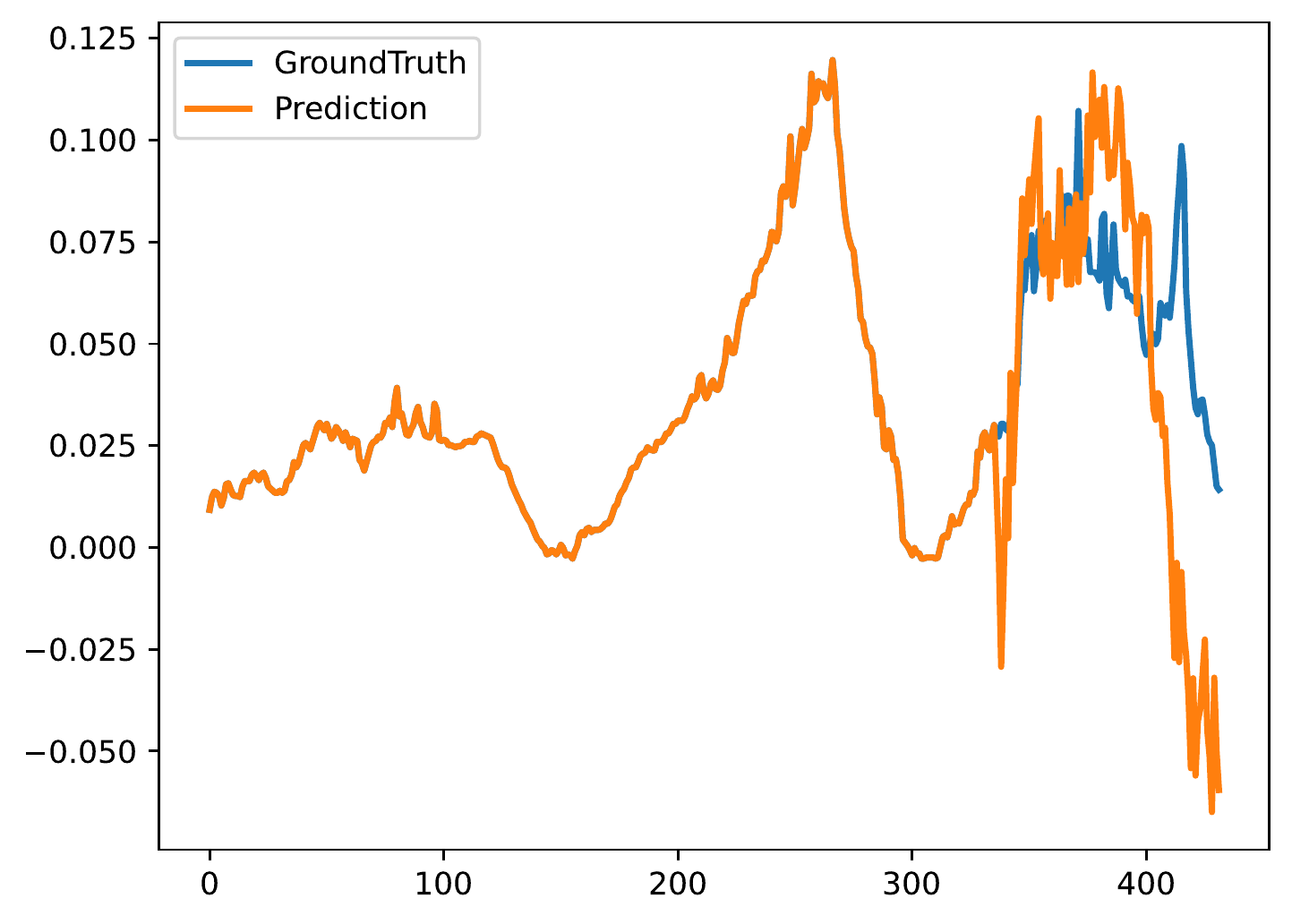}}
    \centerline{(d) FEDformer}
  \end{minipage}
  \hfill 
  \begin{minipage}{0.32\textwidth}
    \centerline{\includegraphics[width=\textwidth]{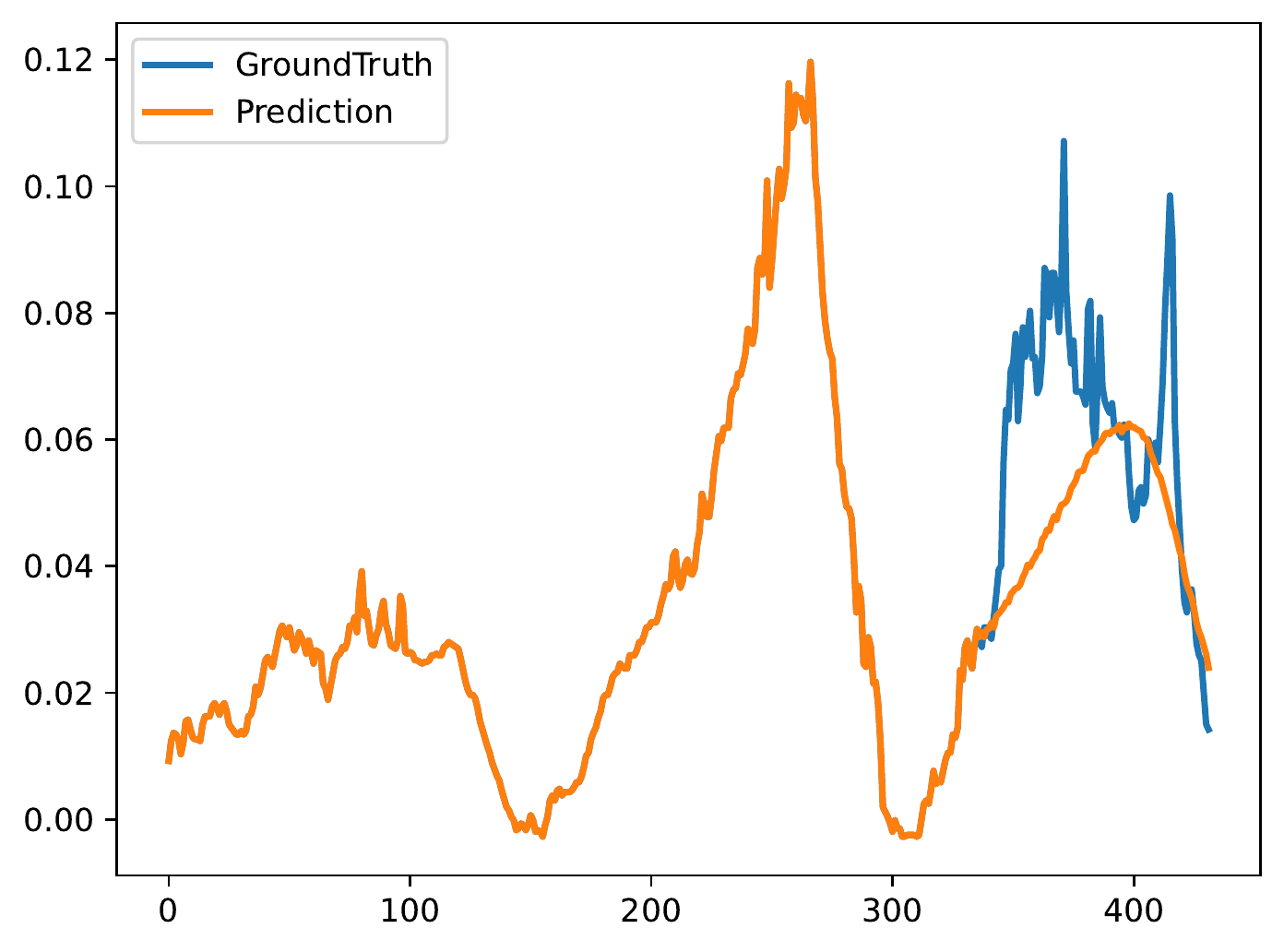}}
    \centerline{(b) DLinear}  
    \centerline{\includegraphics[width=\textwidth]{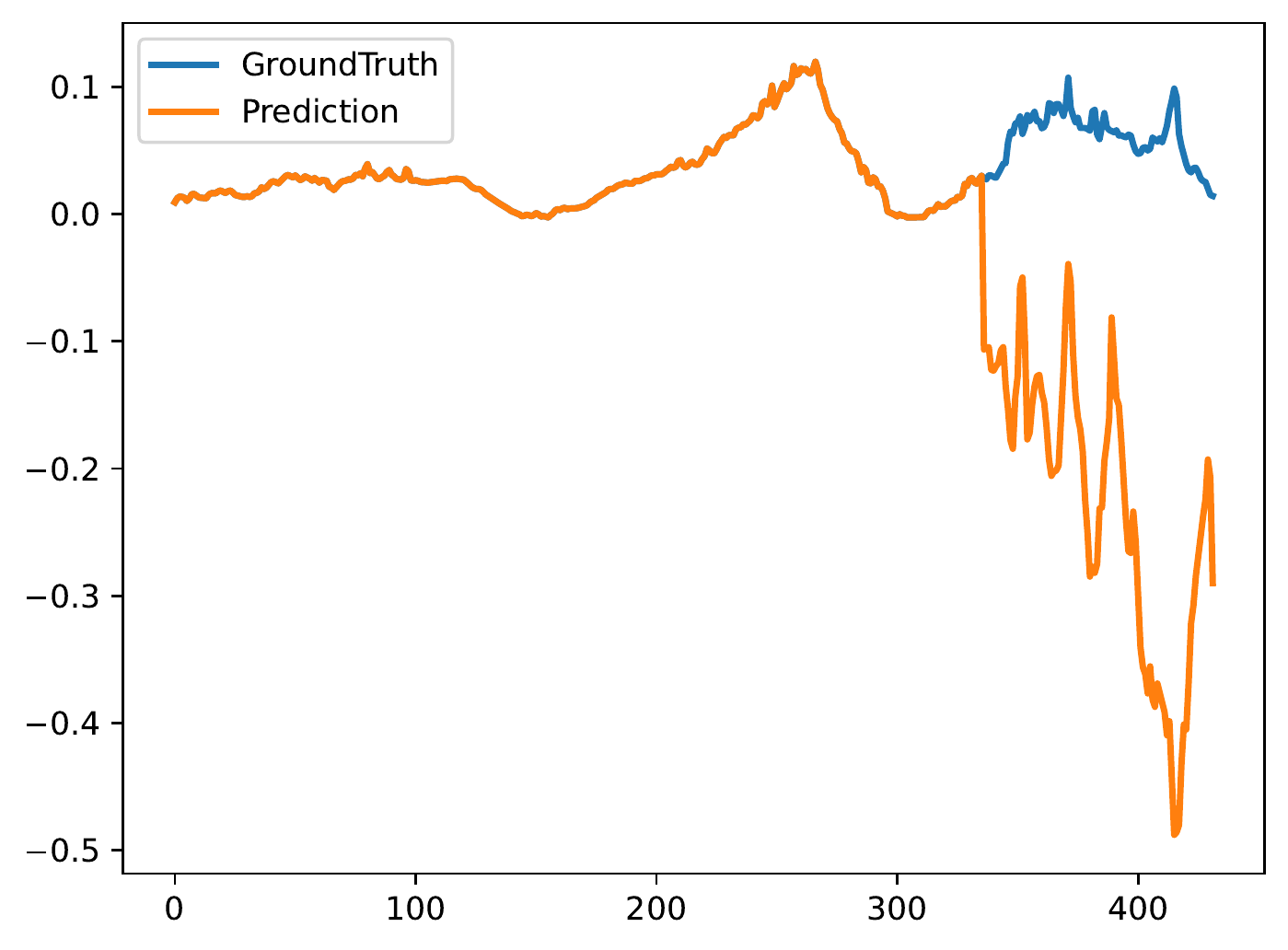}}
    \centerline{(e) Autoformer}
  \end{minipage}
  \hfill
  \begin{minipage}{0.32\textwidth}
    \centerline{\includegraphics[width=\textwidth]{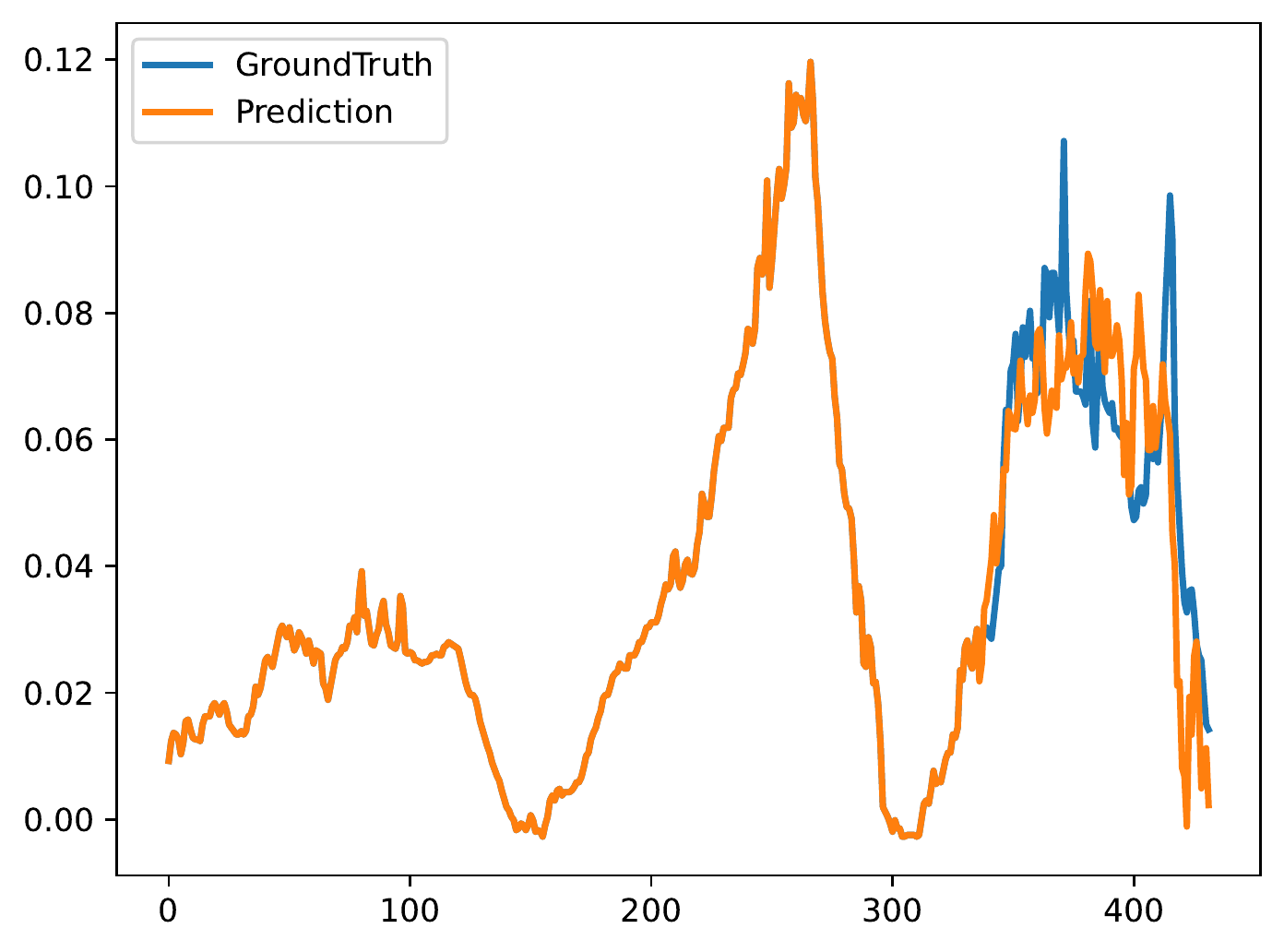}}
    \centerline{(c) MICN}
    \centerline{\includegraphics[width=\textwidth]{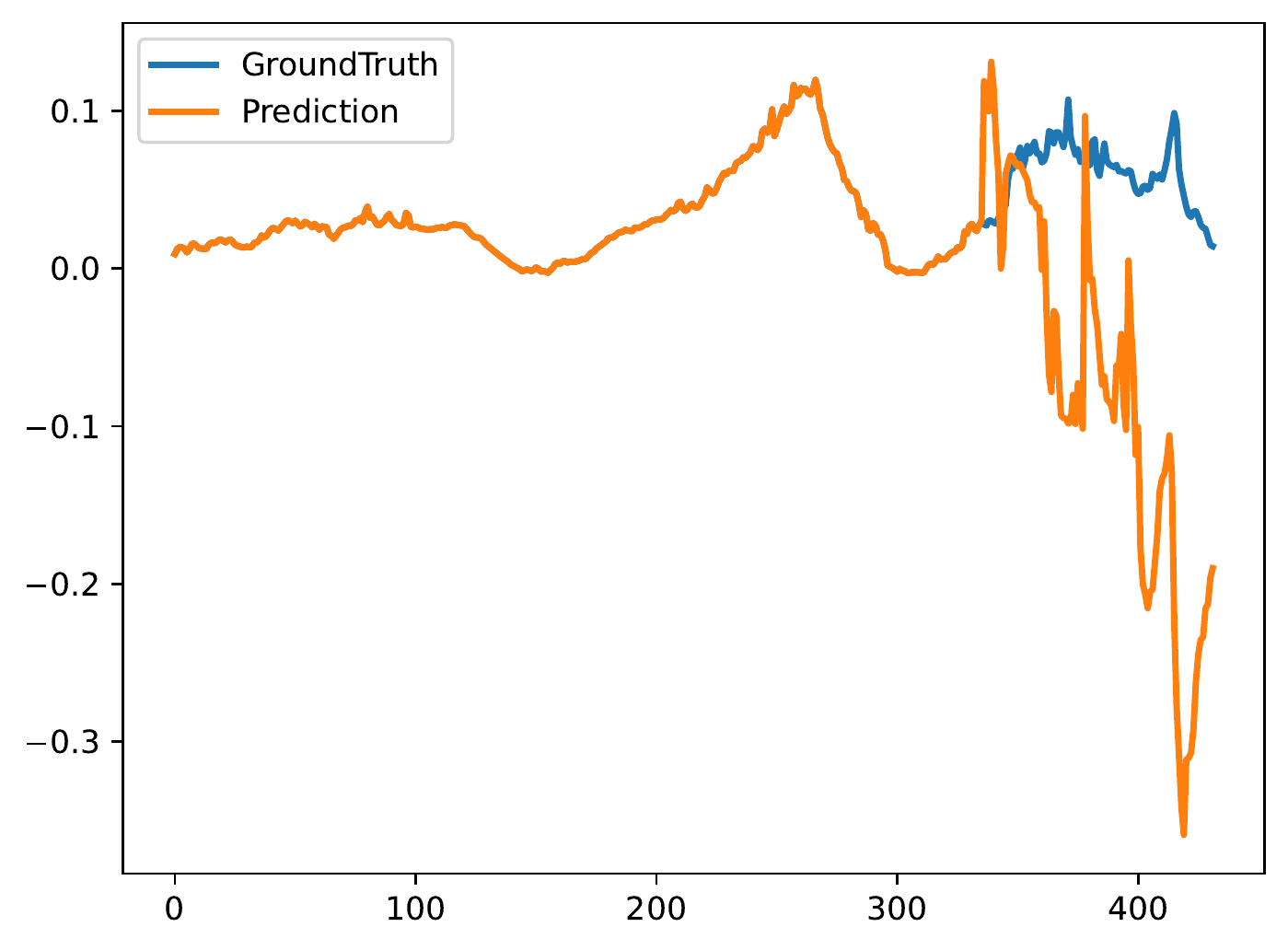}}
    \centerline{(f) Informer}
  \end{minipage}
  \caption{The prediction results on the Weather dataset under the input-336-predict-96 settings.}
  \label{weather_96_plot}
\end{figure}

\begin{figure}[htbp]
  \centering
  \begin{minipage}{0.32\textwidth}
    \centerline{\includegraphics[width=\textwidth]{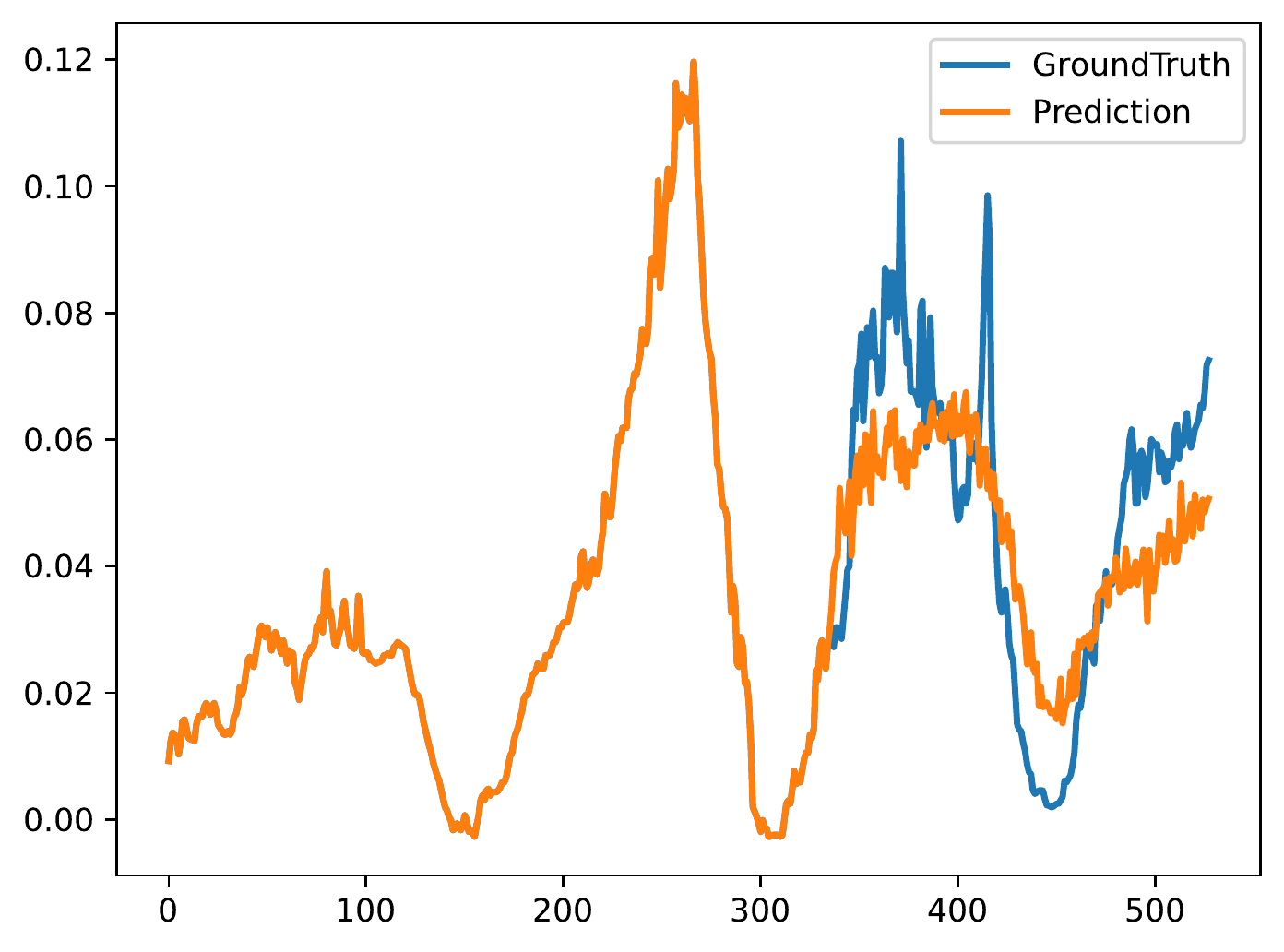}}
    \centerline{(a) MPPN}
    \centerline{\includegraphics[width=\textwidth]{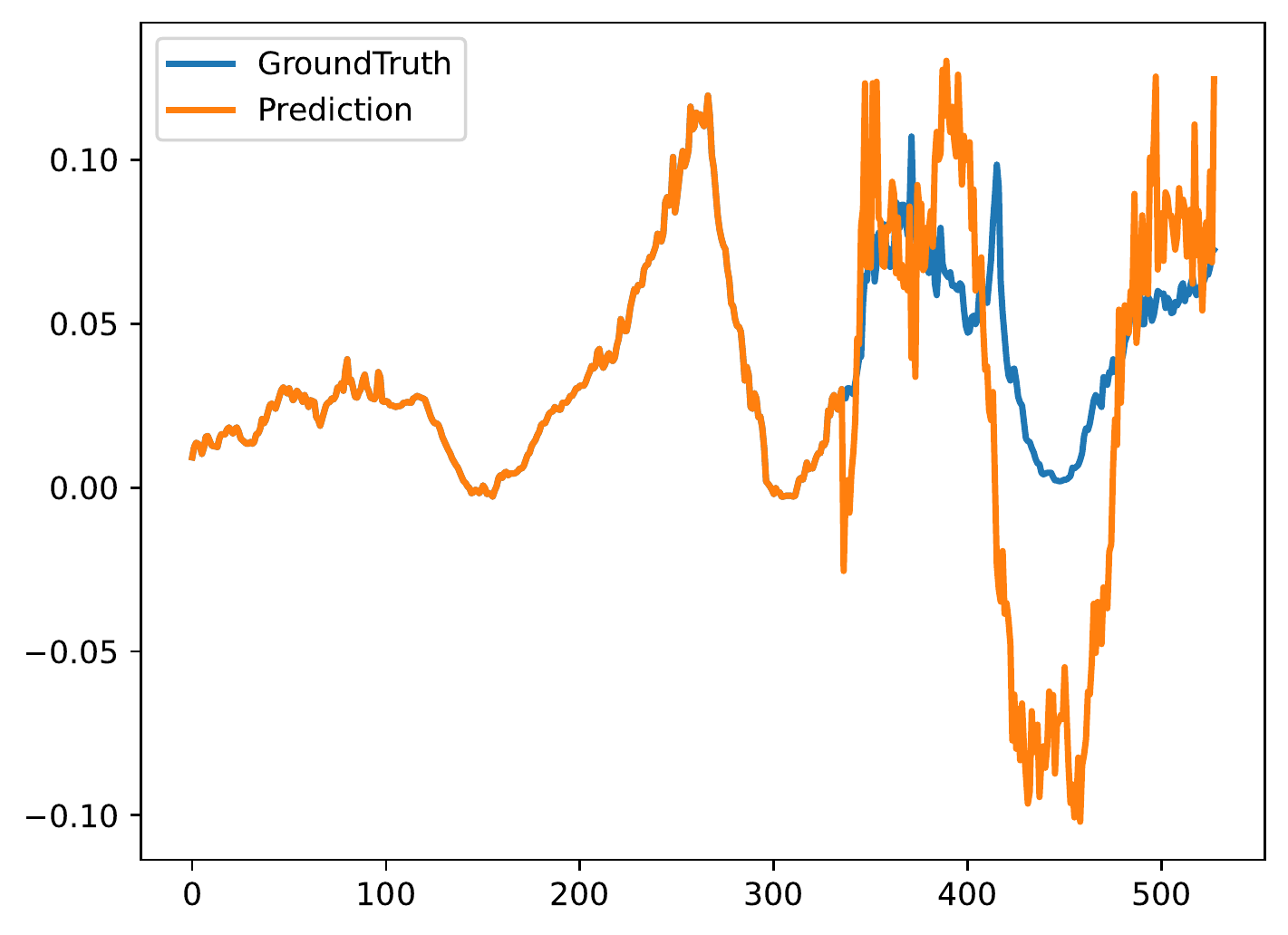}}
    \centerline{(d) FEDformer}
  \end{minipage}
  \hfill 
  \begin{minipage}{0.32\textwidth}
    \centerline{\includegraphics[width=\textwidth]{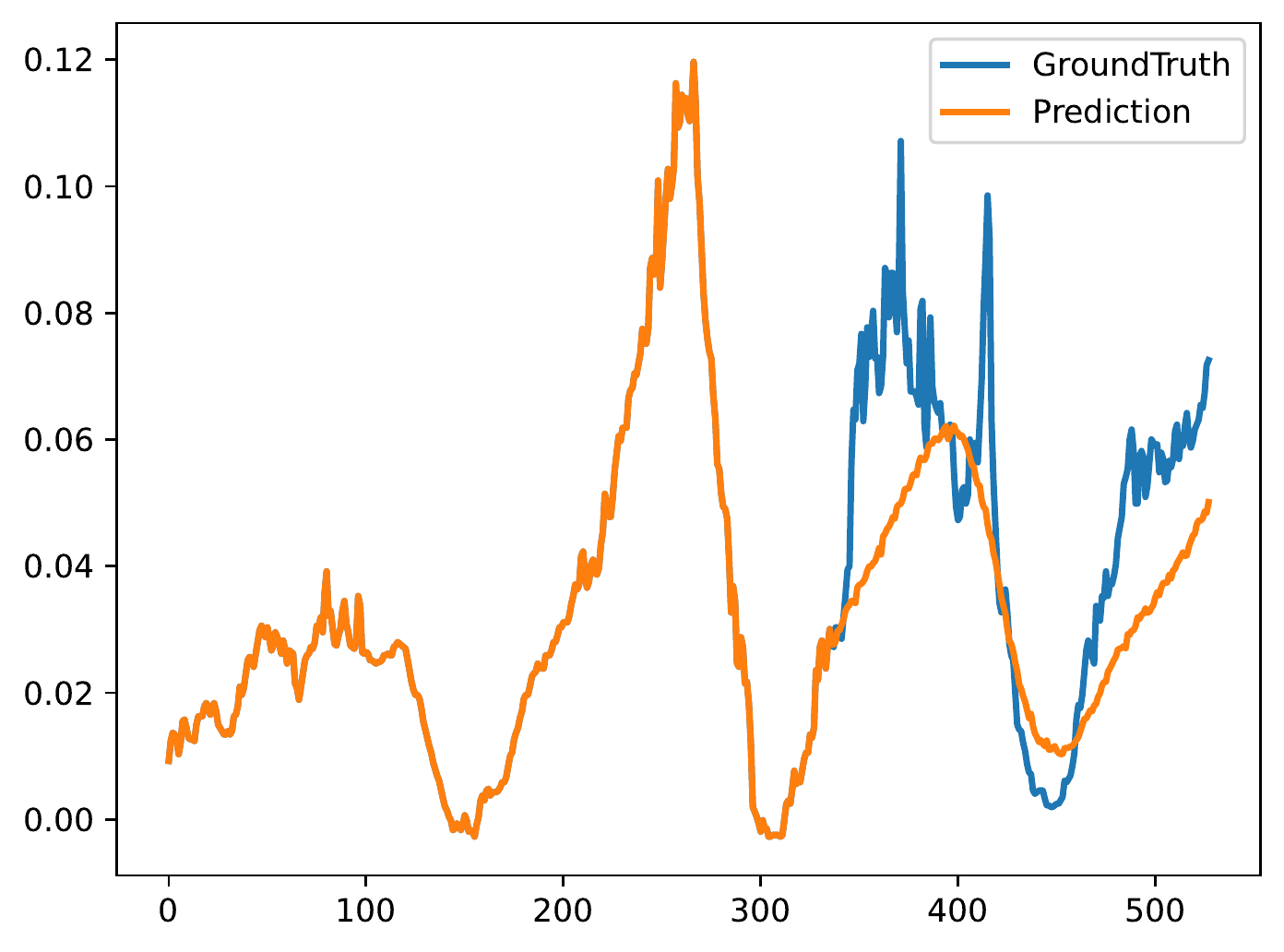}}
    \centerline{(b) DLinear}  
    \centerline{\includegraphics[width=\textwidth]{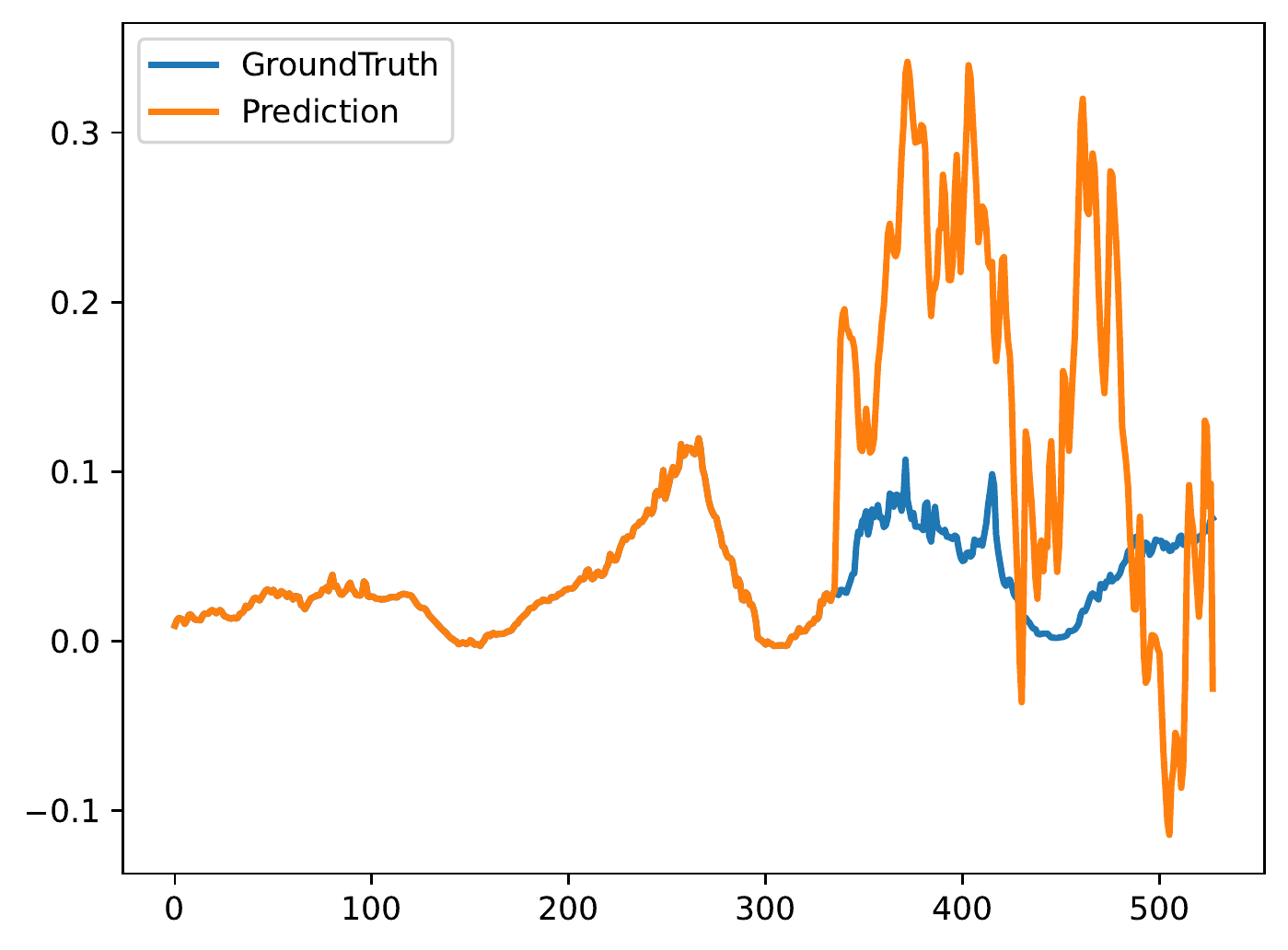}}
    \centerline{(d) Autoformer}
  \end{minipage}
  \hfill
  \begin{minipage}{0.32\textwidth}
    \centerline{\includegraphics[width=\textwidth]{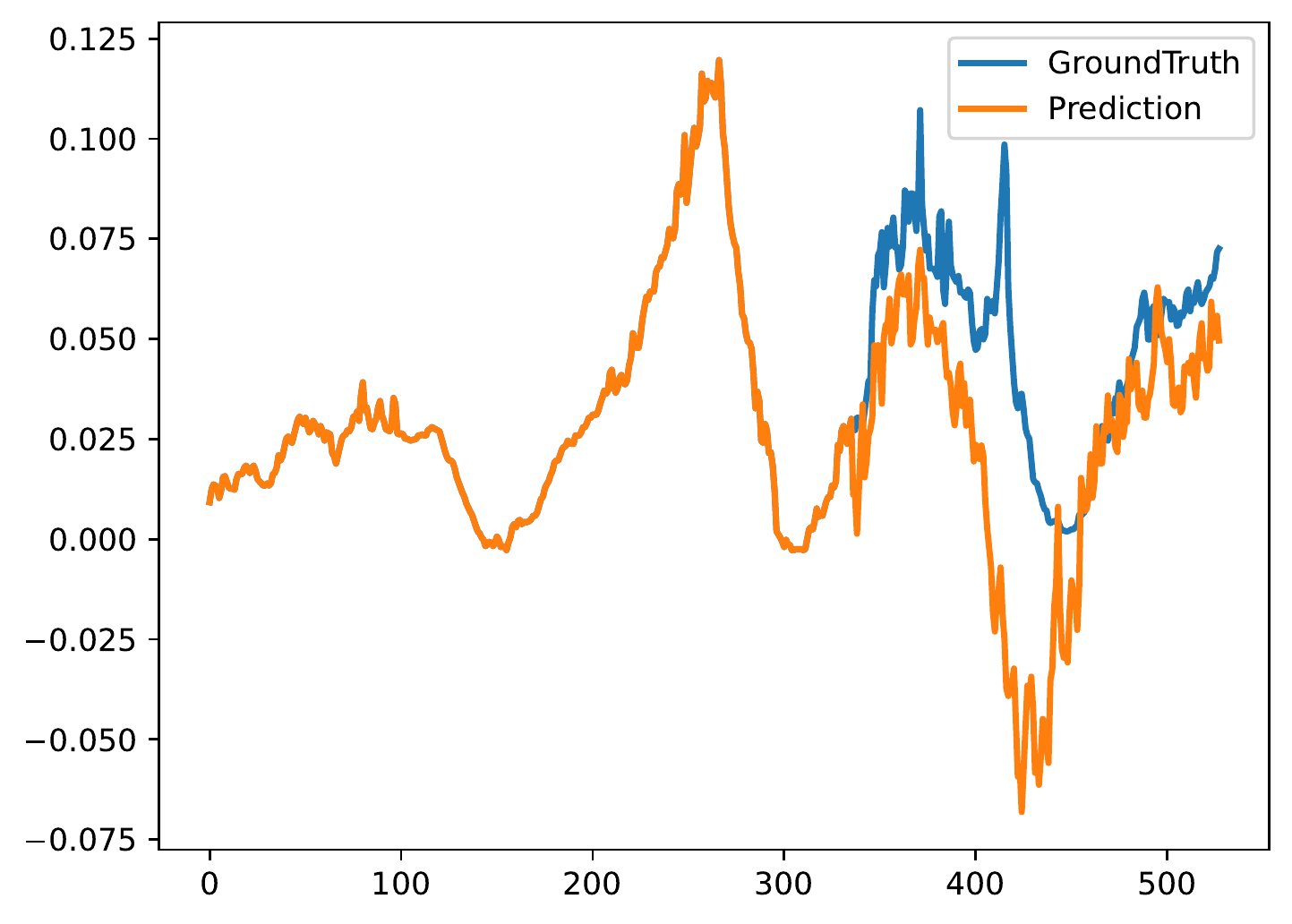}}
    \centerline{(c) MICN}
    \centerline{\includegraphics[width=\textwidth]{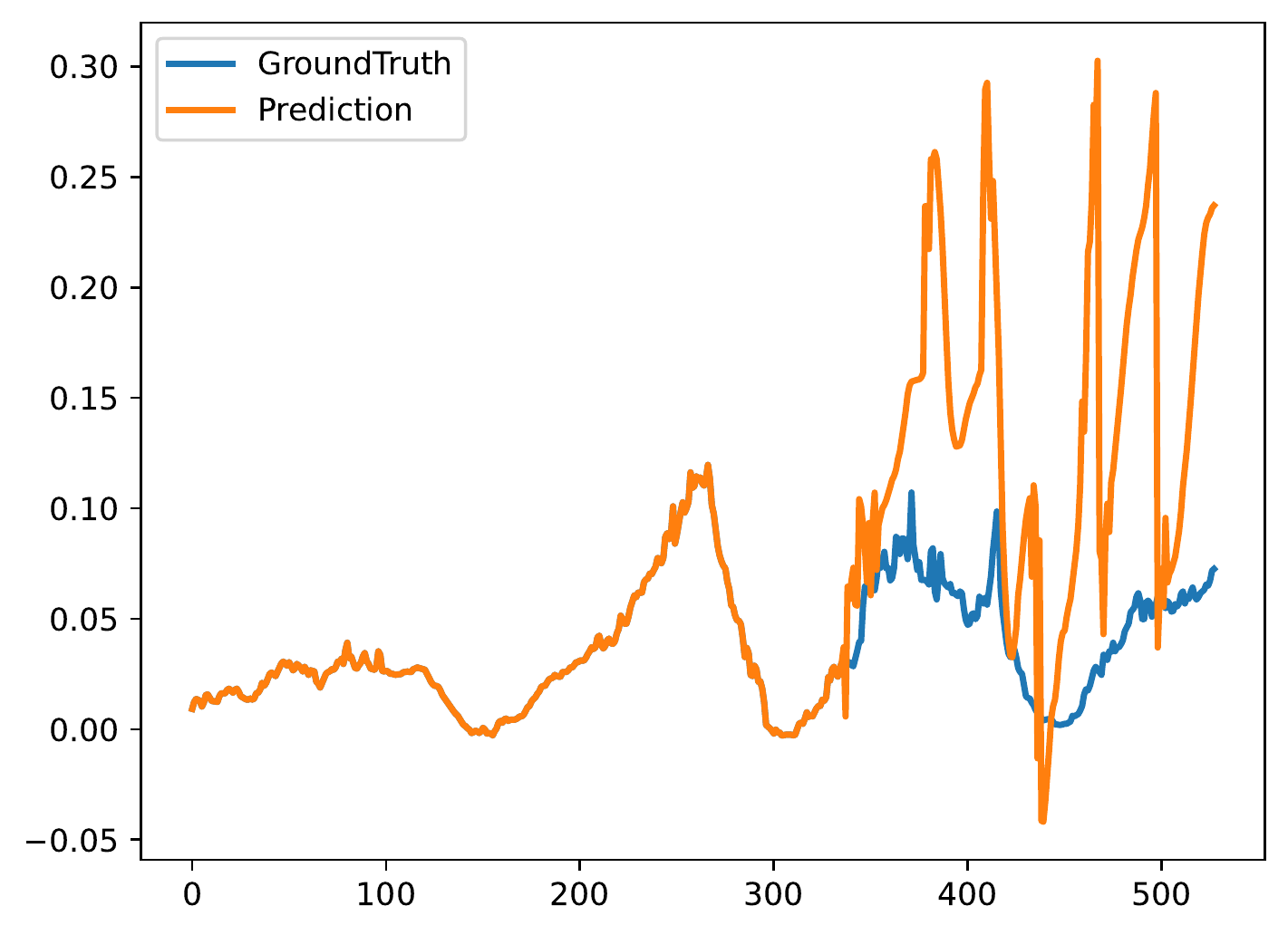}}
    \centerline{(f) Informer}
  \end{minipage}
  \caption{The prediction results on the Weather dataset under the input-336-predict-192 settings.}
  \label{weather_192_plot}
\end{figure}

\begin{figure}[htbp]
  \centering
  \begin{minipage}{0.32\textwidth}
    \centerline{\includegraphics[width=\textwidth]{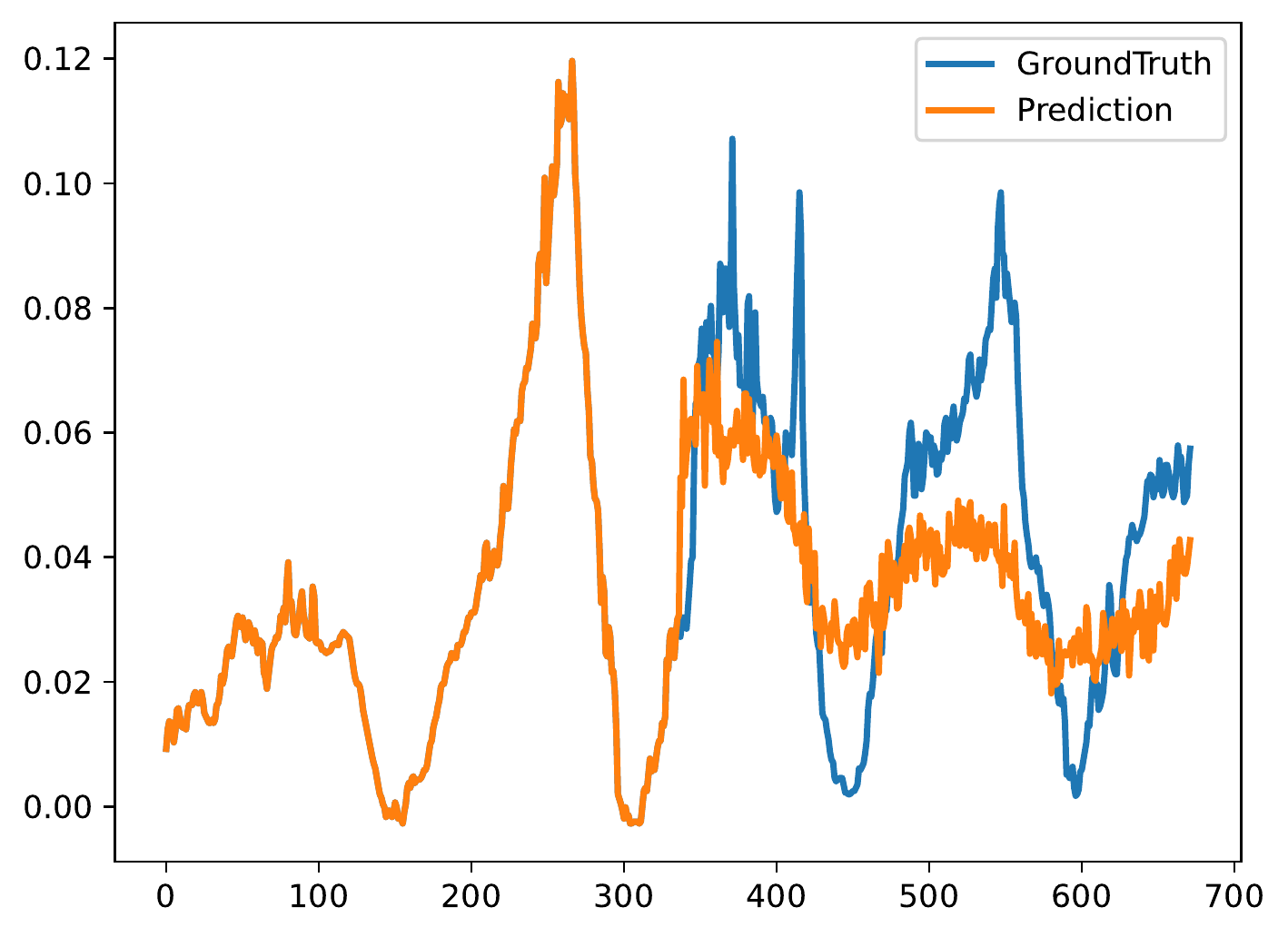}}
    \centerline{(a) MPPN}
    \centerline{\includegraphics[width=\textwidth]{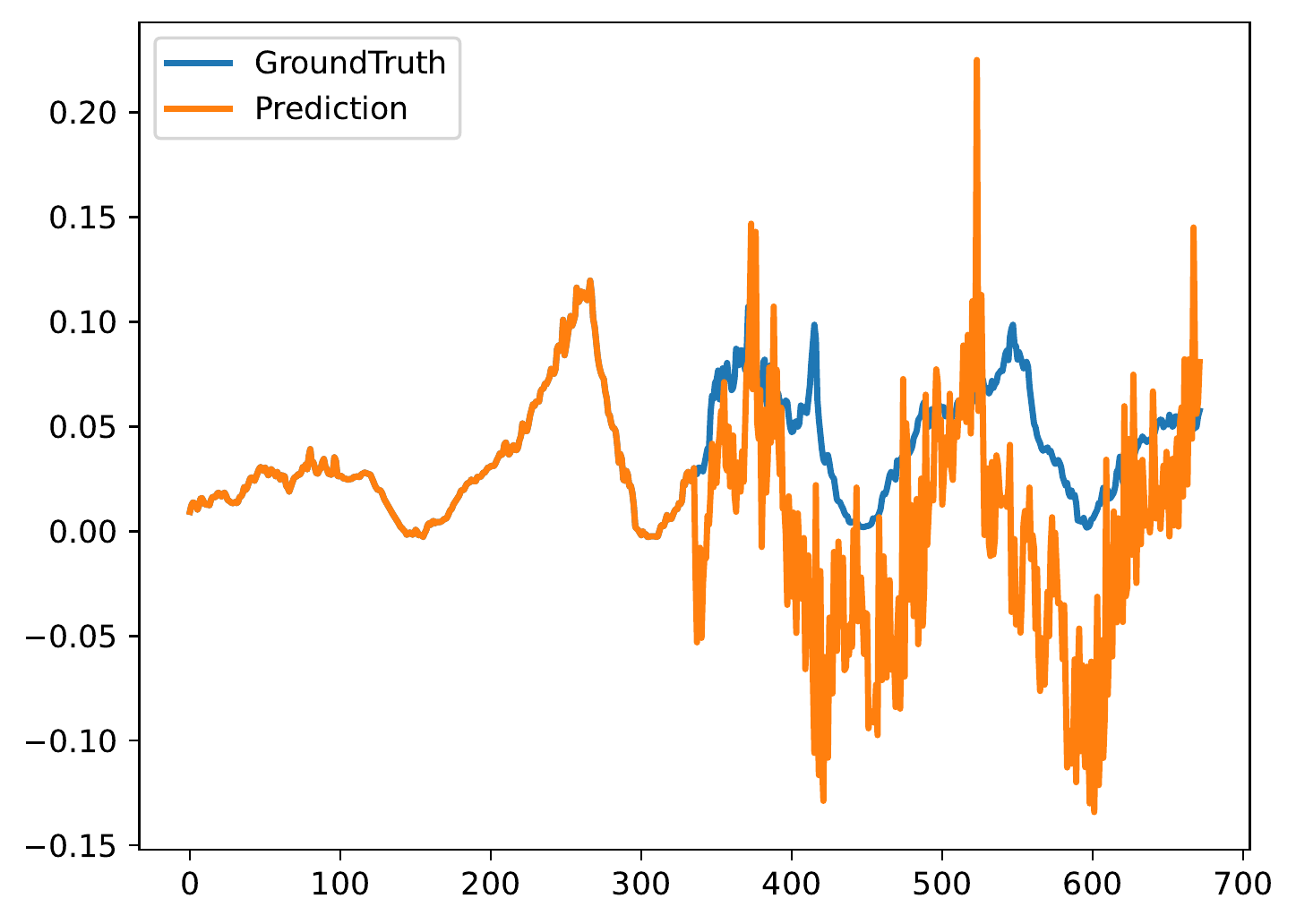}}
    \centerline{(d) FEDformer}
  \end{minipage}
  \hfill 
  \begin{minipage}{0.32\textwidth}
    \centerline{\includegraphics[width=\textwidth]{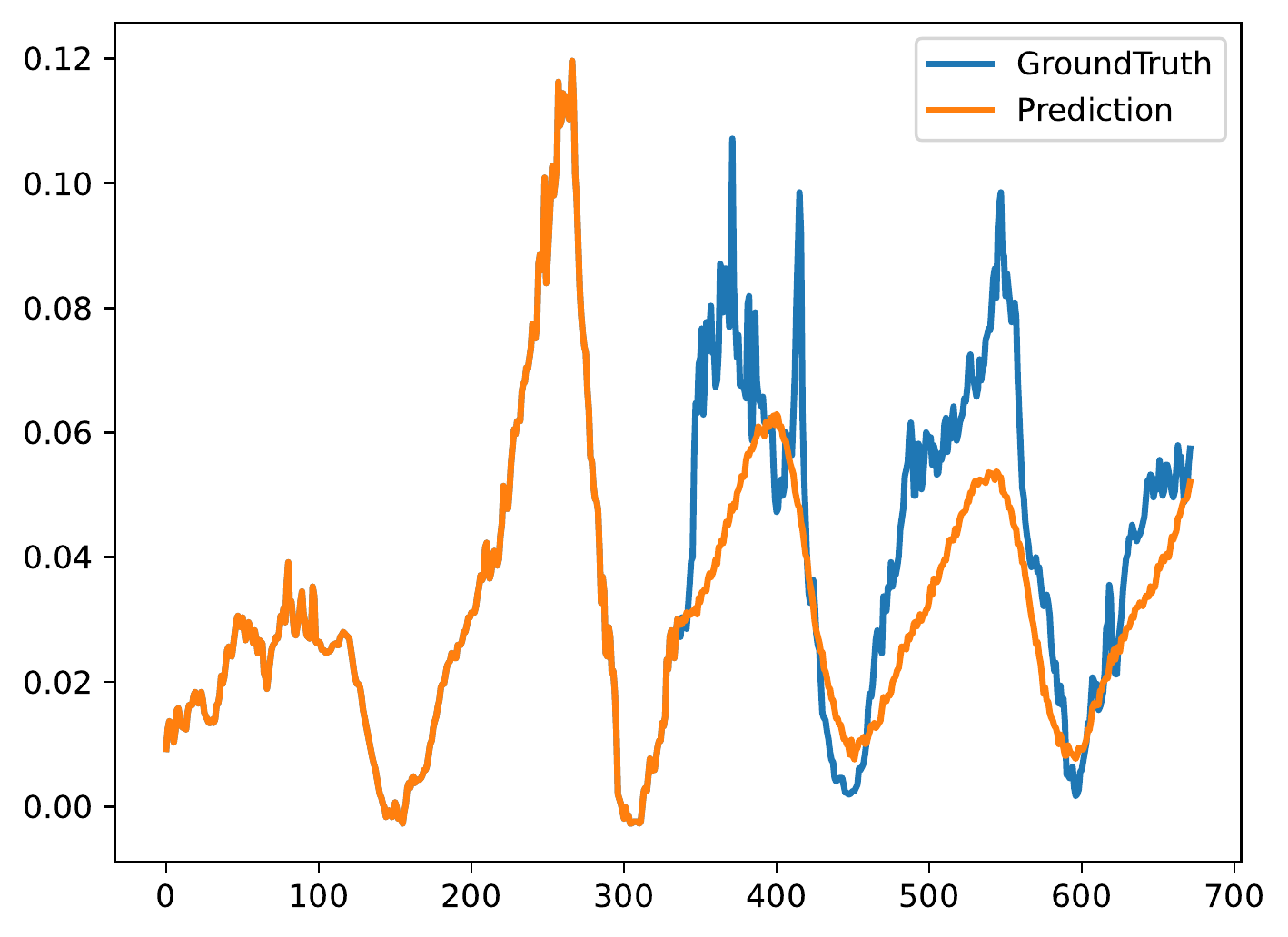}}
    \centerline{(b) DLinear}  
    \centerline{\includegraphics[width=\textwidth]{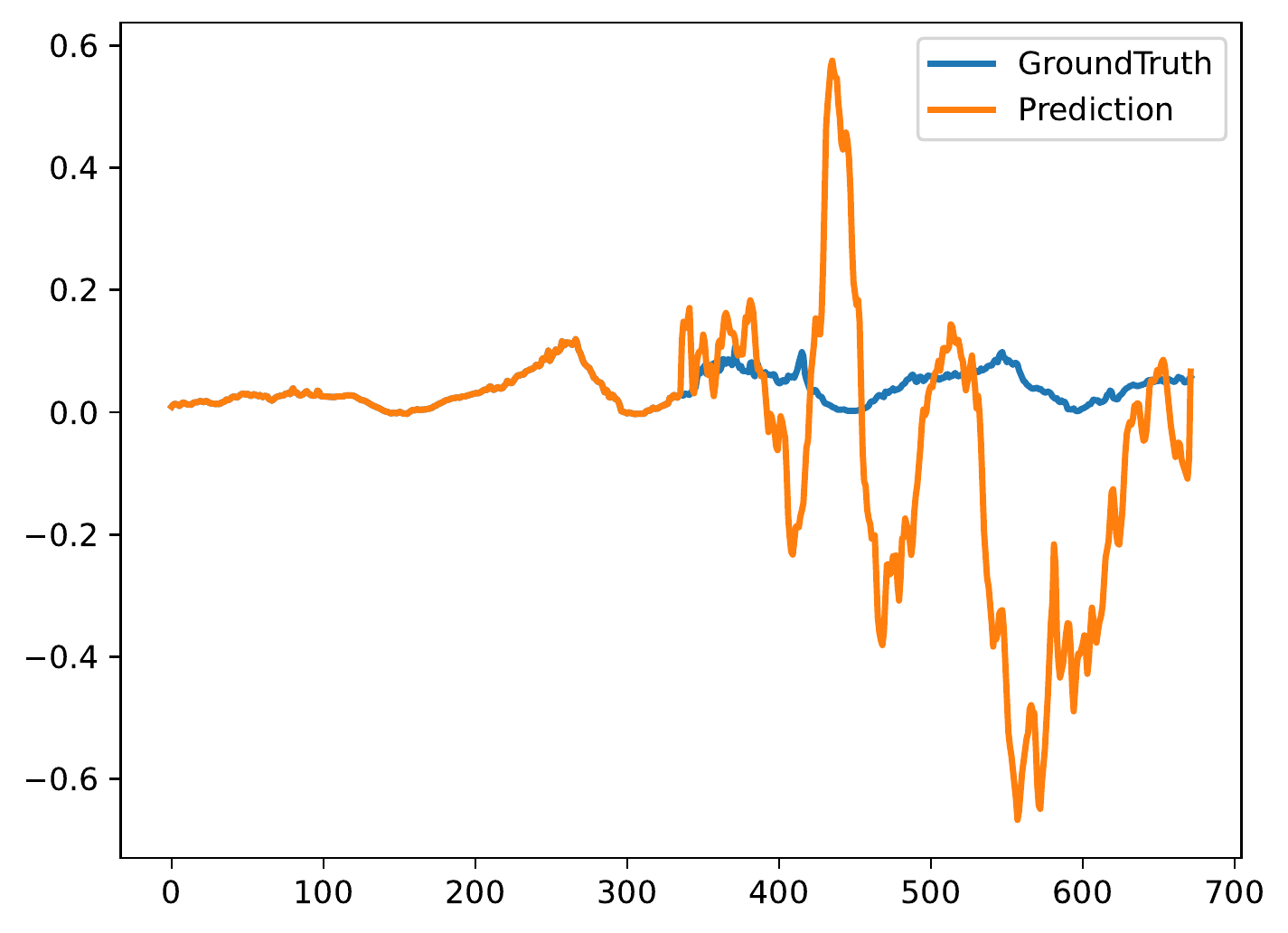}}
    \centerline{(d) Autoformer}
  \end{minipage}
  \hfill
  \begin{minipage}{0.32\textwidth}
    \centerline{\includegraphics[width=\textwidth]{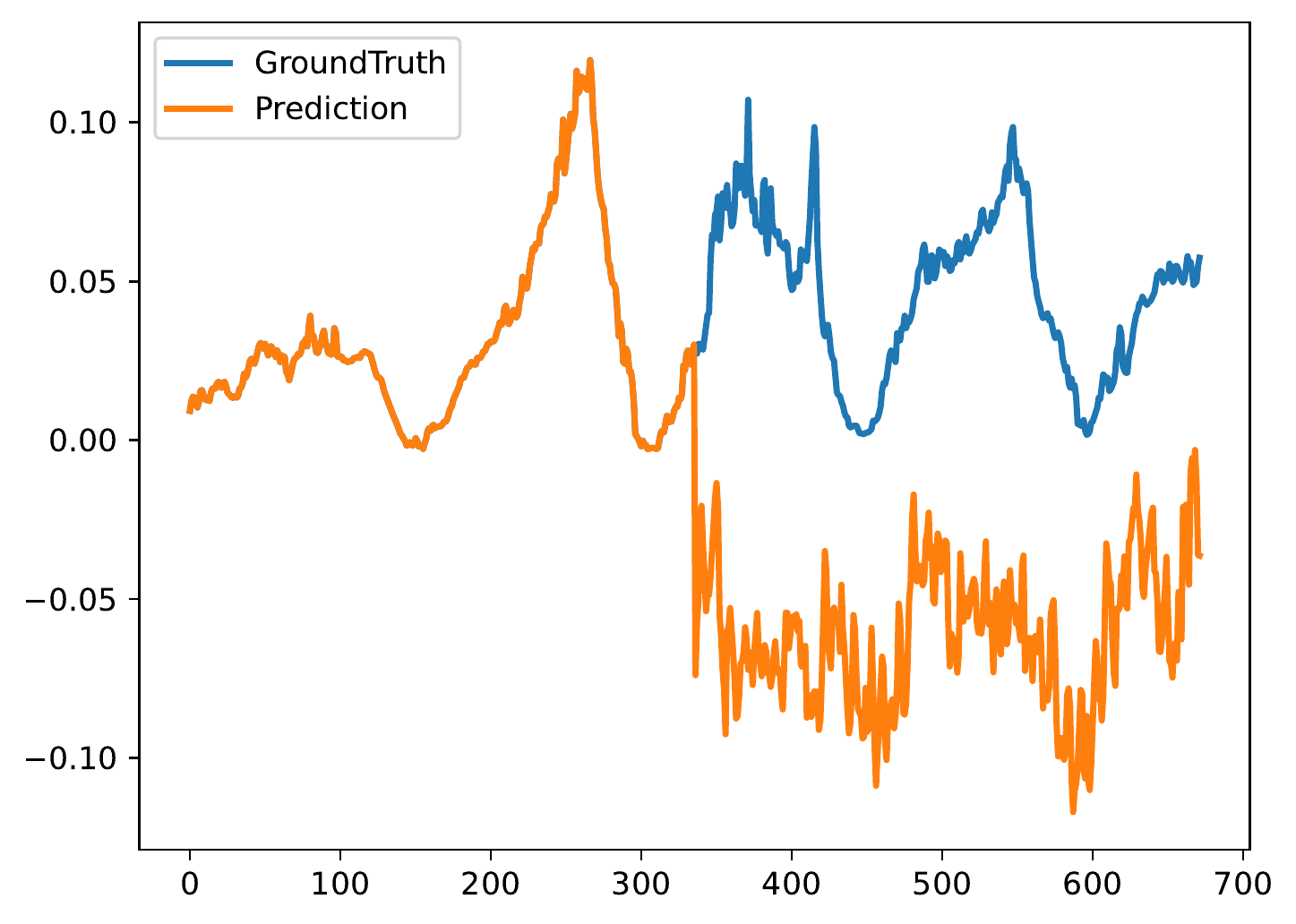}}
    \centerline{(c) MICN}
    \centerline{\includegraphics[width=\textwidth]{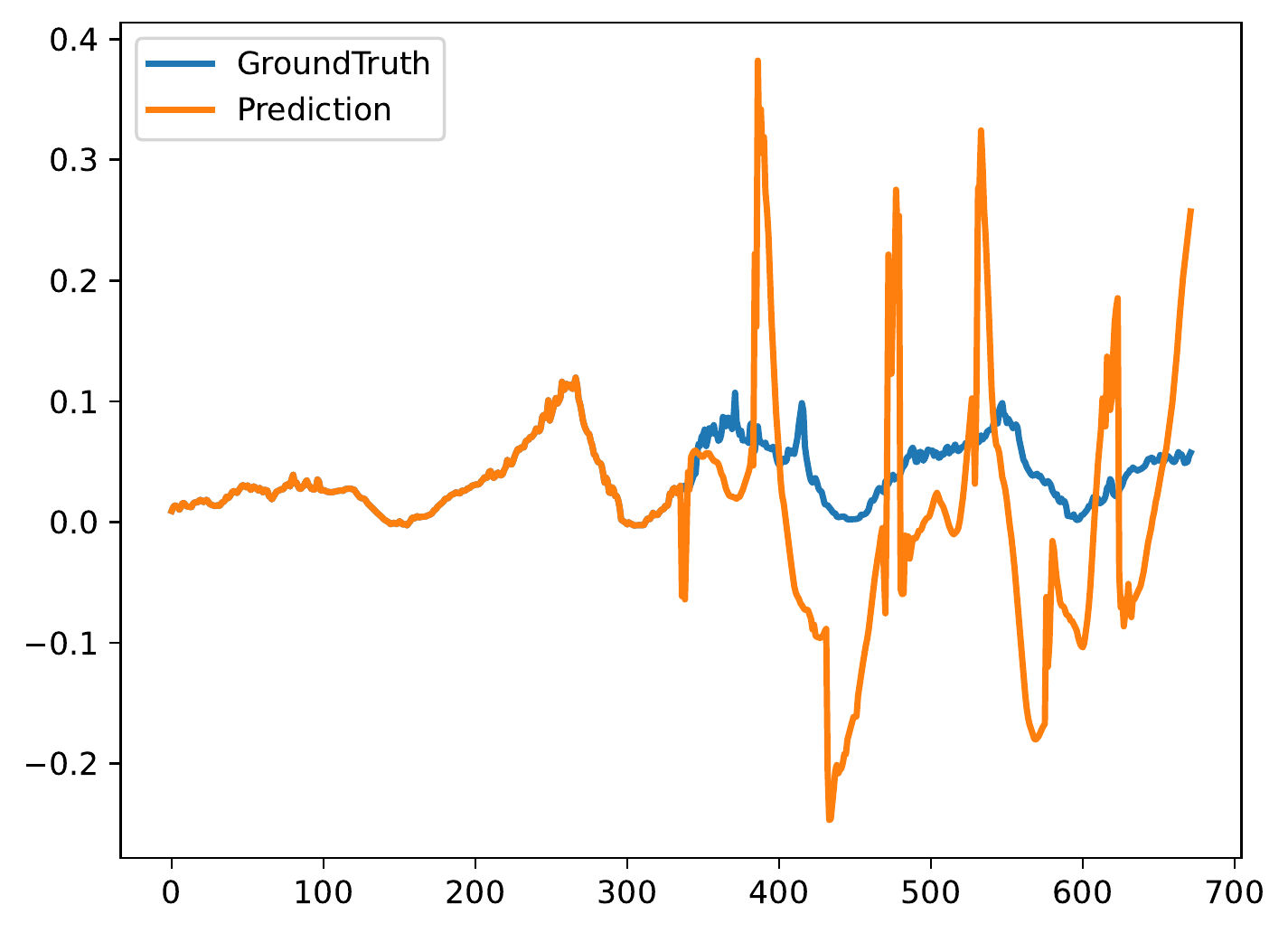}}
    \centerline{(d) Informer}
  \end{minipage}
  \caption{The prediction results on the Weather dataset under the input-336-predict-336 settings.}
  \label{weather_336_plot}
\end{figure}

\begin{figure}[htbp]
  \centering
  \begin{minipage}{0.32\textwidth}
    \centerline{\includegraphics[width=\textwidth]{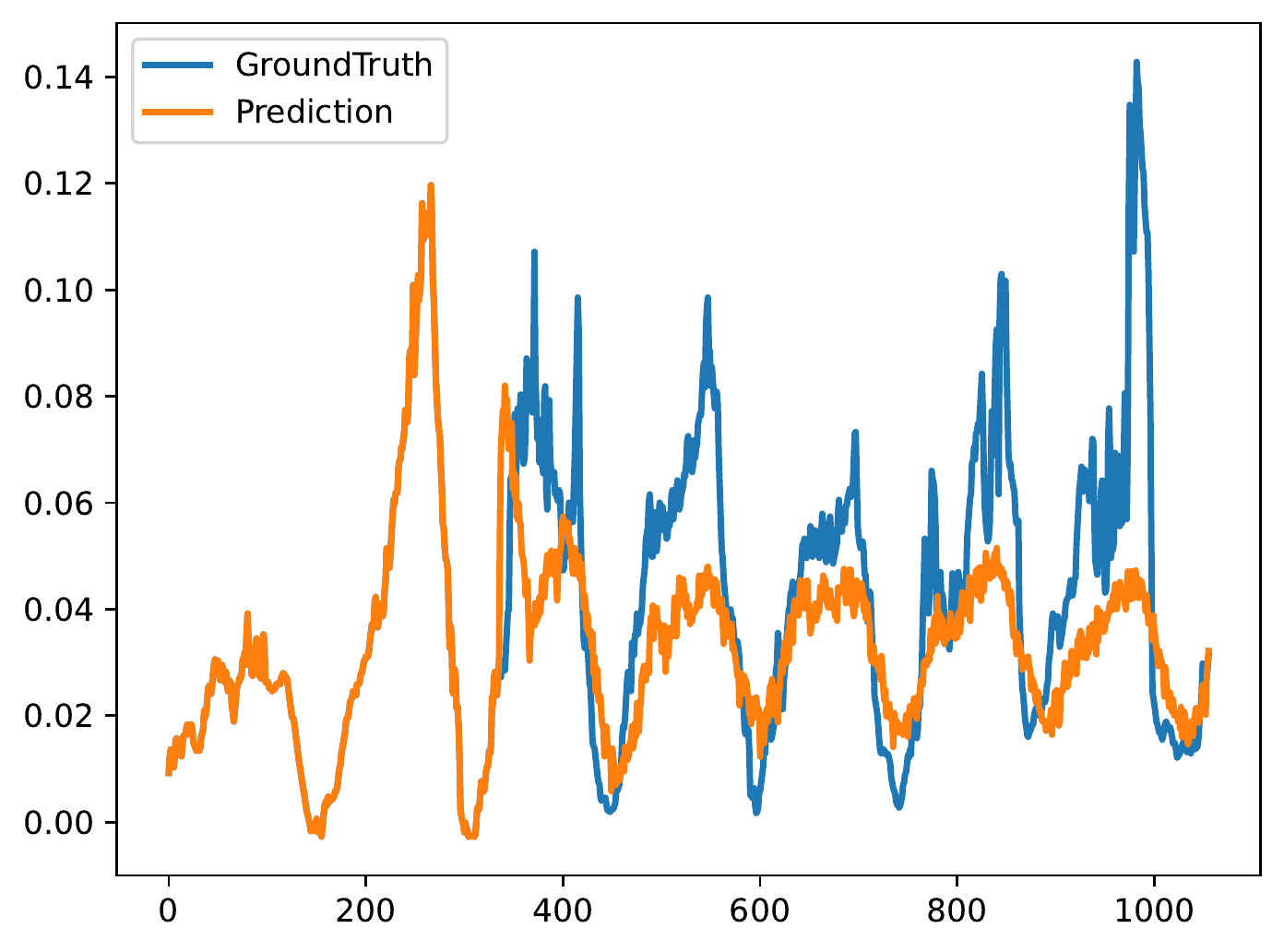}}
    \centerline{(a) MPPN}
    \centerline{\includegraphics[width=\textwidth]{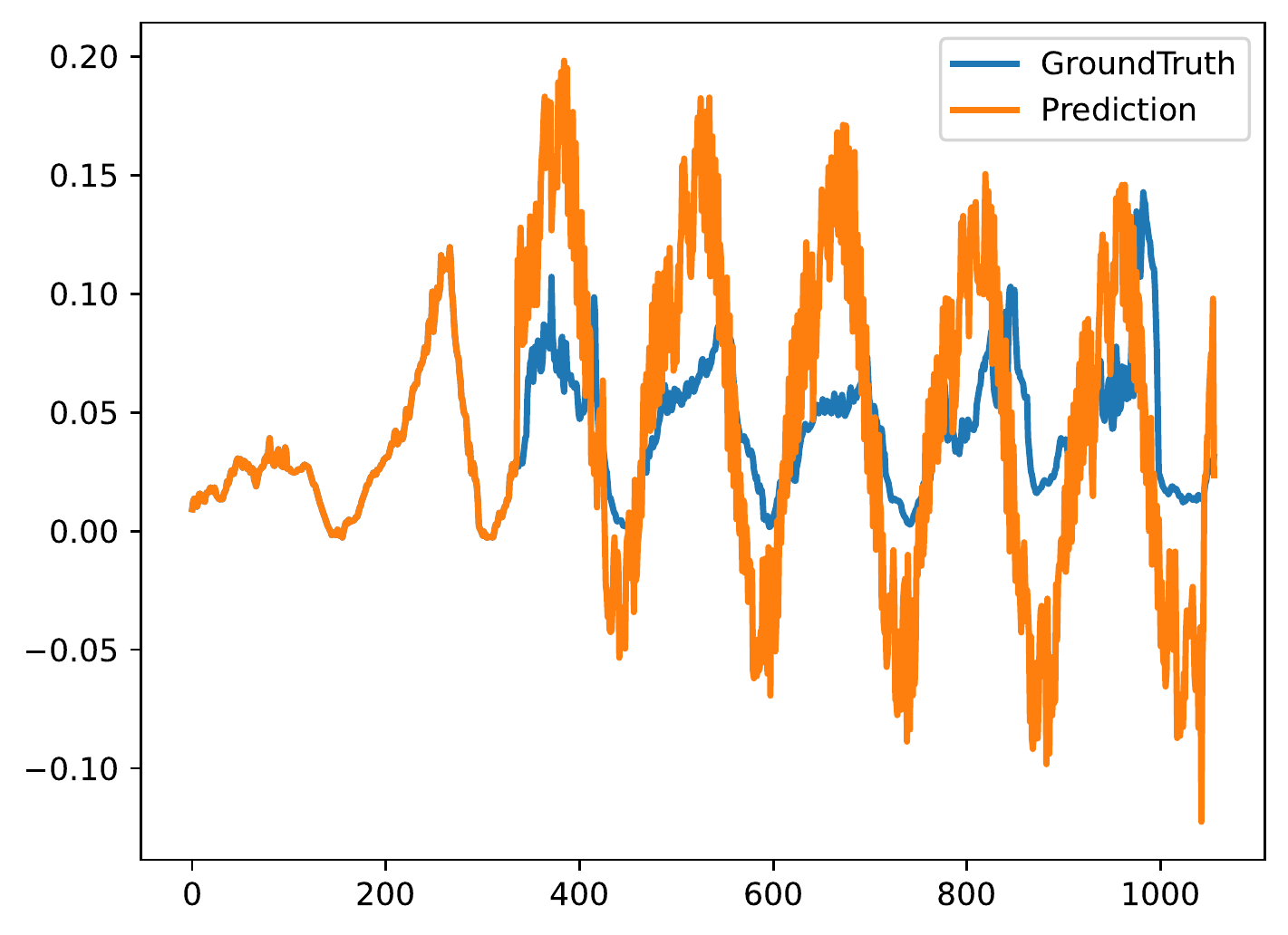}}
    \centerline{(d) FEDformer}
  \end{minipage}
  \hfill 
  \begin{minipage}{0.32\textwidth}
    \centerline{\includegraphics[width=\textwidth]{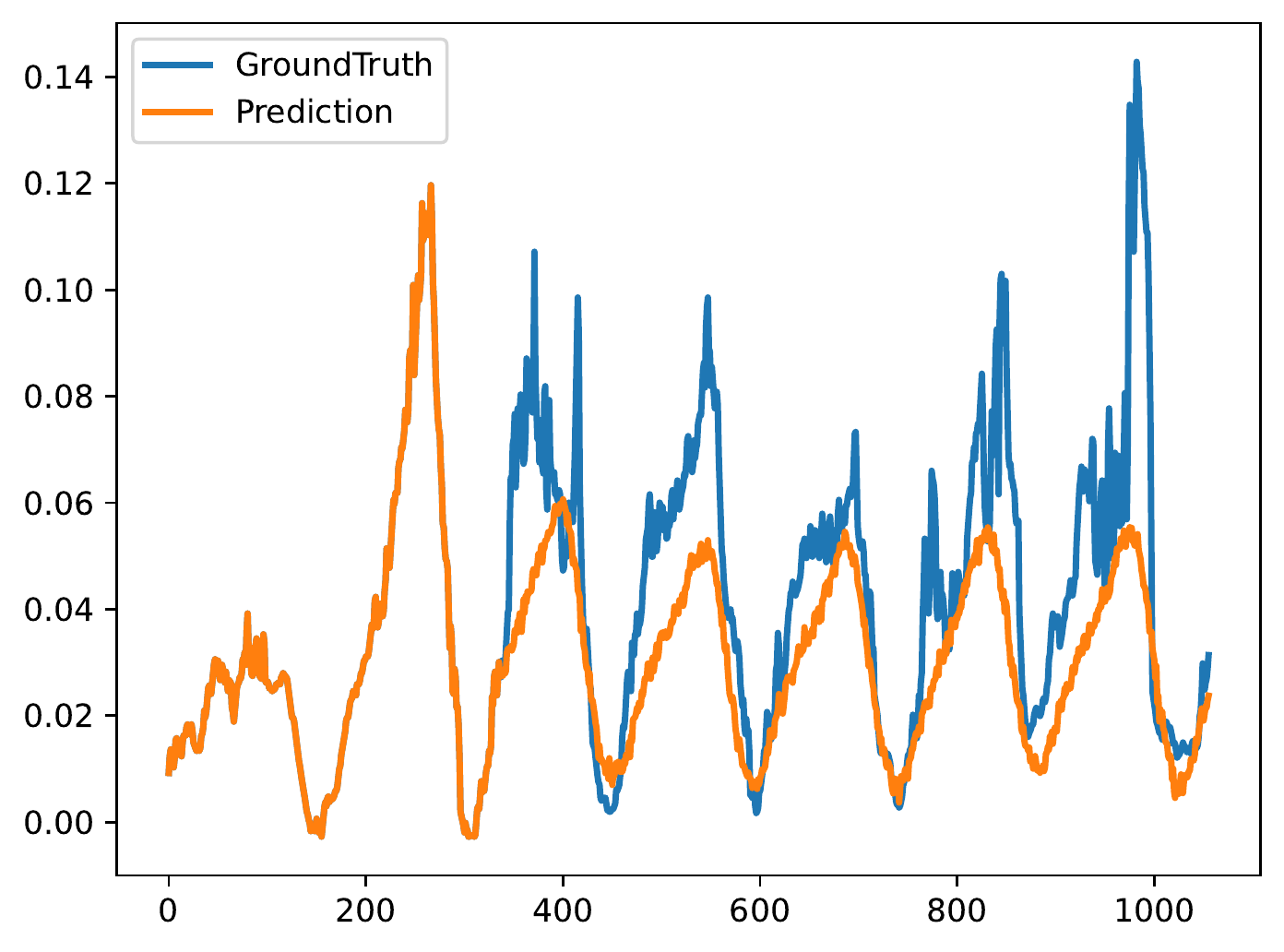}}
    \centerline{(b) DLinear}  
    \centerline{\includegraphics[width=\textwidth]{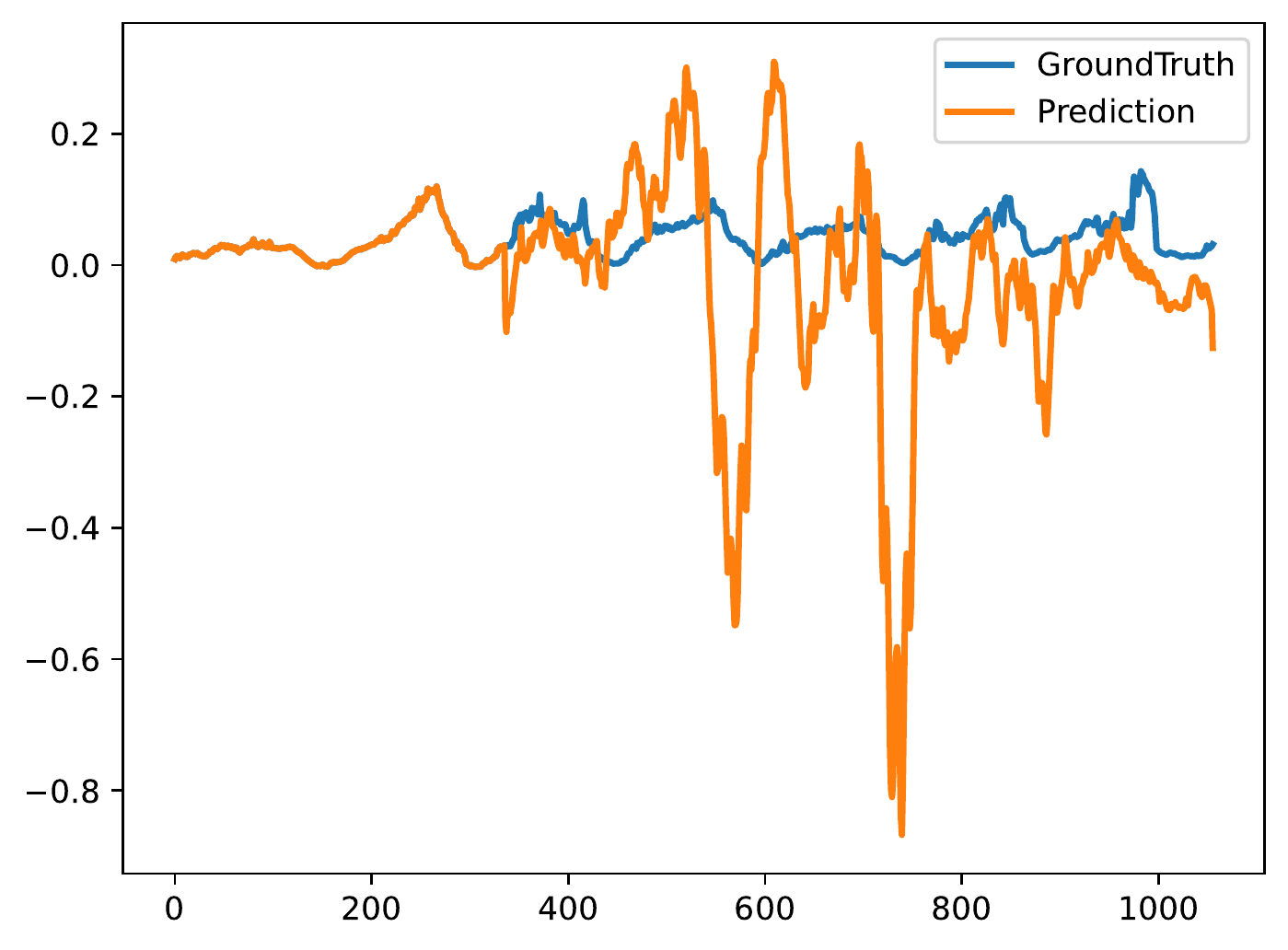}}
    \centerline{(d) Autoformer}
  \end{minipage}
  \hfill
  \begin{minipage}{0.32\textwidth}
    \centerline{\includegraphics[width=\textwidth]{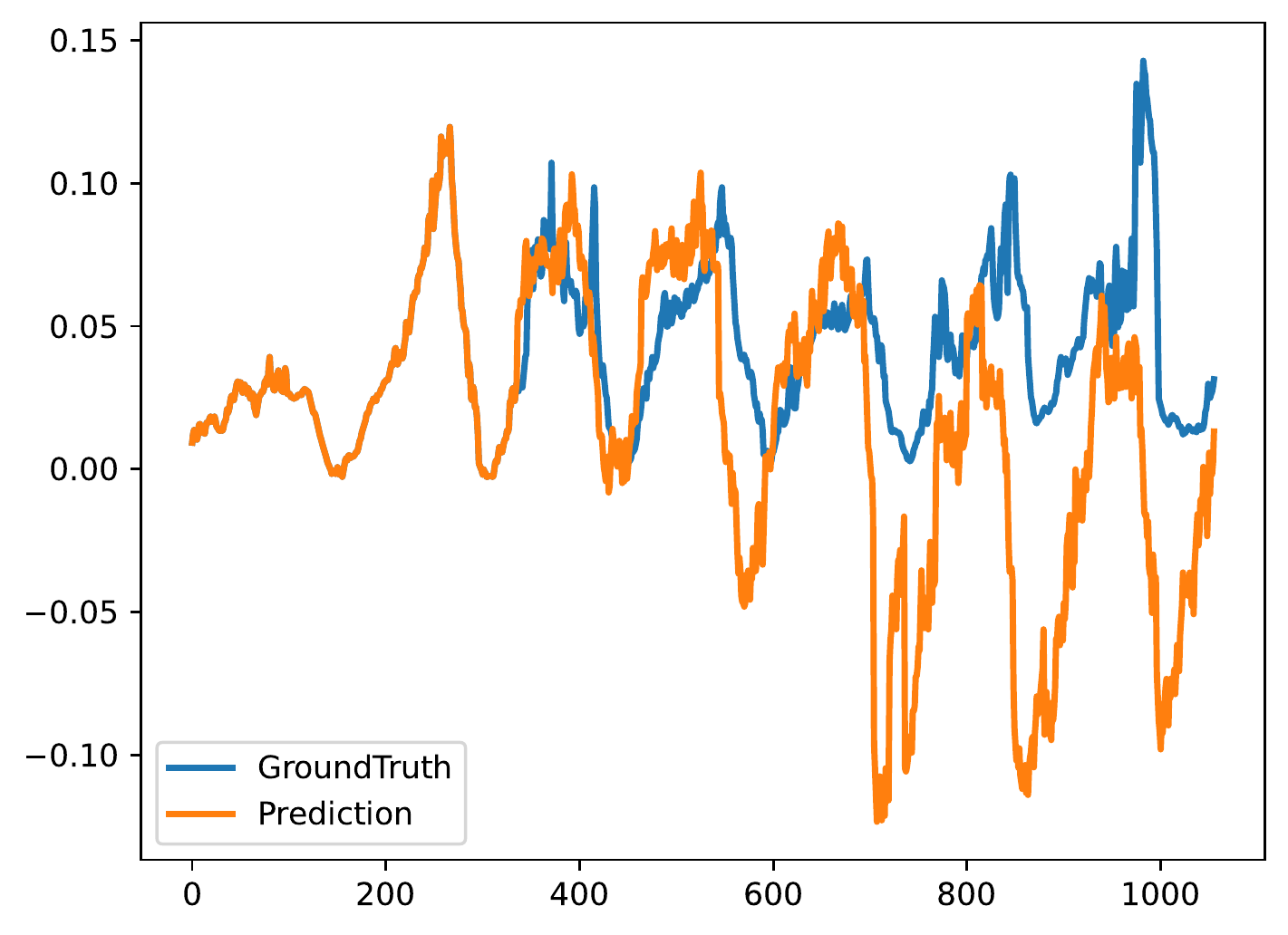}}
    \centerline{(c) MICN}
    \centerline{\includegraphics[width=\textwidth]{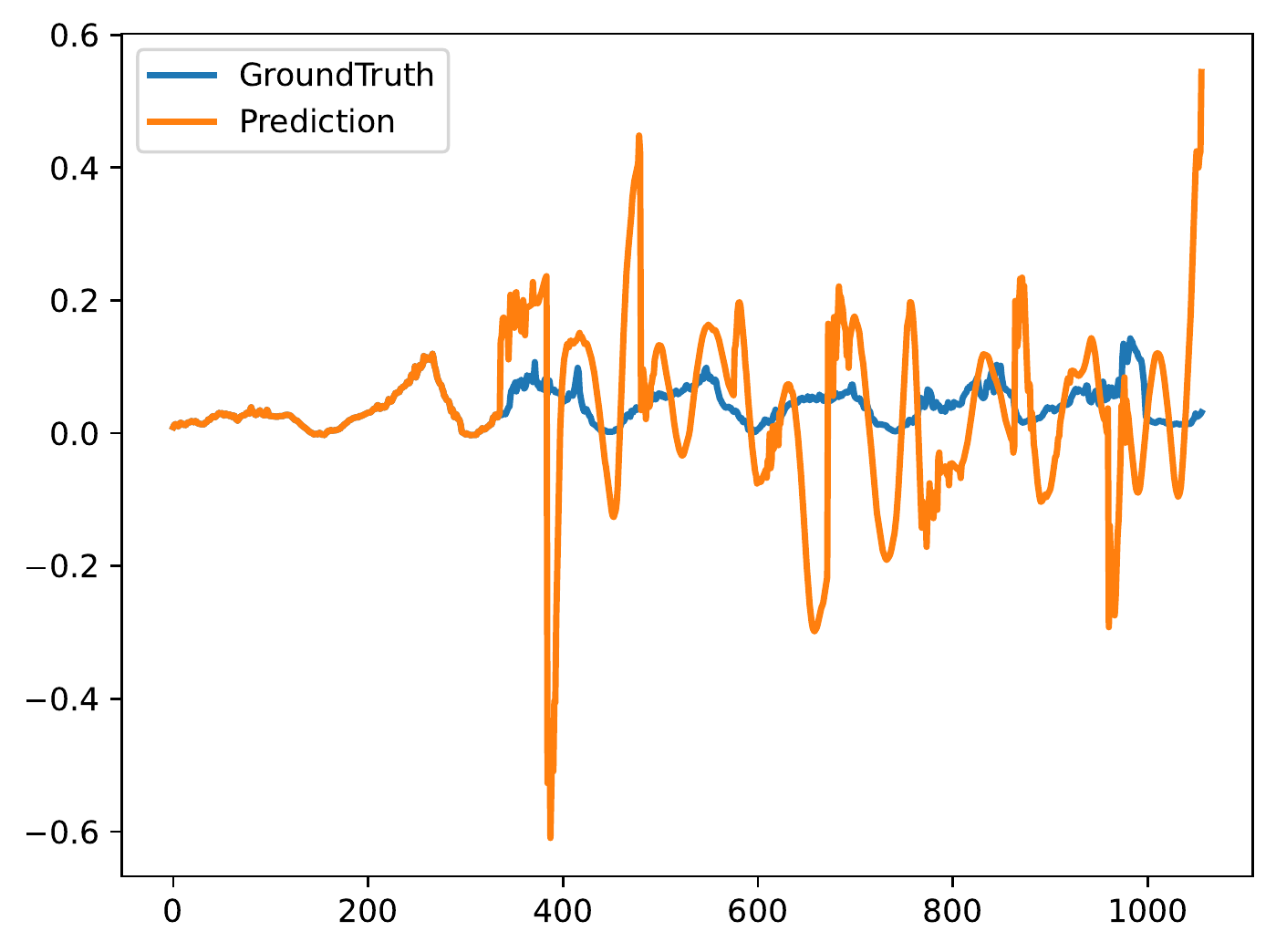}}
    \centerline{(d) Informer}
  \end{minipage}
  \caption{The prediction results on the Weather dataset under the input-336-predict-720 settings.}
  \label{weather_720_plot}
\end{figure}

We plot the multivariate forecasting results of our MPPN and several baseline models on the ETTh1 and Weather dataset. All results are from the testing set for qualitative comparison. As depicted in Figure \ref{etth1_96_plot}-\ref{weather_720_plot}, our model outperforms the baseline models, exhibiting superior performance in modeling  periodicity, short-term fluctuations, and long-term variations of time series. It demonstrates remarkable proficiency in capturing the complex patterns inherent in temporal series data, showcasing its effectiveness in accurately representing and analyzing diverse data characteristics over different time resolutions.

\section{Limitations and future work}

In this paper, we focus on extracting distinctive patterns to represent time series for long-term time series forecasting. We adopt multi-resolution patching and periodic pattern mining mechanism to explicitly extract the predictable patterns from time series data. Although MPPN can effectively capture patterns (e.g., periodicity and trend) in historical time series, most real-world time series can be very complex and influenced by various external factors, which may result in intricate patterns or even previously unseen variations. For example, if a traffic accident occurs by chance on a certain road causing traffic congestion, it will likely affect the speed of the road section; if extreme weather disasters (such as earthquakes and hurricanes) suddenly occur in a certain area, they will greatly affect the weather indicators. Sometimes, accidental events may also cause the peaks and valleys of the time series to arrive earlier or later, such as a customer staying up late to watch a World Cup game, resulting in a delay in electricity usage at home compared to historical baseline. Although MPPN can quickly capture these variations from the data of lookback window, it is difficult to foresee them. 

In the future, we will consider how to capture the varying time series patterns and model the external factors. We plan to introduce knowledge graph for incorporating  the knowledge of external factors to the prediction model. 
With the remarkable performance of foundation models in other fields, we will further investigate the general patterns of time series and spatial-temporal series data and construct task-general models to support multiple downstream time series analysis tasks. We believe that this direction is a particularly intriguing and significant.






\end{document}